\documentclass[10pt,journal,compsoc]{IEEEtran}
\ifCLASSINFOpdf
\else
\fi

\usepackage{cite}
\usepackage{times}
\usepackage{epsfig}
\usepackage{graphicx}
\usepackage{amsmath}
\usepackage{amssymb}
\usepackage{multirow}
\usepackage{amsthm}
\usepackage{subfigure}
\usepackage{color}
\usepackage{csquotes}
\usepackage{soul}
\usepackage{enumerate}
\usepackage[normalem]{ulem}
\usepackage{algorithm}
\usepackage{algorithmicx}
\usepackage{algpseudocode}
\usepackage{booktabs}
\usepackage{array}
\usepackage{rotating}
\usepackage{arydshln}
\usepackage{ragged2e}
\usepackage[colorlinks=true,linkcolor=black,citecolor=blue,urlcolor=blue,]{hyperref}
\newcommand\eg{\emph{e.g.}} 
\newcommand\ie{\emph{i.e.}} 
 
\newcommand\etc{\emph{etc.}}
\newcommand\etcn{\emph{etc. }}
\newcommand\wrt{w.r.t.} 
\newcommand\etal{\emph{et al.}}

\def\Rev#1{\textcolor{black}{#1}}

\begin{document}

% \title{Transformer Transforms Salient Object Detection}

% \title{Calibrated Transformer for Accurate and Reliable Salient Object Detection}
\title{Generative Transformer for Accurate and Reliable Salient Object Detection}


\author{Yuxin~Mao,
        Jing~Zhang,
        Zhexiong~Wan,
        Yuchao~Dai*, \\
        Aixuan~Li,
        Yunqiu~Lv,
        Xinyu~Tian,
        Deng-Ping Fan
        and~Nick~Barnes% <-this % stops a space
\IEEEcompsocitemizethanks{\IEEEcompsocthanksitem Yuxin~Mao, Zhexiong~Wan, Yuchao~Dai, Aixuan~Li, Yunqiu~Lv and Xinyu~Tian are with School of Electronics and Information, Northwestern Polytechnical University, Xi'an, China.
\IEEEcompsocthanksitem Jing Zhang and Nick Barnes are with School of Computing, Australian National University, Canberra, Australia.
\IEEEcompsocthanksitem Deng-Ping Fan is with the CS, Nankai University,
Tianjin, China.
\IEEEcompsocthanksitem Corresponding author: Yuchao~Dai (Email: daiyuchao@nwpu.edu.cn).

\IEEEcompsocthanksitem The source code and experimental results are publicly available via our project page: \url{https://github.com/fupiao1998/TransformerSOD}.

% \IEEEcompsocthanksitem J. Doe and J. Doe are with Anonymous University.
}% <-this % stops an unwanted space
% \thanks{Manuscript received April 19, 2005; revised August 26, 2015.}
}


\markboth{Journal of \LaTeX\ Class Files,~Vol.~14, No.~8, August~2015}%
{Shell \MakeLowercase{\textit{et al.}}: Bare Demo of IEEEtran.cls for Computer Society Journals}



\IEEEtitleabstractindextext{%
\begin{abstract}
\justifying
Transformer,
% networks, 
which originates from machine translation, is particularly powerful at modeling long-range dependencies.
% within a long sequence. 
Currently, the transformer is
% networks are 
making revolutionary progress in various vision tasks,
% ranging from high-level classification tasks to low-level dense prediction tasks, 
leading to significant performance improvements compared with the convolutional neural network (CNN) based frameworks.
In this paper, we conduct extensive research on exploiting the contributions of transformers for \emph{accurate} and \emph{reliable}
% to 
salient object detection. \Rev{For the former, we apply transformer to a deterministic model,}
% , achieving both accurate and reliable saliency predictions. 
% \sout{we first investigate transformers for accurate salient object detection with deterministic neural networks, }
and explain that the effective structure modeling and global context modeling abilities lead to its superior performance compared with the CNN based frameworks. \Rev{For the latter, we observe that both CNN and transformer based frameworks suffer greatly from the over-confidence issue, where the models tend to generate wrong predictions with high confidence.
% a stochastic model is designed
}
% \sout{Then, we design stochastic networks }
% to evaluate the transformers' ability in reliable salient object detection. We observe that both CNN and transformer based frameworks suffer greatly from the over-confidence issue, where the models tend to generate wrong predictions with high confidence.
% , leading to over-confident predictions or a poorly-calibrated model. 
To estimate the reliability
% calibration 
degree of both CNN- and transformer-based frameworks, \Rev{we further present a latent variable model, namely inferential generative adversarial network (iGAN), based on the generative adversarial network (GAN). The stochastic attribute of the latent variable makes it convenient to estimate the predictive uncertainty, serving as an auxiliary output to evaluate the reliability of model prediction.}
% for reliable saliency prediction, 
% we introduce the generative adversarial network (GAN) based models to identify the over-confident regions by sampling in the latent space. Specifically, we present the inferential generative adversarial network (iGAN). 
Different from the conventional GAN,
% based framework, 
which defines the distribution of the latent variable as fixed standard normal distribution $\mathcal{N}(0,\mathbf{I})$, the proposed \enquote{iGAN} infers the latent variable by gradient-based Markov Chain Monte Carlo (MCMC), namely Langevin dynamics, leading to an input-dependent latent variable model. We apply our proposed iGAN
% inferential generative adversarial network (iGAN) 
to both fully and weakly supervised salient object detection, and explain that iGAN within the transformer framework leads to both accurate and reliable salient object detection.
\end{abstract}

}

\maketitle

\IEEEdisplaynontitleabstractindextext

% \fi
\IEEEpeerreviewmaketitle

%==========================================================
\section{Introduction} 
\label{sec:intro}

\IEEEPARstart{V}isual salient object detection (SOD) \Rev{\cite{zhuge2023salient,scrn_sal,wei2020f3net,zhang2021_ucnet,fan2020bbs,jing2020weakly,imagesaliency,SOD_Survey_TPAMI_2021,Zhang_2018_CVPR,deepusps,Miao_2021_ACM_MM,xu2021locate,wu2022edn,yang2022biconnet} aims to localize and segment the regions of an image that attract human attention,
which is
% Given its foreground (salient region) segmentation setting, SOD is 
usually defined as a binary segmentation task.
% \NB{Prob mention pixelwise up front here so generalist readers can differentiate from typical detection}
% Binary segmentation tasks focus on producing segmentation maps indicating the foreground and background, which is usually class-agnostic. In this paper, we investigate in two widely studied binary segmentation tasks, namely salient object detection (SOD) and camouflaged object detection (COD). The former \cite{scrn_sal,wei2020f3net,ucnet_sal,jing2020weakly} aims to localize the region that attract human attention. The latter \cite{fan2020camouflaged,yunqiu_cod21,aixuan_cod_sod21} tries to find the camouflaged objects hiding in the environment.
Depending on whether unimodal data (\ie~RGB image) or multimodal data (\ie~RGB-D data) is used, the majority of salient object detection models can be roughly divided into
% There are two types of settings for static image based saliency detection, namely 
RGB image saliency detection \cite{wei2020f3net,scrn_sal,xu2021locate,wu2022edn,yang2022biconnet} and RGB-D image pair saliency detection \cite{fan2020bbs,zhang2021_ucnet}. The former
% RGB image saliency detection 
involves two variables, namely the RGB image $x$ and its corresponding ground truth saliency map $y$, while the extra depth data $d$ is involved in the latter, making it a multimodal learning task.}
% latter  involves three main variables, including the input image $x$, depth $d$ and ground truth saliency map $y$.}

\Rev{Given the one-to-one mapping formulation, and the backbone-dependent network structures, the main focus of conventional deep RGB image-based saliency detection models 
% aims to train a model to achieve a mapping from the input space (RGB image) to the output space (saliency map), which is usually defined as a single modal problem. The main focus of existing RGB image based saliency detection models 
is achieving structure-preserving
% detail-accurate 
prediction with effective high/low level feature aggregation. On the other hand, as a multimodal learning task, the basic assumption of RGB-D saliency detection is that the extra depth data can bring informative geometric information, which can be complementary to the appearance information from the RGB image. In this case, the main focus of existing RGB-D SOD models \cite{dmra_iccv19,Fu2020JLDCF,fan2020bbs,jing2021_complementary} is to extensively utilize the geometric information for effective multimodal learning.
% design multi-modal learning frameworks fo effective 
% RGB feature and depth feature fusion.
% or structure-aware loss functions.
% RGB-D image pair-based saliency detection involves three main variables, including the input image $x$, depth $d$ and ground truth saliency map $y$. The model is then trained to map $x$ and $d$ to the saliency map $y$. With extra geometric information from the depth data, RGB-D saliency detection is usually defined as a multi-modal learning problem, and the main focus of existing models \cite{dmra_iccv19,Fu2020JLDCF,fan2020bbs,jing2021_complementary} is to design multi-modal learning frameworks for effective RGB feature and depth feature fusion.
}

By thoroughly analyzing the existing saliency detection models, we observe three major issues, namely the less effective global context modeling abilities, the missing structure information issue, and the inconsistent depth distribution issue.

\begin{figure*}[tp]
%  \vspace{-5mm}
  \begin{center}
  \setlength\tabcolsep{2pt}
  \begin{tabular}{*{2}{p{0.075\textwidth}<{\centering}} | *{5}{p{0.075\textwidth}<{\centering}} | *{5}{p{0.075\textwidth}<{\centering}}}
  {\includegraphics[width=\linewidth]{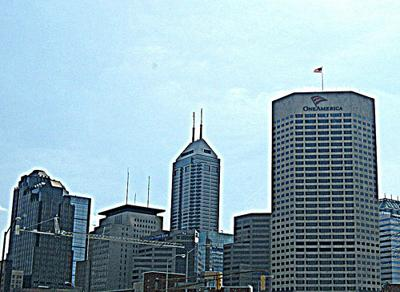}}
  &{\includegraphics[width=\linewidth]{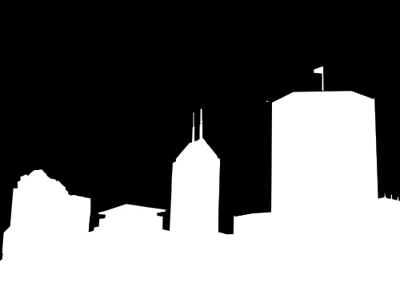}} 
  &{\includegraphics[width=\linewidth]{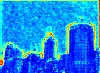}} 
  &{\includegraphics[width=\linewidth]{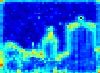}} 
  &{\includegraphics[width=\linewidth]{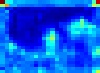}} 
  &{\includegraphics[width=\linewidth]{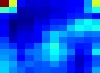}} 
  &{\includegraphics[width=\linewidth]{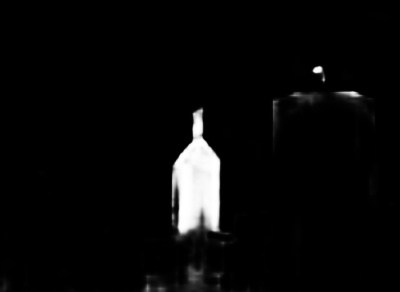}}  
  &{\includegraphics[width=\linewidth]{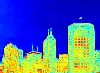}} 
  &{\includegraphics[width=\linewidth]{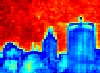}} 
  &{\includegraphics[width=\linewidth]{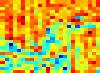}} 
  &{\includegraphics[width=\linewidth]{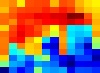}} 
  &{\includegraphics[width=\linewidth]{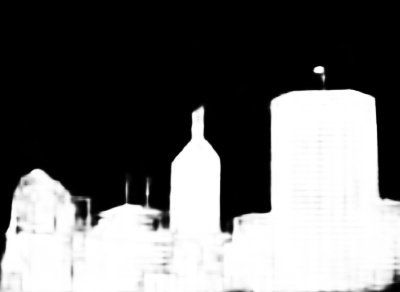}}  \\
  {\includegraphics[width=\linewidth]{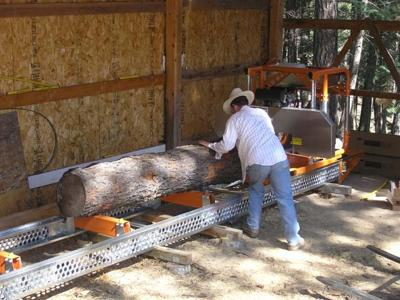}} 
  &{\includegraphics[width=\linewidth]{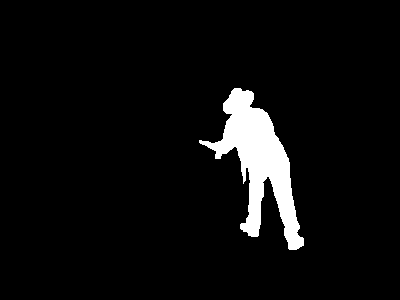}} 
  &{\includegraphics[width=\linewidth]{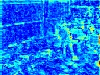}} 
  &{\includegraphics[width=\linewidth]{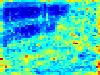}} 
  &{\includegraphics[width=\linewidth]{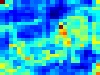}} 
  &{\includegraphics[width=\linewidth]{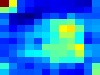}} 
  &{\includegraphics[width=\linewidth]{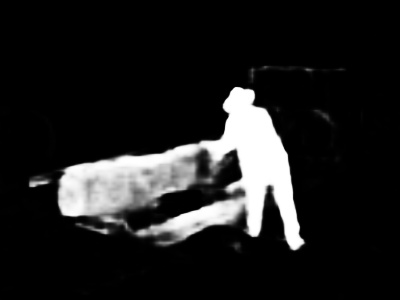}} 
  &{\includegraphics[width=\linewidth]{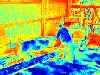}} 
  &{\includegraphics[width=\linewidth]{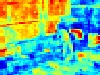}} 
  &{\includegraphics[width=\linewidth]{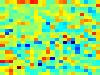}} 
  &{\includegraphics[width=\linewidth]{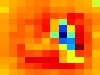}} 
  &{\includegraphics[width=\linewidth]{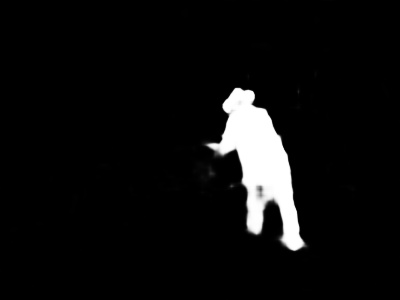}} \\
  \multicolumn{1}{c}{\footnotesize{Image}}&\multicolumn{1}{c}{\footnotesize{GT}}&\multicolumn{5}{c}{\footnotesize{CNN}}&\multicolumn{5}{c}{\footnotesize{Transformer}}\\ 
  % \begin{tabular}{c@{ }}
  % {\includegraphics[width=0.97\linewidth]{imgs/sup.pdf}} \\
  \end{tabular}
  \end{center}
%   \vspace{-15pt}
  \caption{\footnotesize{Visualizing the features of the ResNet50 backbone \cite{resnet} (\enquote{CNN}) and the transformer backbone (\enquote{Transformer})
%   before ($1^{st}$ row of each image) and 
  after fine-tuning them for SOD.
%   ($2^{nd}$ row of each image).
  }
  }
%   \caption{\footnotesize{Different levels of backbone features of the ResNet50 backbone \cite{resnet} (\enquote{CNN}) and the transformer backbone (\enquote{Transformer})
% %   before ($1^{st}$ row of each image) and 
%   after fine-tuning them for salient object detection.
% %   ($2^{nd}$ row of each image).
%   }
%   }
\label{fig:structure_comparison}
\end{figure*}

\begin{table*}[t!]
  \centering
  \footnotesize
  \renewcommand{\arraystretch}{0.9}
  \renewcommand{\tabcolsep}{3.9mm}
  \caption{\Rev{Details of the widely used RGB-D saliency datasets to explain the domain gap issue for RGB-D saliency detection.}
  }\label{tab:existing_rgbd_dataset}
  \begin{tabular}{c||r|r|c|c|r|r}
  \hline
     & & & & & & \\
    \textbf{Dataset} & \textbf{Year} & \textbf{Size} & \textbf{Type} & \textbf{Depth Source}&  \textbf{\#Train}& \textbf{\#Test} \\
  \hline
  \hline
    SSB \cite{niu2012leveraging} & 2012  &1,000  & Internet & Stereo cameras+ optical flow~\cite{liu2010sift}  & - & 1,000 \\
    NLPR \cite{peng2014rgbd} & 2014  &1,000  & Indoor/Outdoor&  Microsoft Kinect~\cite{zhang2012microsoft} & 700 & 300 \\
    DES \cite{cheng2014depth} & 2014 & 135  & Indoor& Microsoft Kinect~\cite{zhang2012microsoft} & - & 135 \\
    NJU2K\cite{NJU2000} & 2014 & 1,985  & Movie/Internet  & FujiW3 camera + Sum's optical flow~\cite{sun2010secrets}  & 1,500 & 485 \\
    LFSD \cite{li2014saliency}  & 2014 &80  & Indoor/Outdoor& Lytro Illum cameras~\cite{ng2005light} & - & 80 \\
    SIP \cite{sip_dataset} & 2020 & 929   & Person in outside & Huawei Mate10 & - & 929  \\
    % DUT \cite{dmra_iccv19} & 2019  & 1,200  & Indoor/Outdoor & Lytro2 camera+~\cite{tao2013depth} & 800 & 400 \\
    % ReDWeb-S~\cite{liu2020learning} & 2020 &  3179  & Indoor/Outdoor & Web stereo images & - & 3179  \\
  \hline
  \end{tabular}
\end{table*}

\begin{figure*}[htp]
%  \vspace{-5mm}
  \begin{center}
  \begin{tabular}{c@{ } c@{ } c@{ } c@{ } c@{ } c@{ } }
  {\includegraphics[width=0.16\linewidth]{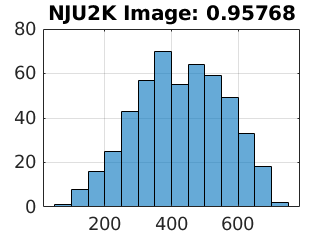}} &
  {\includegraphics[width=0.16\linewidth]{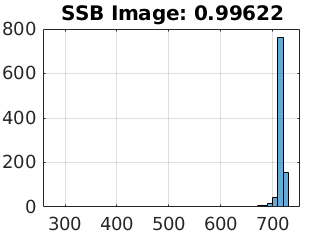}} &
  {\includegraphics[width=0.16\linewidth]{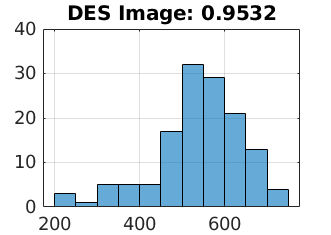}} &
  {\includegraphics[width=0.16\linewidth]{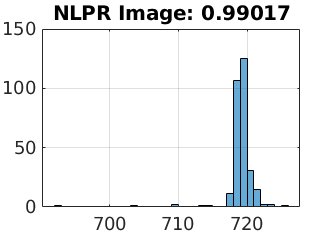}} &
  {\includegraphics[width=0.16\linewidth]{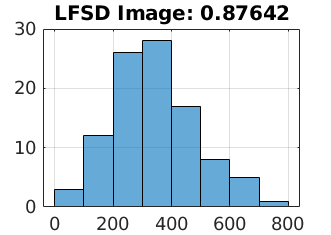}} &
  {\includegraphics[width=0.16\linewidth]{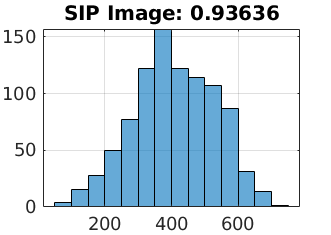}} \\
  \end{tabular}
  \end{center}
%   \vspace{-5pt}
  \caption{\footnotesize{Global contrast of depth from benchmark RGB-D SOD datasets, where the x-axis is the Chi-squared distance between salient foreground and non-salient background within the depth data, and the y-axis is the number of images.}
  }
\label{fig:depth_global_contrast}
\end{figure*}

\noindent\textbf{Less effective global context modeling:} Conventional CNN based saliency detection models usually consist of two main parts: 1) an encoder for feature extraction; and 2) a decoder for high/low feature aggregation, where the encoder is usually adopted from 
% \sout{the}\NB{an} 
an ImageNet pre-trained backbone network, \eg, VGG~\cite{vgg_network}, ResNet~\cite{resnet}. \Rev{In this way, the SOD models are mainly designed to obtain effective decoders for feature aggregation \cite{cpd_sal,scrn_sal,wei2020f3net}. We visualize the different levels of CNN and transformer backbone features of the SOD models in Fig.~\ref{fig:structure_comparison} and find that the former encodes less accurate global context than the latter, especially for the large salient foreground (the first row of Fig.~\ref{fig:structure_comparison}).
% less
% We claim that the sliding window based convolution operation within the CNN backbone limits its context modeling ability, especially the global context modeling ability, which is important for detecting large salient objects (
% see Fig.~\ref{fig:cnn_transformer_scales}). 
% We have thoroughly analyzed the failure cases of existing state-of-the-art saliency detection models, and observed that the less effective global context modeling caused by limited receptive fields is one of the major issue
% % \NB{issue} \sout{reasons} 
% that hinder the performance of deep saliency detection models.
}

\noindent\textbf{Missing structure information:} Conventional CNN backbones have gradually larger receptive fields
% \NB{s} 
with the deeper layers by using
% The usage of the
% \NB{the} 
stride or pooling operation,
% within the
% % \NB{the} 
% existing CNN backbone-based framework 
leading to extensive down-sampling
% of the backbone features 
as shown in Fig.~\ref{fig:structure_comparison} (\enquote{CNN}), where we show the different levels of backbone features of the ResNet50 \cite{resnet} after fine-tuning it for SOD. We observe missing structure information\footnote{\Rev{We define \enquote{structure information} as the detail-alignment of prediction with the input image.}} in both higher and lower level features,
% and lower level features, 
which makes effective high/low feature aggregation especially
% and structure-aware loss function
necessary for CNN backbone based framework to achieve structure-accurate predictions.
% for structure information recovery.
% recovery the missing structure information. Note that, 
However, once the information is lost, it will not be fully
% \NB{not be fully} \sout{never be} 
recovered.

% Conventional CNN backbones have gradually larger receptive fields
% % \NB{s} 
% with the deeper layers.
% % \NB{layers}\sout{ network}, 
% % where the deepest level of network has the largest receptive field. 
% But the larger receptive field is obtained with a loss of structure information as a sacrifice (see Fig.~\ref{fig:structure_comparison}).
% this is the main issue of these backbone networks for SOD.
% \NB{for salient object detection? dense prediction?}.
% The main issue of these backbone networks is that the larger receptive field is obtained with loss of structure information as a sacrifice.
% \NB{. This is}\sout{, and that's} 
% \sout{And t} \NB{T}
% This is the reason for incorporating the additional complicated decoders.
% \NB{incorporating an additional decoder,} \sout{the extra decoder part,} 
% which is used to recover the lost structure information by aggregating higher and lower level features. 
% Note that, once the information is lost, it will not be fully
% % \NB{not be fully} \sout{never be} 
% recovered.

% \noindent\textbf{Domain gap of depth data:}
\noindent\textbf{Inconsistent depth distribution:}
For RGB-D saliency detection, extra depth data is involved, \Rev{and the mainstream is then to effectively explore the complementary information of both modalities for effective multimodal learning. We observe that the depth data of RGB-D saliency detection can come from different sensors as shown in Table~\ref{tab:existing_rgbd_dataset}, thus the training data and the testing data can be treated as from different domains.
% For the conventional monocular depth estimation or other depth involved applications [][][], the parameters about the depth data is usually provided, thus the model is trained and tested on depth from the same sensors, \ie~close-domain training/testing. 
% Differently, depth for RGB-D saliency detection can be generated by differernt sensors,}
% , which serves as part of the input.
Further, we find that depth from different sensors has different contrast distributions, leading to inconsistent input distributions across the training and testing datasets.}
% leads to the domain gap issue of depth data.
We compute the global contrast
% and interior contrast 
of the depth data from different testing datasets and show the global depth contrast in Fig.~\ref{fig:depth_global_contrast}.
% where the x-axis is the corresponding contrast and the y-axis represents the number of images.
The global contrast measures the noticeability of the salient objects.
% while the interior contrast measures the consistency inside the same salient object, representing the intra-class consistency. 
To obtain the global contrast of the depth data, we first compute a $3H$ dimensional color histogram of both salient foreground and background of the depth data. Following~\cite{kim2015salient}, we obtain an $H\!=\!16$ dimensional histogram for the Red, Green, and Blue channel\footnote{For gray image, we obtain its color version by concatenating it channel-wise to obtain the three-channel color image.} of the RGB image respectively, and the histogram of the color image is then the concatenation of the above histograms. Then we adopt the Chi-squared distance to measure the global contrast between salient objects and background. We define the mean of the Chi-squared distance as the global depth contrast. Fig.~\ref{fig:depth_global_contrast} clearly shows that the global contrast of the salient foreground with the depth data varies across the testing datasets. 
Similarly, we obtain the RGB image global contrast and compute the global contrast difference of the RGB image and depth for each testing dataset, which is shown in the title of each figure in Fig.~\ref{fig:depth_global_contrast}. The various global contrast differences between RGB images and depth data further explain the different contributions of depth.

\noindent\textbf{Advantages of Transformer:} Researchers have found that the \enquote{Transformer}~\cite{transformer_nips}
% , which was first proposed in natural language processing (NLP), 
has great potential to solve
% \sout{in solving} \NB{to solve} 
the limited receptive field issue in vision tasks. The advantage of the \enquote{Transformer} lies in the use of self-attention to capture global contextual information to establish a long-range dependency.
% as shown in Fig.~\ref{fig:receptive_field_comparison}.
% , and it can extract more meaningful features.
Different from
% \sout{the} 
convolutional neural networks
% \NB{s} 
that focus on a small patch of the image,
% with a sliding-window-wise convolution operation, 
the transformer network \cite{transformer_nips} performs global context modeling with self-attention. Inspired by \cite{dpt_transformer,liu2021swin} and the accurate structure modeling ability of the transformer (see Fig.~\ref{fig:structure_comparison}), we conduct extensive research to explore
% \sout{on exploring} \NB{to explore} 
the contributions of the
% \NB{the} 
transformer for accurate salient object detection. Specifically, we design transformer based
% to achieve both accurate and reliable predictions.
% Firstly, we investigate transformer networks
% \NB{networks}
% encoder, transformer decoder, and multi-modal transformer 
% for accurate SOD with 
deterministic neural networks for SOD, and explain that the accurate structure modeling and the global context modeling abilities lead to its superior performance compared with the CNN based frameworks (see Table \ref{tab:fully_rgb_sod_expeiments}).

\noindent\textbf{Overcoming the over-confidence issue:} Although significant performance has been achieved with the transformer, we still observe \enquote{over-confidence} issue within the transformer based SOD models,
% (see Fig.~\ref{fig:overfitting_issue_cnn_transformer}), 
where the model tends to generate wrong prediction with high confidence, which is also defined as the model less-calibrated issue in~\cite{on_calibration}.
% we investigate transformer encoder, transformer decoder, and multi-modal transformer for accurate salient object detection by extensively exploring the global context modeling ability of transformer. 
% Secondly,
% considering the huge number of parameters 
% \NB{number of} parameter\NB{s} \sout{amount} 
% of the transformer compared with CNNs, \NB{but why do you connect parameter number to calibration?} 
% calibration 
% Due to the huge parameter numbers, both CNN and transformer based framework suffer greatly from the over-confidence issue, where the models intend to generate wrong predictions with high confidence, leading to over-confident predictions or poorly-calibrated model. To estimate the calibration degree of 
% to achieve
% , we analyze how to achieve 
% for reliable saliency prediction, 
We then present an inferential generative adversarial network (iGAN) to analyze the reliability degree \cite{on_calibration} of the transformer based framework.
% Specifically, we 
Different from the conventional generative adversarial network (GAN) \cite{gan_raw}
% based framework 
which defines the distribution of the latent variable as fixed standard normal distribution $\mathcal{N}(0,\mathbf{I})$, our proposed \enquote{iGAN} infers the latent variable by gradient based Markov Chain Monte Carlo (MCMC), namely Langevin dynamics \cite{mcmc_langevin}. The latent variable within iGAN is sampled directly from its true posterior distribution \cite{ABP}, leading to more informative latent space exploration.
% Lastly, w
We apply the proposed iGAN
% inferential generative adversarial network (iGAN) 
to both fully and weakly
% /weakly/semi-
supervised SOD, and explain that iGAN within the transformer framework leads to both accurate and reliable salient object detection, where the produced uncertainty maps \cite{kendall2017uncertainties} \Rev{can serve as an auxiliary output to explain the reliability of model predictions. Experimental results show that the proposed iGAN within the transformer backbone can fix the \enquote{less effective global context modeling} and \enquote{missing structure information} issues of conventional CNN backbone based framework, and the auxiliary uncertainty outputs can be used to explain model reliability.}

\noindent\Rev{\textbf{Fixing the depth domain gap issue:} Although consistent performance improvement is achieved with the transformer backbone based framework, we still observe the depth domain inconsistency issue, where the testing datasets with depth different from the training datasets achieve a marginal performance improvement, \eg~LFSD \cite{li2014saliency} in Fig.~\ref{fig:depth_global_contrast}.
% still exists within the transformer backbone based framework.
% is limited.
% For the \enquote{inconsistent depth distribution} issue, 
To fix it, we present an \enquote{auxiliary depth} module and perform self-supervised depth estimation. The deep hybrid model structure~\cite{cao2022deep} via depth reconstruction, \ie~RGB-D saliency detection as conditional generation and depth estimation as marginal density estimation, works effectively when a depth domain gap exists.}

Our main contributions can be summarized as: \textbf{1)} We extensively explore the contributions of transformer networks~\cite{transformer_nips,liu2021swin,dpt_transformer,dosovitskiy_ViT_ICLR_2021} for \textbf{accurate} salient object detection and explain that the effective structure and global context modeling abilities lead to the superior performance of the transformer-based saliency detection network; \textbf{2)} We present an inferential generative adversarial network (iGAN) to effectively measure the reliability degree of the transformer-based SOD network, leading to \textbf{reliable} saliency prediction; \textbf{3)} We apply iGAN to fully and weakly supervised salient object detection to extensively explore the proposed new generative model within the transformer framework.

\vspace{-4mm}
\section{Related Work}
% In this section, we will introduce the existing salient object detection models, recent work on transformer network, generative models and weakly supervised
% % . We will also introduce the main research focus of weakly-supervised and semi-supervised 
% segmentation models.
% and their application on dense prediction tasks.
% and confidence-aware learning.

% \noindent\textbf{RGB Image based Fully Supervised Salient Object Detection:}
\noindent\textbf{Salient Object Detection:}
% SOD aims to detect and segment the objects that attract human attention, which is d
Driven by visual attention \cite{itti1998model},
% (see Fig.~\ref{fig:saliency_timeline}). 
% In this way, 
salient objects are defined as objects that have strong contrast~\cite{global_contrast}.
% and
% Salient objects usually have strong global contrast, therefore the 
Early works usually utilized this prior for saliency related feature extraction.
% Wei \etal \cite{wei2012geodesic} use the contrast between the pixel and the background to generate a saliency score. Yang \etal \cite{Manifold-Ranking:CVPR-2013} sort the similarities between the superpixels of the image and the foreground and background to generate a saliency proposal. With the development of deep neural networks\cite{resnet}, convolutional neural networks gradually have a huge impact in the field of salient object detection \cite{scrn_sal,wei2020f3net,qin2019basnet,qin2020u2,ucnet_sal}. 
% Currently, the mainstream of saliency detection models are deep convolutional neural network based,
% SOD models are the main stream, 
Deep SOD models usually take the pretrained backbone networks~\cite{vgg_network,resnet} as an encoder with
% within a 
UNet~\cite{ronneberger_unet_2015}
% or HED-style~\cite{xie_hed_iccv_2015} network 
structure, where
% , and most of the effort is then to design
% As discussed above, 
% Existing fully supervised RGB image based salient object detection models \cite{cpd_sal,nldf_sal,picanet,scrn_sal,wei2020f3net,wang2020progressive} mainly focus on designing 
effective decoders are designed to achieve high-low level feature aggregation~\cite{cpd_sal,nldf_sal,picanet,scrn_sal,wei2020f3net,wang2020progressive, hou_DSS_cvpr_2017, wang_iccv_2017_stagewise, zhao_eccv_2020_suppress,chen2020global,zhang_saliency_hierarchy_ECCV_2022,qin2019basnet,pang_MINet_CVPR_2020}.
% Hou \etal~\cite{hou_DSS_cvpr_2017} proposed to use a deep supervised learning strategy to perform multi-level supervision of features.
Among them, Wu \etal~\cite{scrn_sal} proposed a \enquote{stacked cross refinement network}, and used the interaction between the edge module and the detection module to optimize the two tasks at the same time. Wei \etal~\cite{wei2020f3net} introduced an adaptive selection of complementary information when aggregating multi-scale features with a
% , and presents a 
structure-aware loss function.
% to , which solves the problem of differences in multi-scale features during fusion. 
% Qin \etal~\cite{qin2019basnet} used deep supervision for the encoder-decoder network and designed an extra residual refinement module with high-quality boundary supervision to obtain a more refined saliency map.
%use multiple supervisions and high-quality boundaries to guide the encoder, decoder and residual refinement module to gradually optimize the saliency prediction to obtain a more refined segmentation.
% \cite{pang_MINet_CVPR_2020} integrates the information of adjacent layers, and integrates multi-scale information to retain the internal consistency of each category (salient foreground or non-salient background).
Tang \etal~\cite{tang2021disentangled} modeled the two tasks of discriminating salient regions and identifying accurate edges independently and solved the limitations of low-resolution SOD by using low-resolution images to delineate salient regions and using high-resolution to refine salient regions. Meanwhile, edge detection~\cite{xie_hed_iccv_2015, liu_rcf_cvpr_2017} is likewisely used as a piece of auxiliary information~\cite{basnet_sal, zhao2019EGNet} to improve the performance of SOD.
And different attention mechanisms such as spatial and channel attention~\cite{dmra_iccv19, zhang_progressive_attention_cvpr_2018} or pixel-wise contextual attention~\cite{picanet} are also used to learn more discriminative features.
Unlike the mainstream design refinement prediction networks, Zhang \etal~\cite{zhang2021auto} proposed an automatic consolidation of multi-level features based on neural architecture search for flexible integration of information at different scales.

% \noindent\textbf{RGB Image based Weakly Supervised Salient Object Detection:}
% The weakly supervised saliency models \cite{imagesaliency,Guanbin_weaksalAAAI,deepusps,jing2020weakly,structure_consistency_scribble,xin2018c2s} learn saliency from easy-to-obtain weak labels, including image-level labels \cite{imagesaliency,Guanbin_weaksalAAAI}, noisy labels \cite{deepusps,Zhang_2018_CVPR,Zhang_2017_ICCV} or partial scribble labels \cite{jing2020weakly, structure_consistency_scribble}. Due to the limited structure information in the weak annotations, the performance of existing weakly supervised models is still far from satisfactory, and the main focus of existing weakly supervised salient object detection models is to recover the structure information by designing the pair-wise constraints related regularizer. 

% \noindent\textbf{RGB-D Image Pair based Fully Supervised Salient Object Detection:} 
With extra depth information, RGB-D pair based SOD models \cite{zhang2021_ucnet,fan2020bbs,Fu2020JLDCF,dmra_iccv19,han_cnnbased_fusion_2017,lee_Superpixel_RGBD_ECCV_2022} mainly focus on exploring the complementary information between the RGB image and the depth data for effective multi-modal learning.
% The former provides appearance information of a scene, while the latter introduces geometric information. 
Depending on how information from these two modalities is fused, existing RGB-D SOD models can be roughly divided into three categories: early-fusion models \cite{qu2017rgbd,zhang2021_ucnet}, late-fusion models \cite{wang2019adaptive,han2017cnns,A2dele_cvpr2020} and cross-level fusion models \cite{dmra_iccv19,chen2018progressively,chen2019multi,chen2019three,zhao2019Contrast,ssf_rgbd,self_attention_rgbd,fan2020bbs,ji2020accurate,HDFNet-ECCV2020,zhang2020bilateral,cmms_rgbd,Li_2020_CMWNet,jing2021_complementary}. The early-fusion models fuse RGB image and depth data at the input layer, forming a four-channel feature map.
% The first solution directly concatenates the RGB image with its depth,
% % information, forming a four-channel input,
% while
% , and feed it to the network to obtain both the appearance information and geometric information.
% \noindent\textbf{Early fusion models:} These models directly concatenates the RGB image and the depth data channelwise to obtain a four-channel input data. \cite{DANet,jing2020uc,Fu2020JLDCF}
% \cite{CoNet} define depth as prior, and train an extra depth estimation branch to achieve inference without depth as input.
% \noindent\textbf{Late fusion models:} Instead of fusing in the input layer, t
The late fusion models treat each mode (RGB and depth) separately, and then saliency fusion is achieved at the output layer.
% and two different networks with RGB and depth as inputs are trained to obtain two different predictions for each input RGB-D image pair. Then they fuse the above two predictions at the end of the network.
% \cite{A2dele_cvpr2020}
% \noindent\textbf{Cross-level fusion models:} 
The cross-level fusion models gradually fuse features of RGB and depth~\cite{HDFNet-ECCV2020,Li_2020_CMWNet,fan2020bbs,cmms_rgbd,Luo2020CascadeGN,chen2020eccv,ATSA,self_attention_rgbd,ssf_rgbd,li2021hierarchical,fu2021siamese,jing2021_complementary}, which is the main stream for RGB-D SOD.

\noindent\textbf{Vision Transformer and Its Applications:}
The transformer network~\cite{transformer_nips}
% which uses multi-head self-attention modules with positional encoding to capture long-range dependencies between every two elements in an input sequence, has achieved great success in natural language processing (NLP). 
% In addition, both the encoder and decoder layers have feed forward multilayer perceptron (MLP) blocks with residual connections and layer normalization steps for additional processing of the outputs.
% This breakthrough 
has sparked great interests in the computer vision community to adapt these models for vision tasks such as object detection~\cite{carion_DETR_ECCV_2020, zhu_deformableDETR_ICLR_2021, dai_UPDETR_CVPR_2021, wang_PVT_2021,zhang_metaDETR_arxiv_2021}, object tracking~\cite{xu_TransCenterTracking_2021,yan_SpatialTemporalTransformerTrackingv2_2021}, pose estimation~\cite{mao_TFPose_2021}, optical flow~\cite{Jiang_GMAFlow_2021} \etcn
% The dense prediction tasks aim to perform pixel-level classification or regression on the feature map. In recent years, dense prediction tasks usually usefully convolutional neural networks (FCNs), which adopt convolution and sub-sampling with different feature attention or enhancement methods as their fundamental elements in order to learn multi-scale representations.
Inspired by the success of 
% \NB{the}
the Vision Transformer (ViT)~\cite{dosovitskiy_ViT_ICLR_2021} in image classification which splits the input image into a sequence of patches and feeds them to a standard Transformer encoder,
some works extend transformers for
% such classification model as a backbone for 
dense prediction tasks, \eg, semantic segmentation or depth estimation.
SETR~\cite{zheng_SETR_2020} and PVT~\cite{wang_PVT_2021} use several convolutional layers as the decoder to upsample feature maps and get the dense prediction with the input image size.
% PVT \cite{wang_PVT_2021} is based on the design of ViT, while it
% % also maintaining the global receptive field. This method
% introduces a progressive shrinking pyramid to reduce the sequence length of the transformer with increased network depths,
% % when the network depth increases, 
% thereby it can significantly reduce the computational cost.
% the amount of calculation.
% adopts the U-shape structure 
DPT~\cite{dpt_transformer} uses ViT~\cite{dosovitskiy_ViT_ICLR_2021} as an encoder to extract features from different spatial resolutions of the initial embedding. 
% \sout{And} \NB{Then} 
% Then a decoder assembles the set of tokens into grid representations at each resolution.
% Such feature representations are progressively fused into the final dense prediction module. 
Liu \etal~\cite{liu2021swin} presented the Swin Transformer, a hierarchical transformer with a shifted windowing scheme to achieve an efficient network for vision tasks.
Recently,~\cite{liu_ICCV_2021_VST, zhang2021learning_nips} introduce the
% \NB{the} 
transformer to saliency detection, achieving significant performance improvement.
% compared with the CNN based saliency models.
% Different from \cite{liu_ICCV_2021_VST}, which works on accurate saliency detection with a deterministic network, we present a generative model based framework for both accurate and reliable saliency detection.
% Different from \cite{zhang2021learning_nips}, which adopts an energy-based model~\cite{LeCun06atutorial} for modeling the prior distribution of the latent variable.
% We present an alternative solution by modeling the prior distribution of a generative adversarial network with Langevin Dynamics~\cite{mcmc_langevin}, leading to image conditioned prior distribution, named inferential GAN (iGAN). We also apply the proposed iGAN to weakly supervised SOD and explain the superior performance of iGAN within the transformer based framework, which has never been explored before.

\noindent\textbf{Generative Models and Their Applications:} There mainly exist two types of generative models, namely latent variable models~\cite{gan_raw,VAE_Kingma} and energy-based models~\cite{LeCun06atutorial}. The former usually involves an extra latent variable to model the predictive distribution, and the latter directly estimates the compatibility of the input and output variable with a designed energy function.
% The goal of dense generative models is to produce stochastic both accurate model prediction and meaningful uncertainty map representing model awareness of it's prediction.
% Two types of generative models have been widely studied, namely
% \NB{The} 
The variational auto-encoder (VAE)~\cite{cvae,VAE_Kingma} and generative adversarial network (GAN)~\cite{gan_raw} are two widely studied latent variable models. VAEs use an extra inference model to constrain the distribution of the latent variable, and GANs design a discriminator to distinguish real samples and the generated samples. VAEs have already been successfully applied to image segmentation~\cite{PHiSeg2019, probabilistic_unet} to produce stochastic predictions during testing. For saliency prediction,~\cite{SuperVAE_AAAI19} adopts a
% \NB{a} 
VAE for image background reconstruction and the residual of the raw image and the reconstructed background is then defined as the salient region(s). 
% emodel the image background, and separates salient objects from the background through reconstruction residuals.
Differently,~\cite{zhang2021_ucnet} designs conditional variational auto-encoder (CVAE) to model the subjective nature of saliency, where the latent variable is used to model the prediction variants. GAN-based methods can be divided into two categories, namely fully-supervised and semi-supervised settings. The former~\cite{groenendijk2020benefit,gan_maskerrcnn} uses the discriminator to distinguish model predictions from ground truth, while the latter~\cite{gan_semi_seg,hung2018adversarial} rely on the GAN to explore the contributions of unlabeled data. \cite{chen_IWGAN_arxiv_2021} introduces an inferential Wasserstein GAN model, which is a principled framework to fuse auto-encoders and Wasserstein GAN and jointly learns an encoder network and a generator network motivated by the iterative primal-dual optimization process. Differently, we infer the latent variable via Langevin dynamics~\cite{mcmc_langevin}, which
% is more convenient to implement and 
suffers no posterior collapse issue~\cite{Lagging_Inference_Networks}.

% \noindent\textbf{Weakly-/semi-supervised Segmentation Models:}
\noindent\textbf{Weakly Supervised Segmentation Models:}
Instead of label-consuming pixel-wise annotations, Weakly Supervised Segmentation (WSS) models are designed to explore the possibility of using weak labels, \eg, image tags~\cite{hsu122017weakly,Guanbin_weaksalAAAI,ahn2018learning,huang2018weakly,zhang2021learning}, bounding box~\cite{song2019box,dai2015boxsup,kulharia2020box2seg,lee2021bbam,tian2021boxinst}, scribble~\cite{lin2016scribblesup,vernaza2017learning,jing2020weakly,structure_consistency_scribble}, point~\cite{bearman2016s,chen2021seminar}, as supervision.
% In order to realize the application on large-scale datasets, WSS has been introduced in many computer vision areas such as semantic segmentation, instance segmentation and salient object detection. 
% The common pipeline of WSS is generating the high-quality initial seed regions to locate the object and then growing or refining the object regions \cite{}. 
The typical methods~\cite{dai2015boxsup,lin2016scribblesup,obukhov2019gated} usually consider the initial segmentation map produced by traditional unsupervised methods, such as MCG~\cite{arbelaez2014multiscale} and GrabCut~\cite{rother2004grabcut}, as the supervision to train the deep neural networks and then repeat the iterative process between refinement of the prediction and training of the network. However, the framework tends to introduce accumulated label noise in each step and the iteration is time-consuming. Zhang \etal~\cite{Zhang_2018_CVPR} proposed an end-to-end deep learning framework to predict the latent saliency map from multiple noisy saliency maps created by unsupervised handcrafted saliency methods and mitigated the influence of label noise by a specifically-designed noise modeling module. The idea is further extended~\cite{zhang2020learningeccv}
% \sout{promoted} \NB{extended}
to use a generative model~\cite{ABP} to model the label noise from a single noisy saliency map. 
% \NB{In addition to} \sout{Besides the} 
In addition to noise modeling strategies, some methods refine the segmentation map with structure-aware loss functions.
% refinement into the training of deep learning in the form of the loss function.
Yu \etal~\cite{structure_consistency_scribble} used partial cross-entropy loss to expand the scribble region to the whole object region and refine the segmentation with a
% by the aid of 
local saliency coherency loss.
% borrowing the idea of CRF and saliency structure consistency loss based on self-supervised learning. 
% Tian \etal~\cite{tian2021boxinst} directly refined the object from the coarse label by a deep neural network using the projection loss based on tightness prior to the bounding box and pair-wise affinity loss to assign proximal pixels to the same category.
% However, c
% Currently, m
Most WSS methods are based on CNNs. We explore the potential of transformers for weakly supervised segmentation, especially weakly supervised SOD with scribble supervision~\cite{jing2020weakly,structure_consistency_scribble}.

\section{Accurate and Reliable Salient Object Detection via Generative Transformer}

% \subsection{Necessity of Accurate and Reliable Model}
% \noindent\textbf{Accurate Saliency Model:} 
As a context-based task, SOD strongly relies on both local and global context, \Rev{where the former is necessary for identifying median size salient foreground, and the latter is essential for large salient object detection. As discussed in Sec.~\ref{sec:intro}, conventional CNN backbone-based frameworks~\cite{cpd_sal,scrn_sal} are effective in modeling the local context, and their performance deteriorates for salient objects that expand to a larger region
% , while the limited receptive field makes them less effective in modeling the global context 
(see Fig.~\ref{fig:structure_comparison}).
% , leading to less accurate prediction for larger salient objects. 
The self-attention model within the transformer framework~\cite{transformer_nips} enables it to achieve long-range dependency modeling with global context exploration, which is desirable for accurate salient object detection.
% , thus it can obtain
% % leading to section
% effective global context modeling, which in turn is beneficial for accurate SOD.
}

\Rev{Let's define our training dataset as $D=\{x_i,y_i\}_{i=1}^N$ of size $N$, where $x_i$ and $y_i$ are the input RGB image (or RGB-D image pair for RGB-D saliency detection) and the corresponding ground truth saliency map, and $i$ indexes the samples, which is omitted.
With the transformer backbone (we use $\theta$ to represent its parameters), given any testing sample $x^*$ with ground truth $y^*$, we define its joint distribution as $p(x^*,y^*,\theta|D)=p(y^*|x^*,\theta)p(\theta|D)p(x^*|D)$,
% \begin{equation}
% \label{deep_hybrid_model}
%     \begin{aligned}
%     &p(x^*,y^*,\theta|D)\\
%     &=\frac{p(x^*,y^*,\theta,D)}{p(D)}\\
%     &=\frac{p(y^*|x^*,\theta,D)p(x^*,\theta,D)}{p(D)}\\
%     % &=\frac{p(y^*|x^*,\theta)p(x^*|\theta,D)p(\theta,D)}{p(D)}\\
%     &=\frac{p(y^*|x^*,\theta)p(x^*|D)p(\theta|D)p(D)}{p(D)}\\
%     &=p(y^*|x^*,\theta)p(\theta|D)p(x^*|D),
%     \end{aligned}
% \end{equation}
% \begin{equation}
% \label{deep_hybrid_model}
%     \begin{aligned}
%     &p(x^*,y^*,\theta|D)\\
%     &=\frac{p(x^*,y^*,\theta,D)}{p(D)}\\
%     &=\frac{p(y^*|x^*,\theta,D)p(x^*,\theta,D)}{p(D)}\\
%     % &=\frac{p(y^*|x^*,\theta)p(x^*|\theta,D)p(\theta,D)}{p(D)}\\
%     &=\frac{p(y^*|x^*,\theta)p(x^*|D)p(\theta|D)p(D)}{p(D)}\\
%     &=p(y^*|x^*,\theta)p(\theta|D)p(x^*|D),
%     \end{aligned}
% \end{equation}
where $p(y^*|x^*,\theta)$ represents the predictive distribution or the inherent randomness given $\theta$ as the oracle~\cite{kendall2017uncertainties}. $p(\theta|D)$ explains the ambiguity of the model $\theta$ given the provided training dataset $D$, and $p(x^*|D)$ measures the discrepancy between $x^*$ and the training dataset $D$.}

\Rev{With the global context modeling ability of transformer, the $p(\theta|D)$ term
% in Eq.~\eqref{deep_hybrid_model} 
can be modeled more effectively compared with the CNN frameworks. However, there exists no solution in the transformer framework to model the predictive distribution $p(y^*|x^*,\theta)$. Further, the training/testing discrepancy is not mentioned either, thus it is inconvenient to evaluate the domain gap caused by $p(x^*|D)$. To fix the above-mentioned issues, we introduce a latent variable model with an extra latent variable $z$ involved to model the inherent data noise. Further, we find one of the main domain gap for RGB-D saliency detection lies in the inconsistent depth data as shown in Table~\ref{tab:existing_rgbd_dataset} and Fig.~\ref{fig:depth_global_contrast}. We then introduce an auxiliary depth module, achieving self-supervised learning, with which it's more convenient to evaluate the training/testing discrepancy. Specifically,
% given any testing sample $x^*$ with ground truth $y^*$, their joint distribution based on the latent variable model with latent variable $z$ is defined as:
with extra latent variable $z$, the joint distribution of the testing sample $x^*$
% $p(x^*,y^*,\theta|D)$
% in Eq.~\eqref{deep_hybrid_model} 
can be rewritten as: 
\begin{equation}
\label{deep_hybrid_model_latent_variable}
    \begin{aligned}
    &p(x^*,y^*,\theta,z|D)\\
    &=\frac{p(x^*,y^*,\theta,z,D)}{p(D)}\\
    % &=\frac{p(y^*|x^*,\theta,z,D)p(x^*,\theta,z,D)}{p(D)}\\
    &=\frac{p(y^*|x^*,\theta,z)p(x^*|\theta,z,D)p(\theta,z,D)}{p(D)}\\
    % &=\frac{p(y^*|x^*,\theta,z)p(x^*|z,D)p(\theta,z|D)p(D)}{p(D)}\\
    % &=p(y^*|x^*,\theta,z)p(\theta|D)p(z|D)\frac{p(z|x^*,D)p(x^*|D)}{p(z|D)}\\
    &=p(y^*|x^*,\theta,z)p(\theta|D)p(z|x^*,D)p(x^*|D).
    \end{aligned}
\end{equation}
% \begin{equation}
% \label{deep_hybrid_model_latent_variable}
%     \begin{aligned}
%     &p(x^*,y^*,\theta,z|D)\\
%     &=\frac{p(x^*,y^*,\theta,z,D)}{p(D)}\\
%     &=\frac{p(y^*|x^*,\theta,z,D)p(x^*,\theta,z,D)}{p(D)}\\
%     &=\frac{p(y^*|x^*,\theta,z)p(x^*|\theta,z,D)p(\theta,z,D)}{p(D)}\\
%     &=\frac{p(y^*|x^*,\theta,z)p(x^*|z,D)p(\theta,z|D)p(D)}{p(D)}\\
%     &=p(y^*|x^*,\theta,z)p(\theta|D)p(z|D)\frac{p(z|x^*,D)p(x^*|D)}{p(z|D)}\\
%     &=p(y^*|x^*,\theta,z)p(\theta|D)p(z|x^*,D)p(x^*|D).
%     \end{aligned}
% \end{equation}
}

\Rev{
% Compare Eq.~\eqref{deep_hybrid_model_latent_variable} with Eq.~\eqref{deep_hybrid_model}, we find that t
The extra latent variable in Eq.~\eqref{deep_hybrid_model_latent_variable} makes it convenient to estimate $p(y^*|x^*,\theta,z)$, where the latent variable $z$ can be sampled from $p(z|x^*,D)$ during testing, modeling the discrepancy between training and testing sample. Specifically, for RGB-D saliency detection, the $p(x^*|D)$ term is modeled directly following self-supervised learning with an auxiliary depth module, which can also be explained as an auto-encoder framework. In the following, we will first introduce a transformer for accurate saliency detection (Sec.~\ref{subsec:transformer_accurate_prediction}) for effective model parameter estimation ($p(\theta|D)$). We present the auxiliary depth module in Sec.~\ref{subsec_auxiliary_depth_estimation}, achieving an auto-encoder framework to evaluate the domain gap between training and testing samples, or the $p(x^*|D)$ term. We will introduce the latent variable model in Sec.~\ref{subsec:generative_saliency_model} to achieve modeling of $p(z|x^*,D)$, with which it's convenient to model the inherent randomness of model prediction $p(y^*|x^*,\theta,z)$. 
We present the objective function in Sec.~\ref{subsec:objective_function}.}

\subsection{Transformer for Accurate Saliency Detection}
\label{subsec:transformer_accurate_prediction}
% With
% % \sout{the} 
% global context modeling and accurate structure modeling ability, we claim that the transformer backbone \cite{liu2021swin} is especially effective in producing structure-accurate saliency predictions. We then design a transformer encoder based network
% % with a
% % \sout{very} \NB{a} 
% % simple decoder, 
% % aiming to achieve 
% for accurate salient object detection.
% For fair comparison, we also design the CNN backbone-based framework with the same decoder, and show performance in Table \ref{tab:fully_rgb_sod_expeiments} and Fig.~\ref{fig:viaulization_structure_aware_loss_effects}.
% , leading to less contribution from the widely used strucure-aware loss function \cite{wei2020f3net}. 
% In this paper, we perform experiments with both CNN backbone and the transformer backbone, and conduct experiments with structure-aware loss and the conventional binary cross-entropy loss function. The performance is shown in Table \ref{tab:fully_rgb_sod_expeiments}.
% We further show the produced saliency maps of both backbone with and without structure-aware loss function in Fig.~\ref{fig:viaulization_structure_aware_loss_effects}.

% \noindent\textbf{Network Details:}
% \subsection{Transformer Encoder}
The straightforward solution of
% \NB{for} \sout{of} 
using a
% \NB{a} 
transformer is to
% \NB{to} 
replace the CNN backbone with a transformer backbone, leading to the
% \NB{the} 
\enquote{transformer encoder}. We take the Swin transformer~\cite{liu2021swin} as our transformer encoder, which takes the embedded image features as input and produces a list of feature maps $f_{\theta_{1}}(x)=\{t_l\}_{l=1}^4$ of channel size 128, 256, 512 and 1024 respectively, representing different levels of features. Different from \cite{dosovitskiy_ViT_ICLR_2021,dpt_transformer} that use fixed tokenization,
the Swin transformer~\cite{liu2021swin} is a hierarchical transformer structure whose representation is computed with self-attention in shifted non-overlapped windows, thus it
% It uses shifted windows to build cross-window connection, thus
enables even larger receptive field modeling.
%%% The concat decoder version, channel size is 32
Given the \enquote{transformer encoder},
% $f_{\theta_{1}}(x)=\{t_l\}_{l=1}^4$, to obtain pixel-wise prediction, 
we design a simple \enquote{convolution decoder} to achieve high/low level feature aggregation. Specifically, we first feed each backbone feature $t_l$ to a simple convolutional block and obtain the new backbone feature $\{t'_l\}_{l=1}^4$ of the same channel size $C=32$. 
Such channel reduction operation aims to further enhance context modeling and reduce the huge memory requirement. Our final saliency map $s=f_{\theta_{2}}(\{t'_l\}_{l=1}^4)$ is then obtained via a decoder parameterized by $\theta_{2}$.
% \begin{equation}
%     \label{decoder_structure}
%     s=f_{\theta_{2}}(\{t'_l\}_{l=1}^4)
%     % s=f_{\theta_{msd}}(f_{\theta_{rcab}}[(\{t'_l\}_{l=1}^4)]),
% \end{equation}

\Rev{The detailed structure of the decoder can be formulated as $s=f_{\theta_\text{msd}}(f_{\theta_\text{rcab}}[(\{t'_l\}_{l=1}^4)])$, where $[\cdot]$ denotes the channel-wise concatenation operation, $f_{\theta_\text{rcab}}$ is the residual channel attention block~\cite{rca_eccv}, $f_{\theta_\text{msd}}$ is the multi-scale dilated convolutional block~\cite{denseaspp} to obtain a one-channel saliency map. Note that, $\theta=\{\theta_1,\theta_2\}$ indicates the entire parameters of our salient object detection network.}
$f_\theta(x)$ can directly produce the saliency map for RGB image $x$. For RGB-D saliency detection, we perform early fusion by simply concatenating the RGB image and depth data at the input layer, and feeding it to a $3\times3$ convolutional layer to generate a new input tensor $x'$ with a channel size of 3, which is then fed to the saliency generator $\theta$. With the global context modeling ability, the regression ability of the transformer is proven better than CNN frameworks, leading to better $\theta$ estimation given the same training dataset $D$ with less ambiguity/uncertainty.
% \subsection{Transformer Decoder}
% The goal of transformer encoder is to achieve effective feature extraction with both local and global context modeling. Based on the transformer encoder, we can use CNN decoder or transformer decoder. We claim that both decoders are effective in high/low level feature aggregation. 

% \subsection{Multi-modal Transformer}
% In addition to transformer encoder and transformer decoder, given RGB-D image pair, we aim to build a multi-modal transformer for effective RGB-D feature fusion.

% \noindent\textbf{Auxiliary Depth Estimation}
% The existing RGB-D salient object detection models takes both RGB image and depth data as input at training and test time. We argue that the depth data at test time from different sensor from the training dataset may lead to inferior performance. In this paper, we prove that auxiliary depth estimation is an effective solution to achieve two main benefits: 1) incorporating depth for RGB-D saliency detection; 2) avoiding the performance decoration caused by depth from different sensors.

% \noindent\textbf{RGB-D Fusion via Transformer:}

% \YC{As I understand, y and s are both 2D matrix,  so are inter and union as * and + are element-wise operator, then how to obtain a scalar as in (7)?}
% \YC{Whether * and + are ordinary operator?}

% \subsubsection{Depth Fusion Analysis}

\subsection{Auxiliary Depth Module}
\label{subsec_auxiliary_depth_estimation}
% For the above experiments, we work on RGB saliency detection, and we then analyze how the transformer backbone works for RGB-D saliency detection. 
As we discussed before, the inconsistent depth data (see both Table~\ref{tab:existing_rgbd_dataset} and Fig.~\ref{fig:depth_global_contrast})
% contrast distribution
% depth domain gap 
may hinder the performance of existing RGB-D saliency detection models. We then propose an auxiliary depth module to solve the \enquote{distribution gap} issue within existing RGB-D SOD datasets. The auxiliary depth module with parameters $\theta_3$ has the same structure as our saliency decoder $\theta_2$, which takes the backbone feature $f_{\theta_{1}}(x')$
% \footnote{$x'$ is the early-fused RGB $x$ and depth $d$ feature.} 
as input, and outputs a one-channel depth map $d'=f_{\theta_3}(f_{\theta_{1}}(x'))$. 
% Therefore, the input of the entire network is the RGB, depth image pair through early fusion method, and the output is a saliency map and auxiliary depth output.
Within this framework, the final loss function has extra depth related loss: 
\begin{equation}
    \label{depth_loss}    \mathcal{L}_\text{depth}=\alpha(\beta*\mathcal{L}_\text{ssim}+(1-\beta)*\mathcal{L}_1),
\end{equation}
where $\alpha=0.1$ is used to control the contribution of the auxiliary depth module, and following the conventional setting, we set $\beta=0.85$ in this paper. $\mathcal{L}_{ssim}$ is the SSIM loss function~\cite{Godard_2017_CVPR} and $\mathcal{L}_1$ denotes the L1 loss.

\Rev{\textit{The Auto-encoder for joint distribution modeling:} Given a test sample $x^*$\footnote{we define $x^*$ as RGB image for RGB saliency detection and RGB-D image pair for RGB-D saliency detection.}, for a segmentation model with parameters $\theta$, the output distribution is defined as $p(y^*|x^*,\theta)$, and there is no way to model the marginal data distribution $p(x^*)$. For RGB saliency detection, the \enquote{domain gap} issue is not that significant. However, for RGB-D saliency detection, we observe differences between training depth data and testing depth data (see Table~\ref{tab:existing_rgbd_dataset} and Fig.~\ref{fig:depth_global_contrast}), making reliable $p(x^*)$ estimation necessary for RGB-D saliency detection to take into account the training/testing discrepancy. In our setting, with the auxiliary depth module, we obtain the modeling of $p(x^*)$ or $p(x^*|D)$ via conditional self-supervised learning, where the RGB image is the conditional variable.
% , where $D$ is the training dataset, 
Combining with the main task, \ie~salient object detection, we achieve the joint distribution modeling
% (see Eq.~\eqref{deep_hybrid_model}) 
instead of the conditional distribution modeling. The hybrid model~\cite{cao2022deep} is proven more
% hus our model is more 
robust to the depth distribution gap issue as discussed in Sec.~\ref{sec:intro}.} 
% \YC{Indeed, we take RGB and D as input and make prediction x, then the depth loss is defined correspondingly. What is the difference between the input depth and the depth supervision? Given this, the name depth estimation may not be very accurate. Another issue maybe will this module also works for RGB SOD?}

%=================================================
\subsection{Generative Model for Reliable Saliency Detection}
\label{subsec:generative_saliency_model}
% \subsection{Inferential GAN}
As deep neural networks can fit any random noise~\cite{on_calibration}, the deterministic CNN and transformer backbone based models have serious over-confidence issues, where the model could inaccurately assign a high probability to the wrong prediction.
% , leading to poorly calibrated models. 
% In Fig.~\ref{fig:overfitting_issue_cnn_transformer}, we visualize predictions of deterministic CNN and transformer backbone based models, which clearly explain the over-fitting issue of both models, and make wrong predictions with high confidence. 
To overcome this, a model which is aware of its prediction with reasonable predictive distribution modeling is desired. As discussed in Eq.~\eqref{deep_hybrid_model_latent_variable}, \Rev{a latent variable model makes it convenient to estimate predictive distribution $p(y^*|x^*,\theta,z)$, which can be defined as being Gaussian distribution via:
\begin{equation}
\label{predictive_distribution_definition}
    p(y^*|x^*,\theta,z) = \mathcal{N}(\mu(x^*,\theta,z),\sigma^2(x^*,\theta,z)),
\end{equation}
where the mean is $\mu(x^*,\theta,z)=\mathbb{E}_{z\sim p(z|x^*,D)}p(y^*|x^*,\theta,z)$ and the variance (uncertainty) is $\sigma^2(x^*,\theta,z)=\mathbb{E}_{z\sim p(z|x^*,D)}(p(y^*|x^*,\theta,z)-\mu(x^*,\theta,z))^2$. 
%   \YC{should these equations be $\mu(x^*,\theta,z)=\mathbb{E}_{z\sim  p(z|x^*,D)}z p(y^*|x^*,\theta,z)$ and $\sigma^2(x^*,\theta,z)=\mathbb{E}_{z\sim p(z|x^*,D)}(p(y^*|x^*,\theta,z)-\mu(x^*,\theta,z))^2$? To confirm?}
Eq.~\eqref{predictive_distribution_definition} presents a convenient solution for evaluating uncertainty from a latent variable model, where the randomness or uncertainty of model prediction is controlled by the latent variable $z$, making meaningful $z$ quite desirable for reliable uncertainty estimation. We will first introduce the existing latent variable models and adapt them to our task. We then analyze their advantages and limitations. Based on this, we present the proposed inferential generative adversarial net.}

% relieve the over-confidence issue for model reliability estimation, 
% one can use generative models \cite{gan_raw,VAE_Kingma,ABP} instead of deterministic models to produce stochastic predictions, thus predictive distribution can be estimated. 
% There mainly exist three types of latent variable models, namely Generative Adversarial Nets (GAN) \cite{gan_raw}, Variational Auto-encoder (VAE) \cite{VAE_Kingma} and Alternating back-propagation \cite{ABP}.

% \subsection{Subjective Modeling for Salient Object Detection}
% To model the subjective nature of saliency, one could simply change the deterministic prediction network to a stochastic prediction network via generative models, \ie~\cite{VAE_Kingma,gan_raw}.

\noindent\textbf{Conventional latent variable models}

\noindent\textit{1) Generative Adversarial Nets (GAN)~\cite{gan_raw}:}
Within the GAN-based framework, we design an extra fully convolutional discriminator $g_\beta$ following \cite{hung2018adversarial}, where $\beta$ is the parameter set of the discriminator.
% , and define it as $g_\beta(,.,)$, which is used to distinguish model prediction and the ground truth annotations.
% Similar to the CVAE based model, t
Two different modules (the saliency generator $f_\theta$ and the discriminator $g_\beta$ in our case) play the minimax game in GAN based framework:
\begin{equation}
\label{gan_loss}
\begin{aligned}
    \underset{f_\theta}{\min} \, \underset{g_\beta}{\max} \, V(g_\beta,f_\theta) &= E_{(x,y)\sim p_{data}(x,y)}[\log g_\beta(y|x)]\\ &+ E_{z\sim p(z)}[\log(1-g_\beta(f_\theta(x,z)))],
\end{aligned}
\end{equation}
where
% $G$ and $D$ are the generator model and discriminator model respectively, 
$p_{data}(x,y)$ is the joint distribution of training data, $p(z)$ is the prior distribution of the latent variable $z$, which is usually defined as $p(z)=\mathcal{N}(0,\mathbf{I})$.

In practice, we define the loss function for the generator as the sum of a reconstruction loss $\mathcal{L}_{\text{rec}}$, and an adversarial loss $\mathcal{L}_{\text{adv}}$,
% $\mathcal{L}_{\text{adv}} = \mathcal{L}_{ce}(g_\beta(f_\theta(x,z)),1)$,
which is $\mathcal{L}_{\text{gen}} = \mathcal{L}_{\text{rec}} + \lambda\mathcal{L}_{\text{adv}}$,
% \begin{equation}
%     \label{gan_loss_generator}
%     \mathcal{L}_{gan} = \mathcal{L}_{\text{rec}} + \lambda\mathcal{L}_{\text{adv}},
% \end{equation}
where the hyper-parameter $\lambda$ is tuned, and empirically we set $\lambda=0.1$ for stable training. For SOD, we define the reconstruction loss $\mathcal{L}_{\text{rec}}$ as the structure-aware loss as in Eq.~\eqref{structure_loss}, and the adversarial loss as cross-entropy loss: $\mathcal{L}_\text{adv}=\mathcal{L}_{\text{ce}}(g_\beta(f_\theta(x,z)),\mathbf{1})$,
% where $\mathbf{1}$ is a all-one matrix.
% \YC{Is this 1 a scalar or a matrix?} 
which fools the discriminator that the prediction is real, where $\mathcal{L}_{\text{ce}}$ is the binary cross-entropy loss, and $\mathbf{1}$ is an all-one matrix.
The discriminator $g_\beta$ is trained via the loss function: $\mathcal{L}_{\text{dis}}=\mathcal{L}_{\text{ce}}(g_\beta(f_\theta(x,z)),\mathbf{0})+\mathcal{L}_{\text{ce}}(g_\beta(y),\mathbf{1})$, which aims to correctly distinguish prediction and ground truth. Similarly, $\mathbf{0}$ is an all-zero matrix.
% where $\mathcal{L}_{ce}$ is the binary cross-entropy loss.
% Similar to CVAE, for each iteration of training, we randomly select an annotation $y^m$, which will be treated as $y$ for both generator and discriminator updating. 
In this way, the generator loss and the discriminator loss can be summarized as:
\begin{equation}
    \begin{aligned}
    \label{gan_losses}
        &\mathcal{L}_\text{gen} = \mathcal{L}_{\text{rec}} + \lambda\mathcal{L}_{\text{adv}},\\
        &\mathcal{L}_\text{dis}=\mathcal{L}_\text{ce}(g_\beta(f_\theta(x,z)),\mathbf{0})+\mathcal{L}_\text{ce}(g_\beta(y),\mathbf{1}).\\
    \end{aligned}
\end{equation}

\noindent\textit{2) Variational Auto-encoder (VAE)~\cite{VAE_Kingma}:} For dense prediction task with input variable $x$ and output variable $y$, we refer to conditional variational auto-encoder (CVAE) \cite{cvae} instead, where the input image $x$ is the conditional variable. As a conditional directed graph model, a conventional CVAE mainly contains two modules: a generator model $f_\theta(x)$, which is a saliency generator in this paper, to produce the task related predictions,
% , which generates prediction with image $x$ 
% and a latent variable $z$ as input, 
and an inference model $q_\theta(z|x,y)$, which infers the latent variable $z$ with image $x$ and annotation $y$ as input.
% include a generative process that generates data $x$ by the generative distribution $p_\theta(x|z)$ from a set of latent variable $z$ with a prior distribution $p_\theta(z)$.
Learning a CVAE framework involves approximation of the true posterior distribution of $z$ with an inference model $q_\theta(z|x,y)$, with the loss function as:
\begin{equation}
    \label{CVAE_loss}
    \begin{aligned}
    % \begin{split} 
     \mathcal{L}_\text{cvae} = \underbrace{\mathbb{E}_{h\sim q_\theta(z|x,y)}[-{\rm log} p_\theta(y|x,z)]}_{\mathcal{L}_{\text{rec}}}\\
     + {D_{KL}}(q_\theta(z|x,y)\parallel p_\theta(z|x)).\\
    % \end{split}
    \end{aligned}
\end{equation}
The first term is the reconstruction loss and the second is the Kullback-Leibler divergence of prior distribution $p_\theta(z|x)$ and posterior distribution $q_\theta(z|x,y)$, where both of them are usually parameterized by multi-layer perceptron (MLP).

\begin{figure}[htp]
%  \vspace{-5mm}
   \begin{center}
   \begin{tabular}{c@{ }}
   {\includegraphics[width=0.96\linewidth]{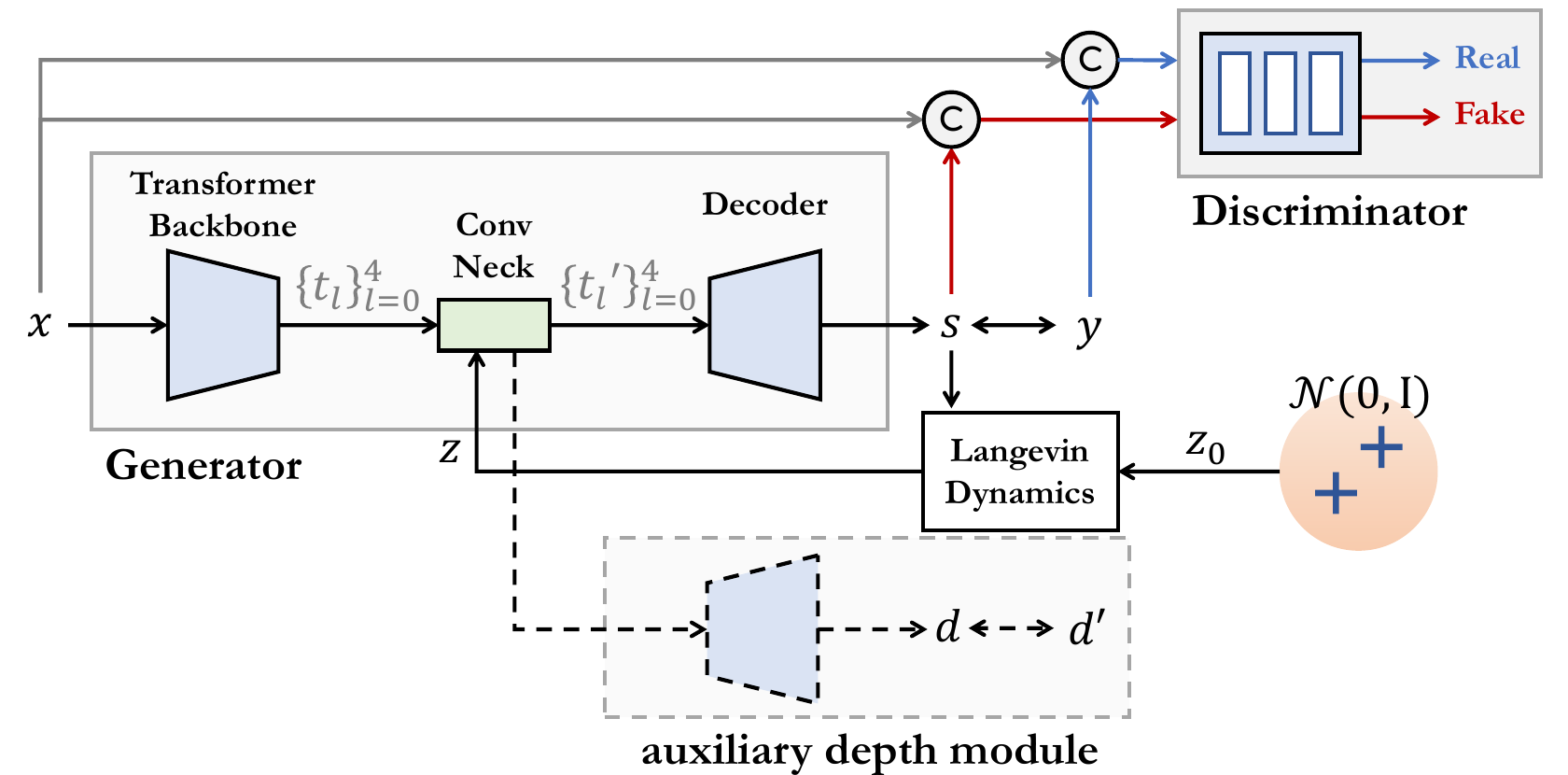}} \\
   \end{tabular}
   \end{center}
%   \vspace{-15pt}
   \caption{\footnotesize{Flowchart of our proposed inferential GAN (iGAN) (note that the \enquote{auxiliary depth} module is removed for RGB SOD). The \enquote{Generator} takes image $x$ and latent variable $z$ as input, and
%   is to 
   generates saliency map $s$, where the latent variable $z$ is updated via
%   with inputting noise sampled by 
   the Langevin dynamics based MCMC~\cite{mcmc_langevin}. The fully convolutional \enquote{Discriminator} is designed to distinguish prediction (fake) and ground truth (real). $d'$ is the predicted depth for RGB-D SOD.
%   \Jing{update figure, change h to z}.
%   \YC{To make it clear, better replace ``Generator'' with ``Transformer Generator'' in the figure.  In the figure, $\mathcal{N}(0,1)$ should be replaced with $\mathcal{N}(0, \mathbf{I})$.}
%   disciminator And use a extra adversarial loss by the \enquote{Discriminator}.
   }
   }
\label{fig:igan_flowchart}
\end{figure}

\noindent\textit{3) Alternating back-propagation (ABP):}
Alternating back-propagation \cite{ABP} updates the latent variable and network parameters in an EM manner. Given the network prediction with the current parameter set, it infers the latent variable by the Langevin dynamics based Markov Chain Monte Carlo (MCMC) \cite{mcmc_langevin}, which is called \enquote{Inferential back-propagation} \cite{ABP}. 
Given the updated latent variable $z$, the network parameter set is updated with gradient descent, which is called \enquote{Learning back-propagation}~\cite{ABP}.
Similar to the VAE \cite{VAE_Kingma} or CVAE \cite{cvae} frameworks, ABP intends to infer $z$ and learn the network parameter $\theta$ to minimize the reconstruction loss. Specifically, ABP \cite{ABP} samples $z$ directly from its posterior distribution with a gradient-based Monte Carlo method, namely Langevin Dynamics~\cite{mcmc_langevin}:
% to sample $z$:
% , which iterates:
\begin{equation}
\begin{aligned}
    z_{t+1}=z_{t}+ \frac{s_t^2}{2}\left[ \frac{\partial}{\partial z}\log p_\theta(y,z_{t}|x)\right]+s_t \mathcal{N}(0,\mathbf{I}),
    \label{langevin_dynamics}
\end{aligned}
\end{equation}
\noindent{where $z_0\sim\mathcal{N}(0,\mathbf{I})$, and the gradient term is defined as:}
\begin{equation}
\label{sigma_related_mcmc}
\frac{\partial}{\partial z}\log p_\theta(y,z|x) = \frac{1}{\sigma^2}(y-f_\theta(x,z))\frac{\partial}{\partial z}f_\theta(x,z) - z.
\end{equation}
$t$ is the time step for Langevin sampling, $s_t$ is the step size, $\sigma^2$ is variance of the inherent labeling noise. As no extra network is involved in the ABP based framework, the final loss function contains only the task-related loss.

\noindent\Rev{\textit{Analysis of the conventional latent variable models:}
As no inference model is included in GAN~\cite{gan_raw}, the latent variable $z$ within GAN is always sampled from the standard normal distribution $\mathcal{N}(0,\mathbf{I})$, which is less informative. For the CVAE based model~\cite{cvae,VAE_Kingma}, it samples the latent variable $z$ from the designed posterior distribution during training, and the distribution gap between the designed posterior and the true distribution leads to the posterior collapse issue \cite{Lagging_Inference_Networks}, where the latent variable is independent of the input image, leading to less representative latent space. For the ABP~\cite{ABP} based framework, although it samples from the true posterior distribution via Eq.~\eqref{langevin_dynamics}, the task related training is not changed, and our experimental results show that the deterministic performance is usually heavily influenced, especially for the conventional CNN backbone based frameworks.
% (see Table \ref{tab:reliable_rgb_sod}).
}

\noindent\textbf{The proposed inferential GAN}

% \YC{As iGAN is one of the major contribution, I prefer to have extended section 4.5 with more details especially how inferential is achieved.}
\Rev{For the VAE~\cite{VAE_Kingma,cvae} based framework, as parameters of the reconstruction model and the posterior net, are updated together, the convergence of the $D_{KL}$ term in Eq.~\eqref{CVAE_loss} may influence the convergence of the reconstruction part, which is also discussed in \cite{higgins2017betavae}. Although beta-VAE~\cite{higgins2017betavae} can slightly balance the convergence of the two parts, carefully picked hyper-parameters are needed. For the ABP~\cite{ABP} based framework, the basic assumption to achieve sampling from the true posterior distribution is that the time step should be large enough, and the step size should be infinitely small (see Eq.~\eqref{sigma_related_mcmc}). For the GAN~\cite{gan_raw} based framework, the less informative latent space makes it not ideal to be directly used to model the predictive distribution. However, its adversarial training strategy can usually lead to better model performance than the other two latent variable models.}
% Among the above three types of generative models, we find: 1) VAE based frameworlwe find the GAN based models usually lead to the best performance due to the higher level adversarial loss function $\mathcal{L}_{\text{adv}}$ within it. 
In this paper, we introduce inferential generative adversarial network (iGAN), a new generative model for SOD, where we infer the latent variable within the proposed framework instead of defining it as fixed $\mathcal{N}(0,\mathbf{I})$.
% Different from the conventional GAN-based framework which define the distribution of the latent variable as fixed standard normal distribution $\mathcal{N}(0,\mathbf{I})$, 
Specifically, the proposed iGAN infers the latent variable by gradient-based Markov Chain Monte Carlo (MCMC), namely Langevin dynamics \cite{mcmc_langevin} (see Fig.~\ref{fig:igan_flowchart}) following ABP~\cite{ABP}, leading to an image conditioned latent variable. Further, as adversarial training is applied, our new generative model can achieve reliable latent space exploration with fewer time steps.
% without sacrificing the deterministic performance.
% with limited time steps.

\begin{algorithm}[!ht]
\small
\caption{iGAN for fully supervised salient object detection}
\textbf{Input}: (1) Training images
% \footnote{We perform early fusion for RGB-D saliency detection model.} \YC{Interestingly, this footnote cannot be found in the paper.} 
$\{x_i\}_1^N$ with associated saliency maps $\{y_i\}_1^N$, where $i$ indexes images, and $N$ is the size of the training dataset (We perform early fusion for RGB-D saliency detection model).
(2) Maximal number of learning epochs $Ep$; (3) Numbers of Langevin steps for posterior $T$; (4) Langevin step sizes for posterior $s_t$ and variance of inherent labeling noise $\sigma^2$.\\
\textbf{Output}: 
Parameters $\theta$ for the generator and $\beta$ for the discriminator.
\begin{algorithmic}[1]
\State Initialize $\theta$ and $\beta$ 
\For{$ep \leftarrow  1$ to $Ep$}
\State Sample image-saliency pairs $\{(x_i,y_i)\}_i^N$
\State For each $(x_i,y_i)$, sample the prior $z_0^{i} \sim \mathcal{N}(0,\mathbf{I})$, and sample the posterior $z_t^{i}$ using $T$ Langevin steps in Eq.~\eqref{langevin_dynamics} with a step size $s_t$ and inherent noise $\sigma^2$. 
\State Update the transformer generator with model prediction $f_\theta(x,z_T^{i})$ using the generator loss function in Eq.~\eqref{gan_losses}.
\State Update the discriminator with loss function in Eq.~\eqref{dis_loss_conditioned}.
\EndFor
\end{algorithmic} \label{alg1}
% \vspace{5mm}
\end{algorithm}

Following the previous variable definitions, given the training example $(x,y)$, we intend to infer $z$ and learn the network parameters $\theta$ to minimize the reconstruction loss as well as a regularization term that corresponds to the prior on $z$.
Our iGAN based framework includes three main parts: a generator for task related predictions, a discriminator to distinguish the prediction and ground truth, and an inference model via Langevin dynamics~\cite{mcmc_langevin} to infer the latent variable with gradient based MCMC. Different from the isotropic Gaussian distribution assumption for the latent variable in GAN~\cite{gan_raw}, or the possible posterior issue~\cite{Lagging_Inference_Networks} within VAE~\cite{VAE_Kingma}, our latent variable $z$ is sampled directly from its real posterior distribution via gradient based MCMC following~\cite{ABP}. Further, we introduce extra adversarial loss and the fully convolutional discriminator, serving as a higher-order loss function for accurate deterministic predictions. Empirically, we set $s_t=0.1$ and $\sigma^2=0.3$ in Eq.~\eqref{langevin_dynamics} and Eq.~\eqref{sigma_related_mcmc}. During training, we sample $z_0$ from $\mathcal{N}(0,\mathbf{I})$, and update $z$ via Eq.~\eqref{langevin_dynamics} by running $T=5$ steps of Langevin sampling~\cite{mcmc_langevin}, and the final $z_T$ is then used to generate saliency prediction in our case. For testing, we can sample directly from the prior distribution $\mathcal{N}(0,\mathbf{I})$.

\noindent\textit{Network Details:}
The proposed iGAN can be applied to any deterministic saliency detection model, and we show the flowchart of the proposed iGAN for saliency detection in Fig.~\ref{fig:igan_flowchart}. Specifically, we first extend the latent variable $z$ to the same spatial size as the highest level backbone feature ($t_4$ in this paper). Then we concatenate $z$ with $t_4$ channel-wise and feed it to a $3\times3$ convolutional layer, which will serve as the new $t_4$ for saliency prediction.
% in Eq.~\eqref{decoder_structure}.
The discriminator contains four $3\times3$ convolutional layers following batch normalization and leakyReLU activation function with 64 channels, which takes the concatenation of image and model prediction (or ground truth) as input to estimate its pixel-wise realness. In this way, the discriminator loss in Eq.~\eqref{gan_losses} can be rewritten as:
\begin{equation}
    % \begin{aligned}
    \label{dis_loss_conditioned}
        \mathcal{L}_\text{dis}=\mathcal{L}_\text{ce}(g_\beta([f_\theta(x,z),x]),\textbf{0})+\mathcal{L}_\text{ce}(g_\beta([y,x]),\textbf{1}),
    % \end{aligned}
\end{equation}
where $[\cdot, \cdot]$ is the channel-wise concatenation operation. The training of the proposed iGAN is the same as the conventional GAN based models in Eq.~\eqref{gan_losses}, except that we have an extra inference model via MCMC \cite{mcmc_langevin}. We show the learning pipeline of iGAN in Algorithm \ref{alg1}.

\noindent\textit{Inferential GAN analysis:} Same as other generative models, iGAN aims to produce reliable uncertainty maps while keeping the deterministic performance unchanged. As the conventional GAN~\cite{gan_raw} has no inference step, the latent variable is independent of the input image $x$, leading to less informative uncertainty maps while sampling from the latent space at test time. Although VAE~\cite{VAE_Kingma} and ABP~\cite{ABP} can produce input-dependent latent space modeling, the possible posterior collapse~\cite{Lagging_Inference_Networks} issue within the former and the less accurate deterministic prediction of the latter limit their applications for SOD. With the proposed iGAN, we can achieve two main benefits: 1) extra inference step is included without increasing model parameters,
% as the VAE \cite{VAE_Kingma} and CVAE \cite{cvae} based frameworks, 
leading to an input-dependent latent variable; 2) with the adversarial loss function serving as a high-order similarity measure, iGAN can lead to more effective model learning compared with ABP~\cite{ABP}. For the former,
% Besides the reliable conParameter analysis: O
our iGAN is built upon GAN~\cite{gan_raw} and ABP~\cite{ABP}, and the fully convolutional discriminator introduces less than 1M extra parameters, which is comparable to both the alternative latent variable models and the deterministic models. For the latter, the adversarial training is proven effective in maintaining the deterministic performance compared with the alternative stochastic models.

% our iGAN contains less than 1M parameters. ABP involves no extra leanable parameters. For the alternative generative models, \ie~VAE~\cite{VAE_Kingma}, the extra posterior net with extra two fully connected layers to produce both mean and variance (which will be used to produce the latent variable $z$ via reparameterization trick) will introduce more parameters, which is around 1M in our experiments. The fully convolutional discriminator in GAN and our iGAN contains less than 1M parameters. ABP involves no extra leanable parameters. In this case, compared with the deterministic models, our iGAN framework introduce less than 1M parameters, which is comparable with the other latent variable models.

\subsection{Objective Function}
\label{subsec:objective_function}
\Rev{For RGB saliency detection, we remove the \enquote{auxiliary depth} module from Fig.~\ref{fig:igan_flowchart}.
% represents the entire training pipeline. 
The objective is shown in Eq.~\eqref{gan_losses}, where the reconstruction loss $\mathcal{L}_{\text{rec}}$ is chosen as the structure-aware loss from~\cite{wei2020f3net},}
% To train the model, the weighted structure-aware loss \cite{wei2020f3net} is adopted, 
which is the sum of the weighted binary cross-entropy loss and the weighted IOU loss:
\begin{equation}
\label{structure_loss}
   \mathcal{L}_{\text{rec}}^{\text{RGB}}=\omega(\mathcal{L}_\text{ce}(s,y)+\mathcal{L}_\text{iou}(s,y)),
\end{equation}
where $y$ is the ground truth saliency map, $\omega$ is the edge-aware weight, and is defined as $\omega=1+5*\left | (ap(y)-y)\right |$, with $ap(.)$ representing the average pooling operation. $\mathcal{L}_\text{ce}$ is the binary cross-entropy loss. $\mathcal{L}_\text{iou}$ is the weighted IOU loss~\cite{wei2020f3net}.

% , defined as:
\Rev{For RGB-D saliency detection, Fig.~\ref{fig:igan_flowchart}
represents the entire training pipeline.
As shown, we introduce an auxiliary depth module to model the joint distribution in Eq.~\eqref{deep_hybrid_model_latent_variable}.
% We show the framework of our RGB-D saliency detection model with the proposed iGAN framework in Fig.~\ref{fig:igan_flowchart}. 
Specifically, as an early fusion model, we concatenate the RGB image and depth data at the input level, which is then fed to a $3\times 3$ convolutional layer to produce a tensor with channel size 3, which is defined as the fused input $x'$. The proposed iGAN framework for RGB-D SOD takes $x'$ as input and produces both saliency map $s$
% (see Eq.~\eqref{decoder_structure}) 
and depth prediction $d'$.
% the
% is fed to the iGAN framework, and the auxiliary depth estimation module takes the 
% backbone feature $f_{\theta_{1}}(x')$ as input to produce a one-channel depth map (see sec.~\ref{subsec_auxiliary_depth_estimation}).
% In this case, t
The reconstruction loss for RGB-D saliency detection is defined as:
\begin{equation}
\mathcal{L}_{\text{rec}}^{\text{RGBD}}=\omega(\mathcal{L}_\text{ce}(s,y)+\mathcal{L}_\text{iou}(s,y))+\lambda\mathcal{L}_\text{depth},
\end{equation}
where $\lambda$ is used to balance the contribution of the auxiliary depth module, and empirically we set $\lambda=1$. The generator loss and discriminator loss are obtained following Eq.~\eqref{gan_losses}.
Note that for both RGB and RGB-D SOD models, the latent variable $z$ is updated via Langevin dynamics as shown in Eq.~\eqref{langevin_dynamics}.}

% \Rev{As shown in Fig.~\ref{fig:model_overview}, the proposed framework includes both the inferential GAN part for stochastic generation and the auxiliary depth estimation, achieving self-supervised learning for training/testing discrepancy modeling.
% where $y$ is the ground truth saliency map, $\omega$ is the edge-aware weight, which is defined as $\omega=1+5*\left | (ap(y)-y)\right |$, with $ap(.)$ representing the average pooling operation. $\mathcal{L}_{ce}$ is the binary cross-entropy loss. $\mathcal{L}_{iou}$ is the weighted IOU loss, defined as:
% \begin{equation}
% \label{weighted-iou-loss}
%   \mathcal{L}_{iou}=\text{mean}\left( 1-\frac{\omega*inter+1}{\omega*union - \omega*inter+1}\right),
% \end{equation}
% where $inter=s*y$ is the intersection of $s$ and $y$, and $union=s+y$ is the union of $s$ and $y$, and we obtain mean weighted IOU loss through $\text{mean}(.)$.
% }

\begin{table*}[t!]
  \centering
  \scriptsize
  \renewcommand{\arraystretch}{1.1}
  \renewcommand{\tabcolsep}{0.62mm}
  \caption{\footnotesize{Performance comparison with benchmark fully-supervised RGB SOD models.}}
  \begin{tabular}{l|cccc|cccc|cccc|cccc|cccc|cccc}
  \hline
% \toprule
  &\multicolumn{4}{c|}{DUTS \cite{imagesaliency}}&\multicolumn{4}{c|}{ECSSD \cite{yan2013hierarchical}}&\multicolumn{4}{c|}{DUT \cite{Manifold-Ranking:CVPR-2013}}&\multicolumn{4}{c|}{HKU-IS \cite{li2015visual}}&\multicolumn{4}{c|}{PASCAL-S \cite{pascal_s_dataset}}&\multicolumn{4}{c}{SOD \cite{sod_dataset}} \\
    Method & $S_{\alpha}\uparrow$&$F_{\beta}\uparrow$&$E_{\xi}\uparrow$&$\mathcal{M}\downarrow$& $S_{\alpha}\uparrow$&$F_{\beta}\uparrow$&$E_{\xi}\uparrow$&$\mathcal{M}\downarrow$& $S_{\alpha}\uparrow$&$F_{\beta}\uparrow$&$E_{\xi}\uparrow$&$\mathcal{M}\downarrow$& $S_{\alpha}\uparrow$&$F_{\beta}\uparrow$&$E_{\xi}\uparrow$&$\mathcal{M}\downarrow$& $S_{\alpha}\uparrow$&$F_{\beta}\uparrow$&$E_{\xi}\uparrow$&$\mathcal{M}\downarrow$& $S_{\alpha}\uparrow$&$F_{\beta}\uparrow$&$E_{\xi}\uparrow$&$\mathcal{M}\downarrow$ \\ \hline
%   $\text{CGAN}$ &.881 &.839 &.917 &.036 &.919 &.916 &.945 &.036 &.818 &.734 &.845 &.056 &.909 &.898 &.945 &.031 &.857 &.845 &.899 &.064 &.818 &.807 &.846 &.078   \\
%   $\text{CABP}$ &.828 &.757 &.859 &.058 &.887 &.877 &.913 &.055 &.778 &.670 &.801 &.078 &.878 &.855 &.913 &.047 &.810 &.782 &.845 &.094 &.773 &.744 &.799 &.102   \\
   $\text{CIGAN}$ &.876 &.820 &.906 &.042 &.923 &.913 &.945 &.037 &.823 &.733 &.848 &.061 &.911 &.892 &.943 &.034 &.856 &.836 &.893 &.068 &.833 &.816 &.862 &.075   \\
%   $\text{TGAN}$ &.907 &.877 &.944 &.029 &.939 &.938 &.964 &.025 &.852 &.789 &.882 &.051 &.927 &.920 &.963 &.024 &.878 &.872 &.918 &.053 &.855 &.849 &.894 &.061   \\
%   $\text{TABP}$ &.910 &.878 &.944 &.028 &.942 &.940 &.966 &.024 &.860 &.799 &.891 &.048 &.929 &.922 &.964 &.024 &.879 &.870 &.918 &.054 &.860 &.858 &.897 &.061   \\
   $\text{TIGAN}$ &\textbf{.909} &.873 &\underline{.941} &\textbf{.028} &\textbf{.941} &\textbf{.936} &\textbf{.964} &\textbf{.025} &\textbf{.861} &\underline{.796} &\underline{.890} &\textbf{.047} &\underline{.929} &.918 &\underline{.962} &\underline{.025} &\textbf{.879} &\textbf{.869} &\textbf{.916} &\textbf{.054} &\textbf{.861} &\underline{.854} &\underline{.894} &\textbf{.060}   \\
 \hline
%  CPD \cite{cpd_sal} & .869 & .821 & .898 & .043 & .913 & .909 & .937 & .040 & .825 & .742 & .847 & .056 & .906 & .892 & .938 & .034 & .848 & .819 & .882 & .071 & .799 & .779 & .811 & .088  \\
   SCRN \cite{scrn_sal} & .885 & .833 & .900 & .040 & .920 & .910 & .933 & .041 & .837 & .749 & .847 & .056 & .916 & .894 & .935 & .034 & .869 & .833 & .892 & .063 & .817 & .790 & .829 & .087\\ 
%   PoolNet \cite{Liu19PoolNet} & .887 & .840 & .910 & .037 & .919 & .913 & .938 & .038 & .831 & .748 & .848 & .054 & .919 & .903 & .945 & .030 & .865 & .835 & .896 & .065 & .820 & .804 & .834 & .084 \\ 
    % BASNet \cite{basnet_sal} & .876 & .823 & .896 & .048 & .910 & .913 & .938 & .040 & .836 & .767 & .865 & .057 & .909 & .903 & .943 & .032 & .838 & .818 & .879 & .076 & .798 & .792 & .827 & .094\\ 
%   EGNet \cite{zhao2019EGNet} & .878 & .824 & .898 & .043 & .914 & .906 & .933 & .043 & .840 & .755 & .855 & .054 & .917 & .900 & .943 & .031 & .852 & .823 & .881 & .074 & .824 & .811 & .843 & .081 \\
   F3Net \cite{wei2020f3net} & .888 & .852 & .920 & .035 & .919 & .921 & .943 & .036 & .839 & .766 & .864 & .053 & .917 & .910 & .952 & .028 & .861 & .835 & .898 & \underline{.062} & .824 & .814 & .850 & .077\\
   ITSD \cite{zhou2020interactive} & .886 & .841 & .917 & .039 & .920 & .916 & .943 & .037 & .842 & .767 & .867 & .056 & .921 & .906 & .950 & .030 & .860 & .830 & .894 & .066 & .836 & .829 & .867 & .076\\
%   UIS \cite{ucnet_jornal}  & .888 & .860 & .927 & .034 & .921 & .926 & .947 & .035 & .839 & .773 & .869 & .051 & .921 & .919 & .957 & .026 & . & . & . & .  & . & . & . & . \\ 
    % UJSC \cite{aixuan_cod_sod21} & .899 & .866 & .937 & .032 & .933 & .935 & .960 & .030 & .850 & .782 & .884 & .051 & .931 & .924 & .867 & .026 & . & . & . & .  & . & . & . & .  \\ 
    PAKRN \cite{xu2021locate} & .900 & \textbf{.876} & .935 & .033  & .928 & .930 & .951 & .032 & .853 & \underline{.796} & .888 & \underline{.050} & .923 & \underline{.919} & .955 & .028 & .859 & \underline{.856} & .898 & .068  & .833 & .836 & .866 & .074 \\ 
    % DCN ~\cite{DCN_TIP} &2021&R50& .892 & .859 & .924 & .035  & .928 & .931 & .954 & .032 & .846 & .779 & .875 & .051 & .922 & .916 & .957 & .027 & . & . & . & . & . & . & . & .  \\ 
    MSFNet \cite{zhang2021auto} & .877 & .855 & .927 & .034  & .915 & .927 & .951 & .033 & .832 & .772 & .873 & \underline{.050} & .909 & .913 & .957 & .027 & .849 & .855 & .900 & .064  & .813 & .822 & .852 & .077  \\ 
    % MINet\cite{pang_MINet_CVPR_2020} &.875 &.823 &.912 &.039 &.919 &.922 &.947 &.036 &.822 &.741 &.864 &.057 &.914 &.906 &.955 &.030 &.855 &.843 &.898 &.065 &- &- &- &- \\
    CTDNet\cite{zhao_CTDNet_ACMMM_2021} &.893 &.862 &.928 &.034 &.925 &.928 &.950 &.032 &.844 &.779 &.874 &.052 &.919 &.915 &.954 &.028 &.861 &\underline{.856} &.901 &.064 &.829 &.832 &.858 &.074 \\\hline
    VST\cite{liu_ICCV_2021_VST} &.896 &.842 &.918 &.037 &.932 &.911 &.943 &.034 &.850 &.771 &.869 &.058 &.928 &.903 &.950 &.030 &.873 &.832 &.900 &.067 &.854 &.833 &.879 &.065 \\
    GTSOD \cite{zhang2021learning_nips} &\underline{.908} &\underline{.875} &\textbf{.942} &\underline{.029} &\underline{.935} &\underline{.935} &\underline{.962} &\underline{.026} &\underline{.858} &\textbf{.797} &\textbf{.892} &.051 &\textbf{.930} &\textbf{.922} &\textbf{.964} &\textbf{.023} &\underline{.877} &.855 &\underline{.915} &\textbf{.054} &\underline{.860} &\textbf{.860} &\textbf{.898} &\underline{.061} \\
   \hline
  \end{tabular}
  \label{tab:benchmark_rgb_sod}
\end{table*}

\begin{table*}[t!]
  \centering
  \scriptsize
  \renewcommand{\arraystretch}{1.2}
  \renewcommand{\tabcolsep}{0.65mm}
  \caption{\footnotesize{Performance comparison with benchmark fully-supervised RGB-D SOD models.}}
  \begin{tabular}{l|cccc|cccc|cccc|cccc|cccc|cccc}
  \hline
% \toprule
  &\multicolumn{4}{c|}{NJU2K \cite{NJU2000}}&\multicolumn{4}{c|}{SSB \cite{niu2012leveraging}}&\multicolumn{4}{c|}{DES \cite{cheng2014depth}}&\multicolumn{4}{c|}{NLPR \cite{peng2014rgbd}}&\multicolumn{4}{c|}{LFSD \cite{li2014saliency}}&\multicolumn{4}{c}{SIP \cite{sip_dataset}} \\
    Method & $S_{\alpha}\uparrow$&$F_{\beta}\uparrow$&$E_{\xi}\uparrow$&$\mathcal{M}\downarrow$& $S_{\alpha}\uparrow$&$F_{\beta}\uparrow$&$E_{\xi}\uparrow$&$\mathcal{M}\downarrow$& $S_{\alpha}\uparrow$&$F_{\beta}\uparrow$&$E_{\xi}\uparrow$&$\mathcal{M}\downarrow$& $S_{\alpha}\uparrow$&$F_{\beta}\uparrow$&$E_{\xi}\uparrow$&$\mathcal{M}\downarrow$& $S_{\alpha}\uparrow$&$F_{\beta}\uparrow$&$E_{\xi}\uparrow$&$\mathcal{M}\downarrow$& $S_{\alpha}\uparrow$&$F_{\beta}\uparrow$&$E_{\xi}\uparrow$&$\mathcal{M}\downarrow$ \\ \hline
%     $\text{CGAN}$ &.909 &.896 &.938 &.038 &.903 &.880 &.934 &.040 &.931 &.919 &.963 &.019 &.914 &.889 &.948 &.026 &.829 &.809 &.860 &.084 &.880 &.867 &.917 &.050   \\
%   $\text{CABP}$ &.907 &.891 &.934 &.040 &.904 &.877 &.931 &.041 &.925 &.907 &.949 &.022 &.913 &.881 &.942 &.027 &.824 &.795 &.847 &.089 &.872 &.852 &.906 &.056  \\
%   $\text{CIGAN}$ &.910 &.892 &.934 &.040 &.904 &.874 &.930 &.042 &.932 &.913 &.960 &.021 &.911 &.874 &.940 &.029 &.832 &.806 &.858 &.087 &.876 &.855 &.910 &.054   \\
$\text{CIGAN}$ &.914 &.900 &.939 &.036 &.903 &.876 &.934 &.040 &.937 &.921 &\underline{.970} &.018 &.922 &.890 &.952 &.025 &.851 &.832 &.889 &.075 &.884 &.870 &.917 &.049   \\ 
%   $\text{TGAN}$ &.925 &.918 &.955 &.029 &.921 &.905 &.952 &.030 &.918 &.908 &.943 &.022 &.932 &.915 &.963 &.020 &.871 &.856 &.898 &.068 &.895 &.899 &.937 &.040   \\
%   $\text{TABP}$ &.921 &.908 &.948 &.032 &.910 &.884 &.942 &.037 &.936 &.916 &.966 &.018 &.927 &.904 &.957 &.022 &.888 &.878 &.917 &.055 &.895 &.894 &.929 &.043   \\
%   $\text{TIGAN}$ &.922 &.909 &.948 &.032 &.910 &.884 &.942 &.037 &.933 &.913 &.964 &.019 &.927 &.902 &.956 &.023 &.884 &.871 &.912 &.056 &.893 &.890 &.929 &.044   \\
$\text{TIGAN}$ &\underline{.928} &\underline{.919} &\textbf{.956} &\textbf{.028} &\underline{.915} &\underline{.893} &\underline{.947} &\underline{.034} &.940 &\textbf{.929} &\underline{.970} &\textbf{.016} &\underline{.932} &\underline{.911} &\underline{.961} &\underline{.020} &\textbf{.884} &\underline{.868} &\underline{.911} &\textbf{.057} &\underline{.905} &\underline{.901} &\textbf{.941} &\textbf{.037}   \\
 \hline
   BBSNet \cite{fan2020bbs}  &.921 &.902 &.938 &\underline{.035}  &.908 &.883 &.928 &.041 &.933 &.910 &.949 &.021 &.930 &.896 &.950 &.023 &.864 &.843 &.883 &.072 &.879 &.868 &.906 &.055 \\
   BiaNet \cite{zhang2020bilateral}  &.915 &.903 &.934 &.039  &.904 &.879 &.926 &.043 &.931 &.910 &.948 &.021 &.925 &.894 &.948 &.024 &.845 &.834 &.871 &.085 &.883 &.873 &.913 &.052 \\
   CoNet \cite{ji2020accurate}  &.911 &.903 &\underline{.944} &.036  &.896 &.877 &.939 &.040 &.906 &.880 &.939 &.026 &.900 &.859 &.937 &.030 &.842 &.834 &.886 &.077 &.868 &.855 &.915 &.054 \\
   UCNet \cite{zhang2021_ucnet} &.897 &.886 &.930 &.043 &.903 &.884 &.938 &.039 &.934 &.919 &.967 &.019 &.920 &.891 &.951 &.025 &.864 &.855 &.901 &.066 &.875 &.867 &.914 &.051 \\
   JLDCF \cite{Fu2020JLDCF} &.902 &.885 &.935 &.041  &.903 &.873 &.936 &.040 &.931 &.907 &.959 &.021 &.925 &.894 &.955 &.022 &.862 &.848 &.894 &.070 &.880 &.873 &.918 &.049 \\ \hline
   VST \cite{liu_ICCV_2021_VST} &.922 &.898 &.939 &\underline{.035} &.913 &.879 &.937 &.038 &\underline{.943} &.920 &.965 &\underline{.017} &\underline{.932} &.897 &.951 &.024 &\underline{.882} &\textbf{.871} &\textbf{.917} &\underline{.061} &.904 &.894 &.933 &\underline{.040} \\
   GTSOD \cite{zhang2021learning_nips} &\textbf{.929} &\textbf{.924} &\textbf{.956} &\textbf{.028} &\textbf{.916} &\textbf{.898} &\textbf{.950} &\textbf{.032} &\textbf{.945} &\underline{.928} &\textbf{.971} &\textbf{.016} &\textbf{.938} &\textbf{.921} &\textbf{.966} &\textbf{.018} &.872 &.862 &.901 &.066 &\textbf{.906} &\textbf{.908} &\underline{.940} &\textbf{.037}\\
%   $\text{TADE}$ &.925 &.917 &.953 &.029 &.911 &.890 &.946 &.034 &.944 &.930 &.977 &.015 &.934 &.913 &.965 &.018 &.879 &.869 &.910 &.056 &.902 &.895 &.939 &.038  \\
% Ours &\textbf{.928} &\textbf{.919} &\textbf{.955} &\textbf{.028} &\textbf{.915} &\textbf{.894} &\textbf{.947} &\textbf{.034} &\textbf{.942} &\textbf{.931} &\textbf{.971} &\textbf{.015} &\textbf{.932} &\textbf{.913} &\textbf{.962} &\textbf{.020} &\textbf{.877} &\textbf{.864} &\textbf{.904} &\textbf{.062} &\textbf{.897} &\textbf{.897} &\textbf{.933} &\textbf{.041} \\
% Ours &\textbf{.927} &\textbf{.922} &\textbf{.956} &\textbf{.028} &\textbf{.919} &\textbf{.902} &\textbf{.951} &\textbf{.031} &\textbf{.939} &\textbf{.931} &\textbf{.970} &\textbf{.015} &\textbf{.935} &\textbf{.917} &\textbf{.964} &\textbf{.019} &\textbf{.877} &\textbf{.867} &\textbf{.907} &\textbf{.059} &\textbf{.905} &\textbf{.911} &\textbf{.943} &\textbf{.036} \\
% Ours\_1k &\textbf{.927} &\textbf{.922} &\textbf{.956} &\textbf{.028} &\textbf{.919} &\textbf{.902} &\textbf{.951} &\textbf{.031} &\textbf{.939} &\textbf{.931} &\textbf{.970} &\textbf{.015} &\textbf{.935} &\textbf{.917} &\textbf{.964} &\textbf{.019} &\textbf{.877} &\textbf{.867} &\textbf{.907} &\textbf{.059} &\textbf{.905} &\textbf{.911} &\textbf{.943} &\textbf{.036} \\
% Ours\_22k &.934 &.929 &.962 &.024 &.922 &.905 &.952 &.029 &.944 &.936 &.975 &.014 &.937 &.921 &.966 &.019 &.884 &.868 &.910 &.056 &.901 &.910 &.939 &.038 \\
   \hline 
% \bottomrule
  \end{tabular}
  \label{tab:benchmark_rgbd_sod}
%   \vspace{-5mm}
\end{table*}

%=======================================================================================
\section{Experiments}
% \Jing{Yunqiu, Aixuan: introduce dataset and metric for sod, cod}
% Before we present the proposed framework, we will first introduce the datasets and evaluation metrics for easier reference.

\noindent\textbf{Dataset:} 
In this paper, we conduct research on salient object detection (SOD), including both RGB image-based SOD and RGB-D image pair-based SOD. For the former, we perform fully and weakly supervised saliency detection. Within the fully supervised learning frameworks,
% RGB image based salient object detection task, 
we train the models by using the DUTS training dataset \cite{imagesaliency} $D1=\{x_i,y_i\}_{i=1}^N$ of size $N=10,553$, and test on six other widely used datasets: the DUTS testing dataset, ECSSD~\cite{yan2013hierarchical}, DUT~\cite{Manifold-Ranking:CVPR-2013}, HKU-IS~\cite{li2015visual}, PASCAL-S~\cite{pascal_s_dataset} and the SOD testing dataset \cite{sod_dataset}. For the weakly supervised models,
% RGB image based salient object detection, 
we use the DUTS-S training dataset~\cite{jing2020weakly}, $D2=\{x_i,y_i\}_{i=1}^N$ of size $N=10,553$, where $y_i$ is the scribble annotation.
% instead of pixel-wise annotation in $D1$.
% For semi-supervised salient object detection, we random select $N^l=1,000$ images from DUTS training dataset \cite{imagesaliency} serving as the labeled set $D^l=\{x_i,y_i\}_{i=1}^{N^l}$, and the $N^u=9,553$ images serves as the unlabeled set $D^u=\{x_i,y_i\}_{i=1}^{N^u}$. 
The testing dataset is the same as the fully supervised RGB SOD models.
% We use the same testing dataset for models related to the two different settings.
% For the salient object detection task, we train the models by using DUTS training  dataset\cite{wang2017learning}, and testing them on six other widely used data sets: the DUTS testing dataset, ECSSD \cite{yan2013hierarchical}, DUT \cite{Manifold-Ranking:CVPR-2013}, HKU-IS \cite{li2015visual}, THUR \cite{THUR} and SOC testing dataset \cite{fan2018SOC}.
% For the camouflaged object detection task, we train the models using the COD10K training dataset\cite{fan2020camouflaged}, and test them on four test datasets:  CAMO \cite{le2019anabranch}, CHAMELEON \cite{Chameleon2018}, the COD10K testing dataset, and NC4K\cite{yunqiu_cod21}.

\begin{table*}[t!]
  \centering
  \scriptsize
  \renewcommand{\arraystretch}{1.1}
  \renewcommand{\tabcolsep}{0.658mm}
  \caption{\footnotesize{Performance comparison with benchmark weakly-supervised RGB SOD models.}}
  \begin{tabular}{l|cccc|cccc|cccc|cccc|cccc|cccc}
  \hline
% \toprule
  &\multicolumn{4}{c|}{DUTS \cite{imagesaliency}}&\multicolumn{4}{c|}{ECSSD \cite{yan2013hierarchical}}&\multicolumn{4}{c|}{DUT \cite{Manifold-Ranking:CVPR-2013}}&\multicolumn{4}{c|}{HKU-IS \cite{li2015visual}}&\multicolumn{4}{c|}{PASCAL-S \cite{pascal_s_dataset}}&\multicolumn{4}{c}{SOD \cite{sod_dataset}} \\
    Method & $S_{\alpha}\uparrow$&$F_{\beta}\uparrow$&$E_{\xi}\uparrow$&$\mathcal{M}\downarrow$& $S_{\alpha}\uparrow$&$F_{\beta}\uparrow$&$E_{\xi}\uparrow$&$\mathcal{M}\downarrow$& $S_{\alpha}\uparrow$&$F_{\beta}\uparrow$&$E_{\xi}\uparrow$&$\mathcal{M}\downarrow$& $S_{\alpha}\uparrow$&$F_{\beta}\uparrow$&$E_{\xi}\uparrow$&$\mathcal{M}\downarrow$& $S_{\alpha}\uparrow$&$F_{\beta}\uparrow$&$E_{\xi}\uparrow$&$\mathcal{M}\downarrow$& $S_{\alpha}\uparrow$&$F_{\beta}\uparrow$&$E_{\xi}\uparrow$&$\mathcal{M}\downarrow$ \\ \hline
%   $\text{CGAN}$ &.830 &.779 &.883 &.054 &.890 &.888 &.932 &.047 &.793 &.708 &.829 &.068 &.881 &.870 &.933 &.041 &.831 &.819 &.884 &.076 &.789 &.789 &.844 &.088   \\
%   $\text{CABP}$ &. &. &. &. &. &. &. &. &. &. &. &. &. &. &. &. &. &. &. &. &. &. &. &.   \\
%   $\text{CIGAN}$ &.834 &.779 &.887 &.056 &.896 &.890 &.938 &.044 &.799 &.713 &.838 &.070 &.886 &.873 &.938 &.039 &.827 &.810 &.880 &.079 &.800 &.793 &.855 &.083   \\
   $\text{CIGAN}$ &.834 &.779 &.887 &.056 &\underline{.896} &.890 &\underline{.938} &\underline{.044} &.799 &.713 &.838 &.070 &\underline{.886} &.873 &\underline{.938} &.039 &\underline{.827} &.810 &\underline{.880} &.079 &\underline{.800} &\underline{.793} &\underline{.855} &\underline{.083}   \\ 
%   $\text{TGAN}$ &.859 &.819 &.920 &.041 &.908 &.909 &.952 &.035 &.833 &.768 &.879 &.054 &.895 &.888 &.950 &.035 &.845 &.841 &.902 &.066 &.813 &.812 &.869 &.082   \\
%   $\text{TABP}$ &. &. &. &. &. &. &. &. &. &. &. &. &. &. &. &. &. &. &. &. &. &. &. &.   \\
%   $\text{TIGAN}$ &.851 &.806 &.914 &.044 &.903 &.902 &.949 &.037 &.823 &.754 &.869 &.059 &.889 &.879 &.946 &.037 &.842 &.833 &.900 &.067 &.808 &.800 &.867 &.084   \\
   $\text{TIGAN}$ &\textbf{.855} &\underline{.814} &\textbf{.918} &\textbf{.043} &\textbf{.905} &\textbf{.905} &\textbf{.950} &\textbf{.037} &\textbf{.826} &\textbf{.760} &\textbf{.874} &\textbf{.058} &\textbf{.893} &\underline{.887} &\textbf{.949} &\textbf{.035} &\textbf{.844} &\textbf{.839} &\textbf{.902} &\textbf{.066} &\textbf{.811} &\textbf{.810} &\textbf{.872} &\textbf{.082}   \\
 \hline
 SSAL~\cite{jing2020weakly} & .803 & .747 & .865 & .062 & .863 & .865 & .908 & .061 & .785 & .702 & .835 & .068 & .865 & .858 & .923 & .047 & .798 & .773 & .854 & .093 & .750 & .743 & .801 & .108\\ 
   WSS~\cite{imagesaliency}  &.748 &.633 &.806 &.100 &.808 &.774 &.801 &.106 &.730 &.590 &.729 &.110 &.822&.773 &.819 &.079 & .701 & .691 & .687 & .187 & .698 & .635 & .687 & .152  \\
   C2S~\cite{xin2018c2s} &.805 &.718 &.845 &.071 &- &- &- &- &.773 &.665 &.810 &.082 &.869 &.837 &.910 &.053 & .784 & .806 & .813 & .130 & .770 & .741 & .799 & .117\\
   SCWS~\cite{structure_consistency_scribble}  &\underline{.841} &\textbf{.818} &\underline{.901} &\underline{.049} &.879 &\underline{.894} &.924 &.051 &\underline{.813} &\underline{.751} &\underline{.856} &\underline{.060} &.883 &\textbf{.892} &\underline{.938} &\underline{.038} &.821 &\underline{.815} &.877 &\underline{.078} &.782 &.791 &.833 &.090 \\ \hline
  \end{tabular}
  \label{tab:benchmark_weakly_supervised_rgb_sod}
\end{table*}

For RGB-D SOD, we follow the conventional setting, where the training set $D3=\{x_i,y_i\}_{i=1}^N$ is a combination of 1,485 images from the NJU2K dataset~\cite{NJU2000} and 700 images from the NLPR dataset~\cite{peng2014rgbd}. We then test the performance of our model and competing models on the NJU2K testing set, NLPR testing set, LFSD~\cite{li2014saliency}, DES~\cite{cheng2014depth}, SSB~\cite{niu2012leveraging} and SIP~\cite{sip_dataset} dataset.
% , DUT-RGBD dataset \cite{dmra_iccv19} and ReDWeb-S testing set \cite{ReDWeb-S_dataset}.

% To further explore how the transformer performs with different sets of the
% \NB{the} 
% training dataset, we also train the RGB-D salient object detection models with the newly released COME15K training dataset $D4=\{x_i,y_i\}_{i=1}^N$ of size 8,025 and test on the same benchmark RGB-D salient object detection dataset. 

\noindent\textbf{Evaluation Metrics:} For all three tasks, we use four evaluation metrics to measure the performance, including Mean Absolute Error $\mathcal{M}$, Mean F-measure ($F_{\beta}$), Mean E-measure ($E_{\xi}$)~\cite{fan2018enhanced} and S-measure ($S_{\alpha}$)~\cite{fan2017structure}.
% \Jing{Xinyu: introduce dataset and metric for medical image segmentation}

\textbf{MAE $\mathcal{M}$} is defined as the pixel-wise difference between the prediction $s$
% predicted $c$ 
and the
% pixel-wise binary 
ground truth $y$: $\mathcal{M} = \frac{1}{H\times W}|c-y|$,
% \begin{equation}
%     \begin{aligned}
%     \text{MAE} = \frac{1}{H\times W}|c-y|,
%     \end{aligned}
% \end{equation}
where $H$ and $W$ are the height and width of $c$ correspondingly.
% MAE provides a direct estimation of the conformity between the estimated camouflage map and ground-truth camouflage map. However, for the MAE metric, small objects naturally assign a smaller error and larger objects have larger errors.

% The metric also can not tell where the
% error occurs~\cite{tsai2010motion}.
\textbf{F-measure $F_{\beta}$} is a region-based similarity metric, and we provide the mean F-measure using varying fixed (0-255) thresholds.

\textbf{E-measure $E_{\xi}$} is the recently proposed Enhanced alignment measure~\cite{fan2018enhanced} in the binary map evaluation field to jointly capture image-level statistics and local pixel matching information.

\textbf{S-measure $S_{\alpha}$} is a structure based measure~\cite{fan2017structure}, which combines the region-aware ($S_r$) and object-aware ($S_o$) structural similarity as their final structure metric: $S_{\alpha} = \alpha S_o+(1-\alpha) S_r$,
% \begin{equation}
% \label{equ:S-measure}
% S_{\alpha} = \alpha S_o+(1-\alpha) S_r,
% \end{equation}
where $\alpha\!\in\![0,1]$ is set to 0.5 by default.

\noindent\Rev{\textbf{Calibration measures:} Due to the close correlation between uncertainty estimation and model calibration~\cite{on_calibration}, uncertainty is usually evaluated with modal calibration measures~\cite{franchi2022latent}, \ie expected calibration error (ECE)~\cite{degroot1983comparison}.
% , negative log likelihood (NLL)
% % , expected normalized calibration error (ENCE)~\cite{levi2022evaluating} 
% and \etc. 
The basic assumption is that a reliable uncertainty output should lead to a well-calibrated model, where model confidence is consistent with model accuracy.}

\subsection{Accurate and Reliable Fully-supervised Salient Object Detection}
\label{sec:accurate_reliable_full_sod}

\subsubsection{Performance Comparison with Benchmark Models}
% In Section \ref{sec:accurate_reliable_full_sod} and \ref{sec:accurate_reliable_weak_sod}, we analyzed the proposed accurate and reliable transformer based framework for both fully and weakly supervised SOD.
% We then comprehensively discussed the transformer backbone based saliency models in Section \ref{sec:model_discussion}. 
% In this Section, w
We compare the proposed framework with benchmark saliency models and show model performance in Tables \ref{tab:benchmark_rgb_sod}, \ref{tab:benchmark_rgbd_sod} and  \ref{tab:benchmark_weakly_supervised_rgb_sod}.
% for fully supervised RGB saliency detection, RGB-D saliency detection and weakly supervised RGB saliency respectively. 
Note that, VST \cite{liu_ICCV_2021_VST} and GTSOD \cite{zhang2021learning_nips} are two existing transformer based saliency detection models.
% , and both of them provide models for RGB and RGB-D saliency detection as shown in Table \ref{tab:benchmark_rgb_sod} and Table \ref{tab:benchmark_rgbd_sod}.

% respectively, , we compare the proposed iGAN based framework with benchmark fully supervised RGB saliency detection, RGB-D saliency detection and weakly supervised RGB saliency detection respectively. 

% Comparing our CNN based generative model ($\text{CIGAN}$) with existing techniques
% % CNN backbone based saliency models 
% in Tables \ref{tab:benchmark_rgb_sod} and \ref{tab:benchmark_rgbd_sod}, w
We observe the competitive performance of our CNN based generative model ($\text{CIGAN}$) with existing techniques
% CNN backbone based saliency models 
in Tables \ref{tab:benchmark_rgb_sod} and \ref{tab:benchmark_rgbd_sod}. To focus on explaining the superior performance of the transformer backbone for SOD, our decoder has
% is simple, with 
only 1M parameters, which is around 5\% of model parameters of existing techniques. Further, we find a better performance of our generative model ($\text{TIGAN}$) compared with VST \cite{liu_ICCV_2021_VST}, indicating the superiority of the proposed model. Different from the deterministic VST~\cite{liu_ICCV_2021_VST}, as a generative model, we aim to produce stochastic predictions leading to reliable saliency prediction. In this way, we compare with GTSOD~\cite{zhang2021learning_nips}, another generative transformer SOD model, in the way of both accurate and reliable saliency prediction. Tables \ref{tab:benchmark_rgb_sod} and \ref{tab:benchmark_rgbd_sod} show that the proposed iGAN achieves comparable performance compared with GTSOD \cite{zhang2021learning_nips}, leading to an alternative generative saliency transformer. In Fig.~\ref{fig:generative_performance_comparison}, we further visualize the produced uncertainty maps of GTSOD \cite{zhang2021learning_nips} and ours for RGB SOD. The more reliable uncertainty maps, highlighting the less confident or hard regions, further explain our superiority. \Rev{Besides the visual comparison, we also compute calibration measures of GTSOD \cite{zhang2021learning_nips} and ours and show the results
% the conventional latent variable models and the proposed iGAN and show performance 
in Table~\ref{tab:mc_dropout_analysis}, which clearly shows the advantages of our model in achieving better calibrated models compared with GTSOD \cite{zhang2021learning_nips}.}

\subsubsection{Accurate Saliency Model}
\begin{figure}[tp]
%  \vspace{-5mm}
  \begin{center}
  \begin{tabular}{c@{ } c@{ } c@{ } c@{ } c@{ } c@{ } }
  {\includegraphics[width=0.152\linewidth]{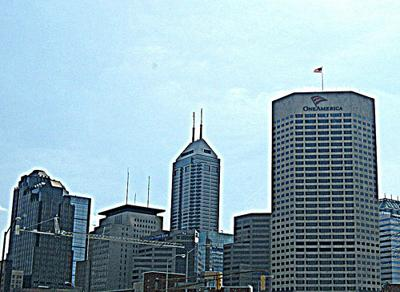}} &
  {\includegraphics[width=0.152\linewidth]{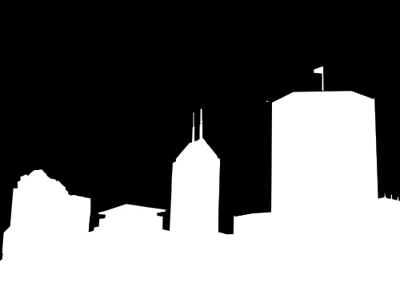}} &
  {\includegraphics[width=0.152\linewidth]{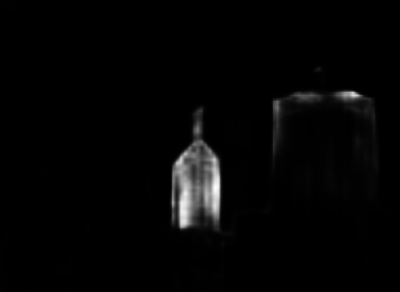}} &
  {\includegraphics[width=0.152\linewidth]{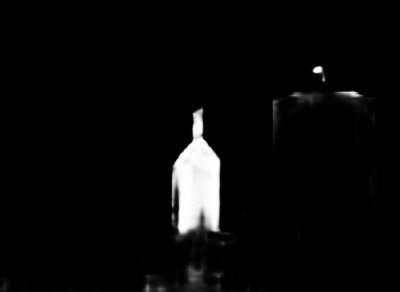}} &
  {\includegraphics[width=0.152\linewidth]{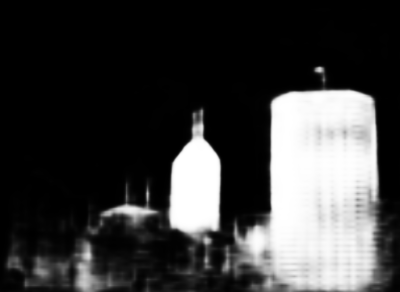}} &
  {\includegraphics[width=0.152\linewidth]{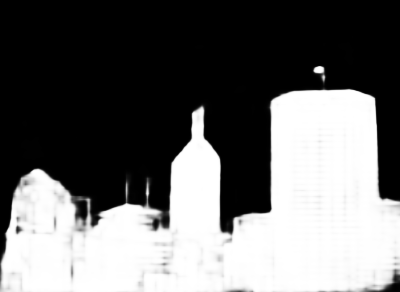}} \\
  {\includegraphics[width=0.152\linewidth]{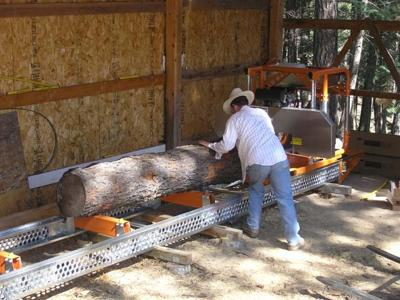}} &
  {\includegraphics[width=0.152\linewidth]{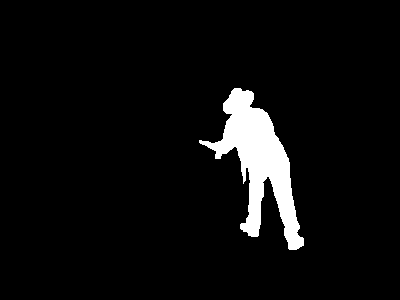}} &
  {\includegraphics[width=0.152\linewidth]{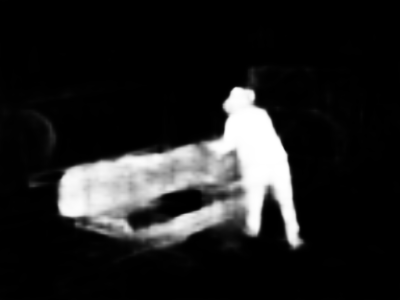}} &
  {\includegraphics[width=0.152\linewidth]{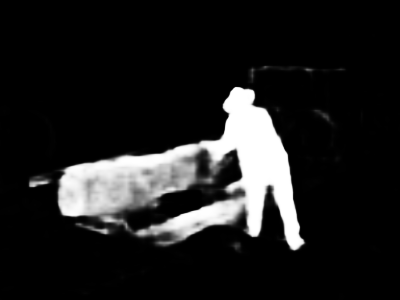}} &
  {\includegraphics[width=0.152\linewidth]{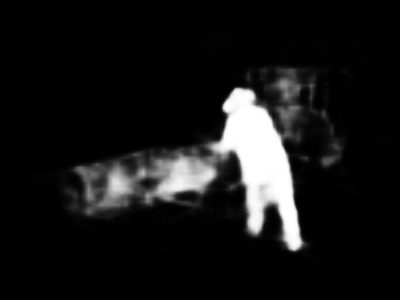}} &
  {\includegraphics[width=0.152\linewidth]{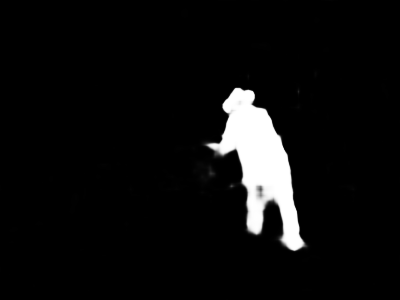}} \\
  \footnotesize{Image}&\footnotesize{GT}&\footnotesize{$\text{B\_{cnn}}$}&\footnotesize{$\text{B'\_{cnn}}$}&\footnotesize{$\text{B\_{tr}}$}&\footnotesize{$\text{B'\_{tr}}$}
  \end{tabular}
  \end{center}
%   \vspace{-5pt}
  \caption{\footnotesize{Predictions of CNN and transformer backbone models
%   ($\text{B\_{cnn}}$ and $\text{B'\_{cnn}}$) and transformer backbone models ($\text{B\_{tr}}$ and $\text{B'\_{tr}}$) 
  without ($\text{B\_{cnn}}$ and $\text{B\_{tr}}$) and with ($\text{B'\_{cnn}}$ and $\text{B'\_{tr}}$) structure-aware loss.}
  }
\label{fig:viaulization_structure_aware_loss_effects}
\end{figure}

In Sec.~\ref{sec:intro}, we discuss that the CNN backbone is not effective in detecting salient objects that rely on global context and the stride and pooling operation lead to less accurate structure information of CNN backbone features.
% \subsubsection{Experimental Results}
We then compare the performance of
% the structure-modeling ability of 
CNN backbone ($\text{B\_{cnn}}$ with ResNet50 \cite{resnet} backbone) and transformer backbone model ($\text{B\_{tr}}$ with Swin transformer backbone \cite{liu2021swin}) for RGB image based SOD, and show performance in Table \ref{tab:fully_rgb_sod_expeiments}, where the models share the same decoder\footnote{We adjust the decoder accordingly to the backbone features.}. Note that, for both $\text{B\_{cnn}}$ and $\text{B\_{tr}}$, we use binary cross-entropy loss for the saliency generator. To further explain how the two types of backbone based models perform with structure-aware loss~\cite{wei2020f3net} in Eq.~\eqref{structure_loss}, we train $\text{B\_{cnn}}$ and $\text{B\_{tr}}$ with a structure-aware loss instead and obtain model $\text{B'\_{cnn}}$ and $\text{B'\_{tr}}$ respectively. We also visualize predictions from the two different backbone networks in Fig.~\ref{fig:viaulization_structure_aware_loss_effects}.
% with the
% % \NB{the} 
% structure-aware loss function \cite{wei2020f3net} in Eq.~\eqref{structure_loss} and the conventional binary cross-entropy loss function.
The significant performance gap between
% \NB{between} \sout{of} 
the two different
% \NB{different} 
backbones
% \sout{based frameworks} 
($\text{B\_{cnn}}$ and $\text{B\_{tr}}$) indicates the superiority of the transformer backbone for SOD. Further, we observe, although the transformer has encoded accurate structure information, the structure-aware loss function ($\text{B'\_{tr}}$) that penalizes the wrong prediction along object boundaries can lead to
% further strengthen the structure-accurate constraint, leading to 
more accurate predictions (see Table \ref{tab:fully_rgb_sod_expeiments} and Fig.~\ref{fig:viaulization_structure_aware_loss_effects}).

\begin{table*}[t!]
  \centering
  \scriptsize
  \renewcommand{\arraystretch}{1.1}
  \renewcommand{\tabcolsep}{0.74mm}
  \caption{\footnotesize{Baseline model performance,
%   Performance comparison with CNN and transformer backbone based baseline models,
where $\text{B\_{cnn}}$ and $\text{B\_{tr}}$ are models with CNN and transformer backbones respectively using a binary cross-entropy loss function, $\text{B'\_{cnn}}$ and $\text{B'\_{tr}}$ are the corresponding models with the structure-aware loss function in Eq.~\eqref{structure_loss}.}}
  \begin{tabular}{l|cccc|cccc|cccc|cccc|cccc|cccc}
  \hline
% \toprule
  &\multicolumn{4}{c|}{DUTS \cite{imagesaliency}}&\multicolumn{4}{c|}{ECSSD \cite{yan2013hierarchical}}&\multicolumn{4}{c|}{DUT \cite{Manifold-Ranking:CVPR-2013}}&\multicolumn{4}{c|}{HKU-IS \cite{li2015visual}}&\multicolumn{4}{c|}{PASCAL-S \cite{pascal_s_dataset}}&\multicolumn{4}{c}{SOD \cite{sod_dataset}} \\
    Method & $S_{\alpha}\uparrow$&$F_{\beta}\uparrow$&$E_{\xi}\uparrow$&$\mathcal{M}\downarrow$& $S_{\alpha}\uparrow$&$F_{\beta}\uparrow$&$E_{\xi}\uparrow$&$\mathcal{M}\downarrow$& $S_{\alpha}\uparrow$&$F_{\beta}\uparrow$&$E_{\xi}\uparrow$&$\mathcal{M}\downarrow$& $S_{\alpha}\uparrow$&$F_{\beta}\uparrow$&$E_{\xi}\uparrow$&$\mathcal{M}\downarrow$& $S_{\alpha}\uparrow$&$F_{\beta}\uparrow$&$E_{\xi}\uparrow$&$\mathcal{M}\downarrow$& $S_{\alpha}\uparrow$&$F_{\beta}\uparrow$&$E_{\xi}\uparrow$&$\mathcal{M}\downarrow$ \\ \hline
   $\text{B\_{cnn}}$ &.878 &.818 &.895 &.042 &\underline{.922} &.907 &.937 &.039 &.822 &.724 &.834 &.061 &.912 &.886 &.933 &.035 &.862 &.838 &.891 &.067 &.831 &.808 &.846 &.079  \\
   $\text{B'\_{cnn}}$ &.882 &.840 &.916 &.037 &\underline{.922} &.919 &.947 &.035 &.823 &.742 &.851 &.057 &.912 &.901 &.947 &.030 &.855 &.841 &.896 &.065 &.832 &.825 &.863 &.073  \\ \hline
   $\text{B\_{tr}}$ &\underline{.907} &\underline{.863} &\underline{.930} &\underline{.031} &\textbf{.939} &\underline{.929} &\underline{.957} &\underline{.028} &\underline{.858} &\underline{.786} &\underline{.878} &\underline{.051} &\textbf{.929} &\underline{.912} &\underline{.954} &\underline{.027} &\textbf{.881} &\underline{.866} &\underline{.911} &\underline{.056} &\underline{.854} &\underline{.841} &\underline{.882} &\underline{.065}  \\
   $\text{B'\_{tr}}$ &\textbf{.911} &\textbf{.882} &\textbf{.947} &\textbf{.026} &\textbf{.939} &\textbf{.940} &\textbf{.965} &\textbf{.024} &\textbf{.860} &\textbf{.801} &\textbf{.894} &\textbf{.045} &\underline{.927} &\textbf{.921} &\textbf{.964} &\textbf{.023} &\underline{.876} &\textbf{.872} &\textbf{.917} &\textbf{.053} &\textbf{.858} &\textbf{.853} &\textbf{.897} &\textbf{.059}  \\
 \hline
  \end{tabular}
  \label{tab:fully_rgb_sod_expeiments}
\end{table*}

\begin{figure}[tp]
%  \vspace{-5mm}
  \begin{center}
  \setlength\tabcolsep{2pt}
  \begin{tabular}{*{2}{p{0.15\linewidth}<{\centering}} | *{2}{p{0.15\linewidth}<{\centering}} | *{2}{p{0.15\linewidth}<{\centering}}}
  % \begin{tabular}{c@{ } c@{ } c@{ } c@{ } c@{ } c@{ } }
  {\includegraphics[width=\linewidth]{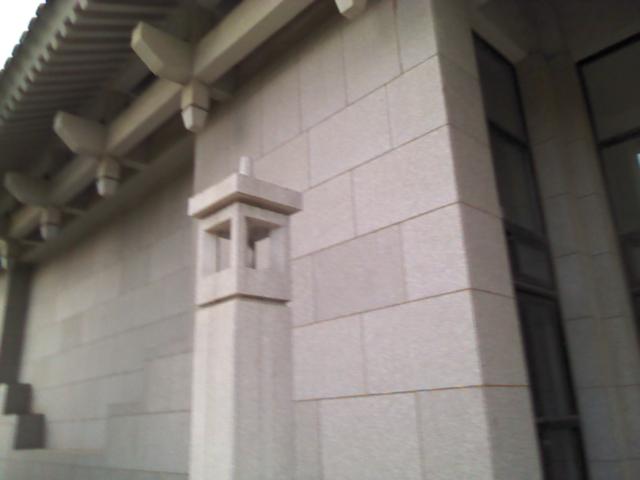}} &
  {\includegraphics[width=\linewidth]{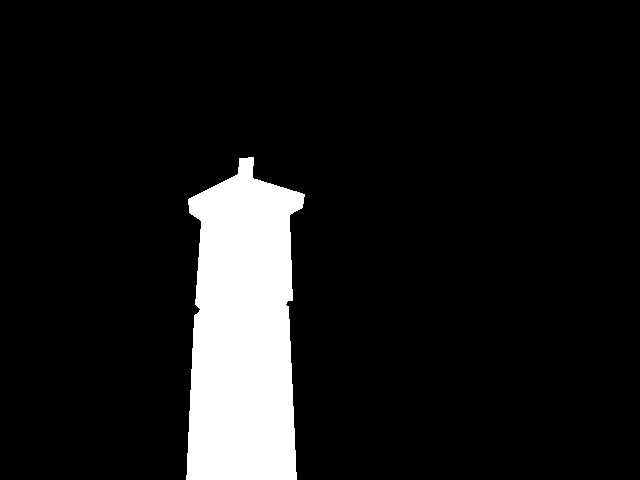}} &
  {\includegraphics[width=\linewidth]{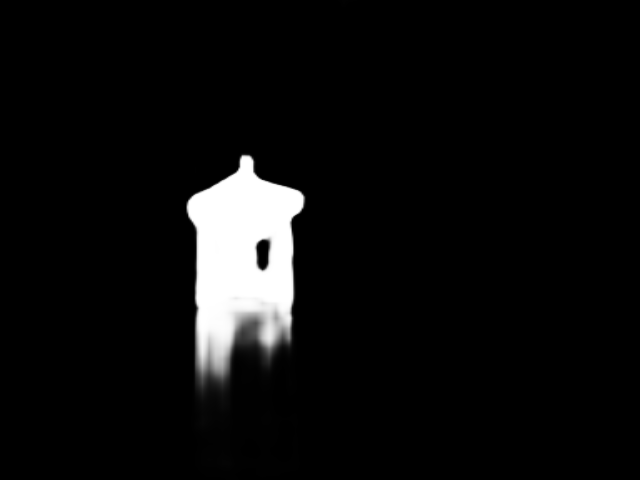}} &
  {\includegraphics[width=\linewidth]{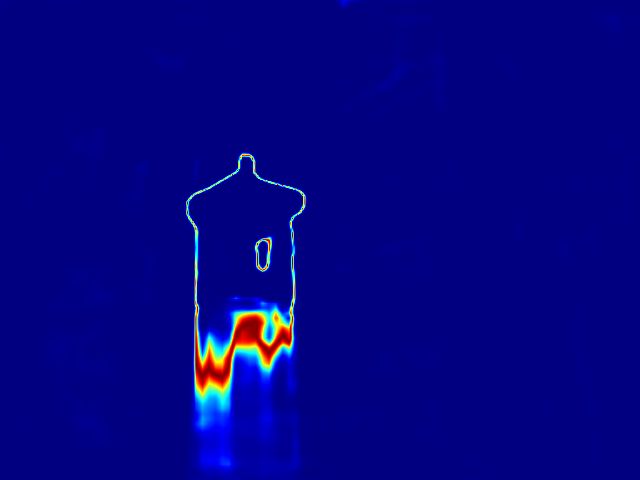}} &
  {\includegraphics[width=\linewidth]{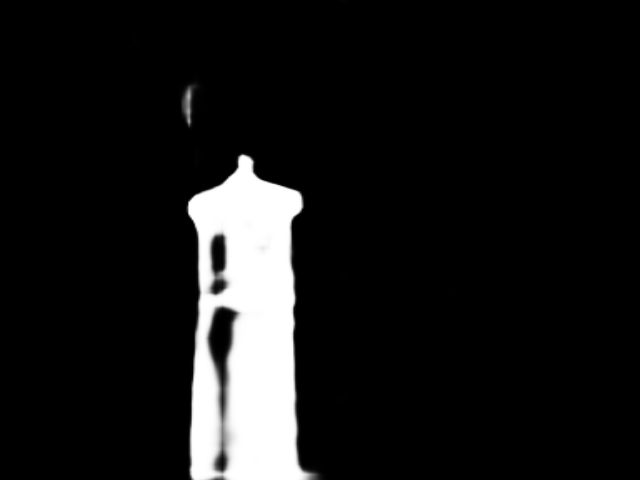}} &
  {\includegraphics[width=\linewidth]{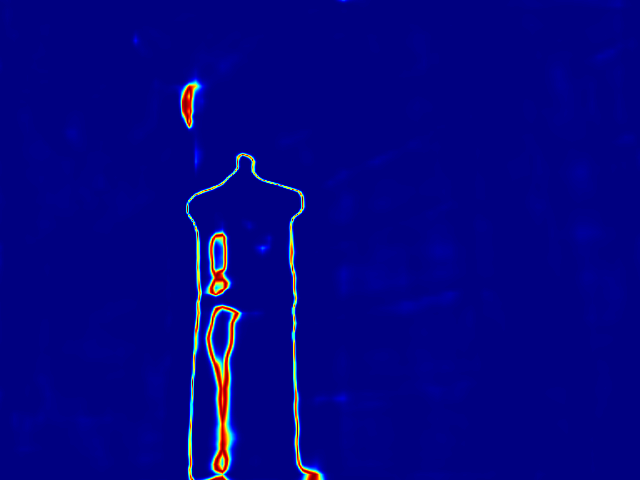}}\\
  % {\includegraphics[width=\linewidth]{fig19/25_img.jpg}}  &
  % {\includegraphics[width=\linewidth]{fig19/25_gt.png}}  &
  % {\includegraphics[width=\linewidth]{fig19/25_gtsod.png}} &
  % {\includegraphics[width=\linewidth]{fig19/25_gtsod_uncer.png}} &
  % {\includegraphics[width=\linewidth]{fig19/25_igan.png}} &
  % {\includegraphics[width=\linewidth]{fig19/25_igan_uncer.png}}\\
  {\includegraphics[width=\linewidth]{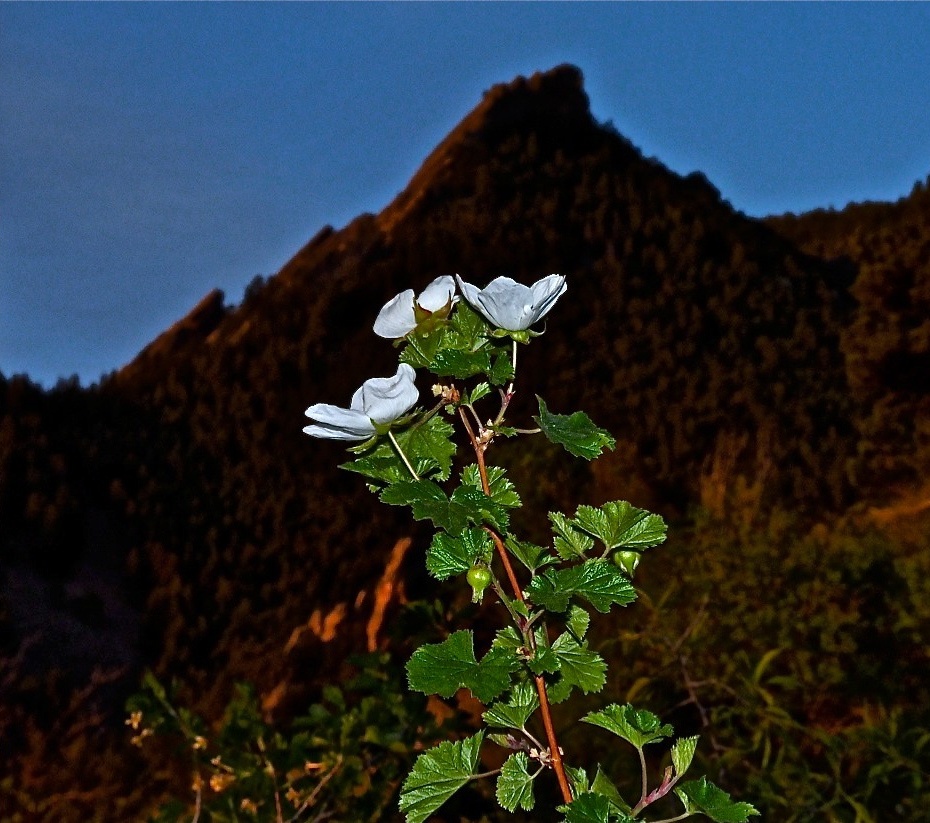}}  &
  {\includegraphics[width=\linewidth]{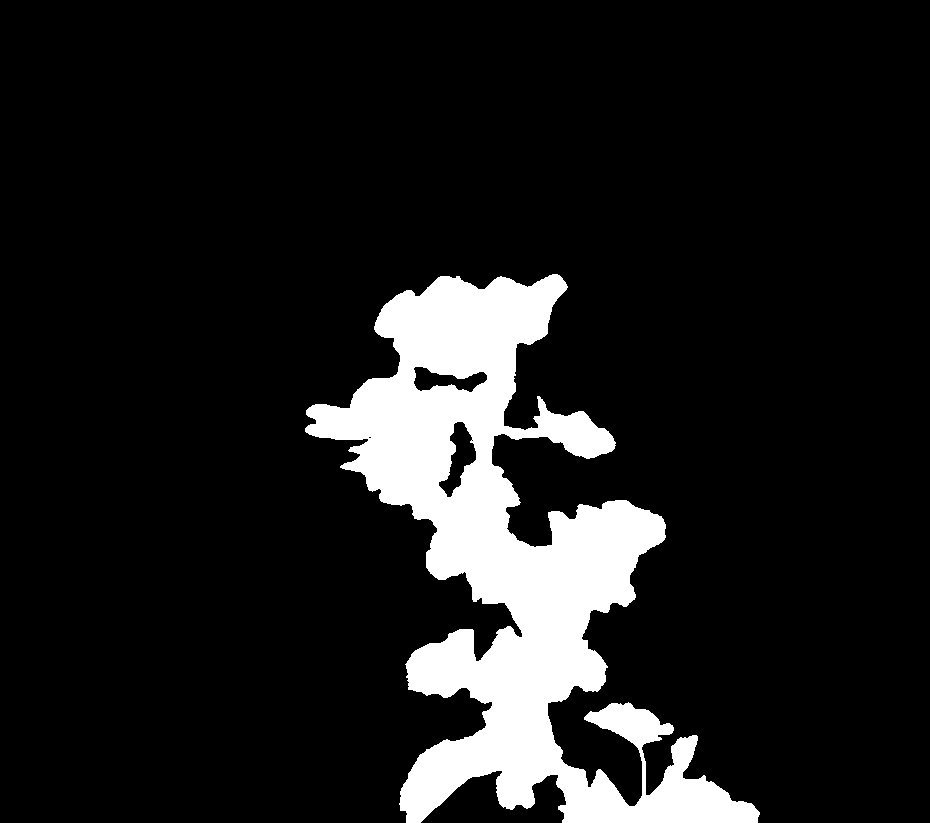}}   &
  {\includegraphics[width=\linewidth]{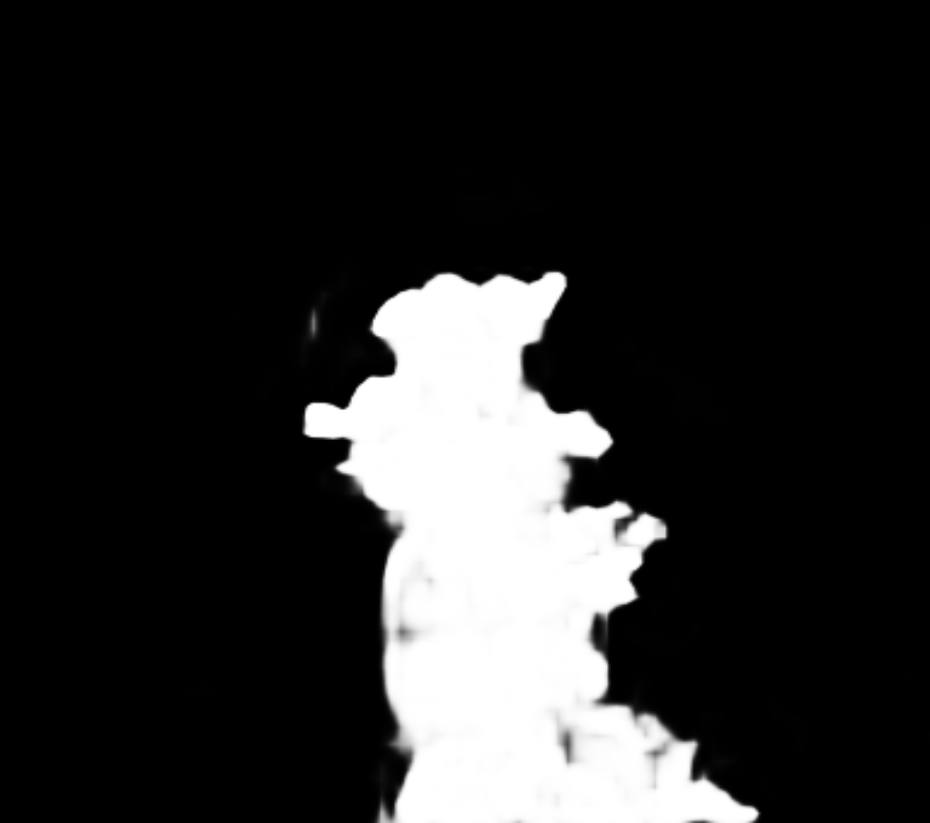}} &
  {\includegraphics[width=\linewidth]{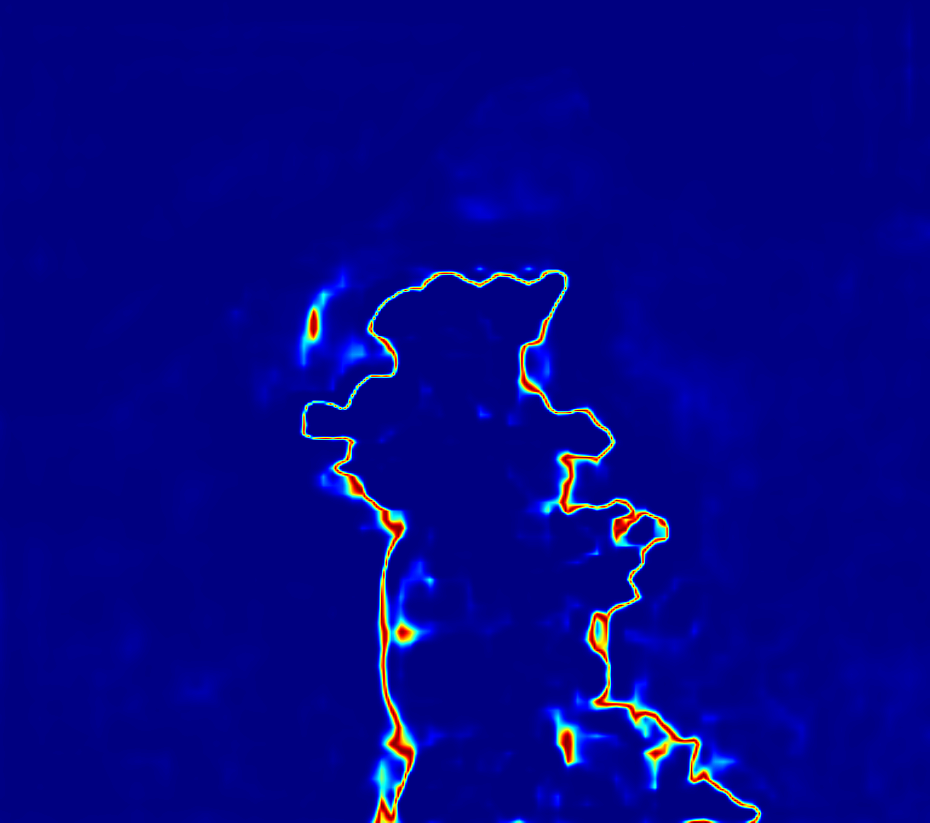}} &
  {\includegraphics[width=\linewidth]{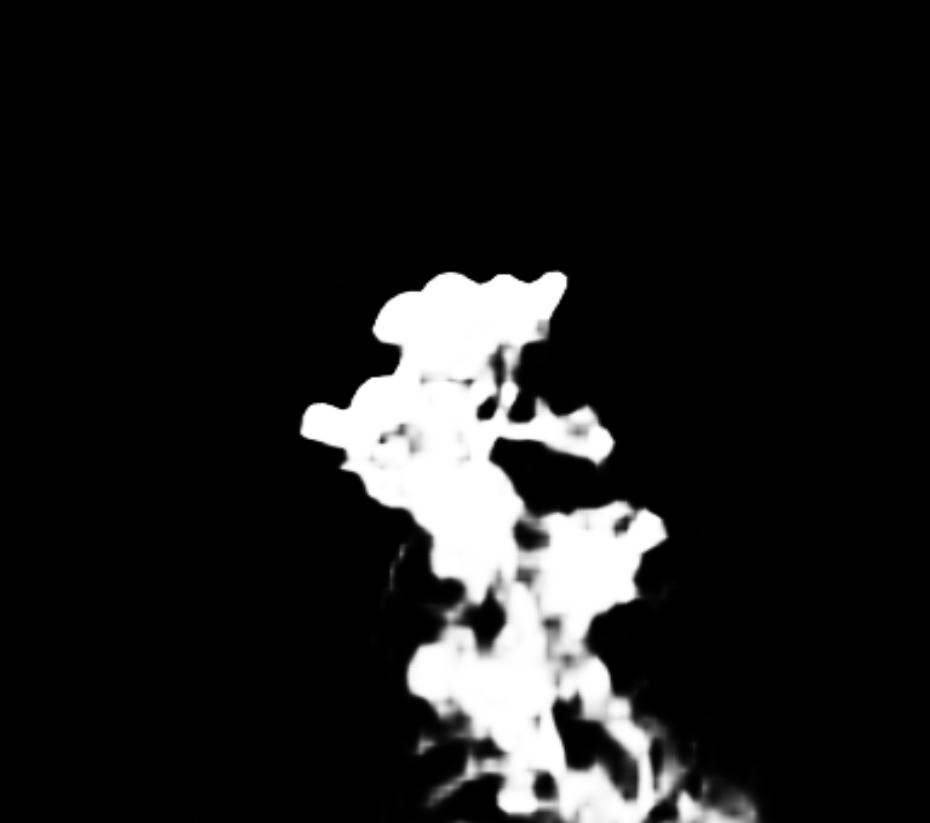}} &
  {\includegraphics[width=\linewidth]{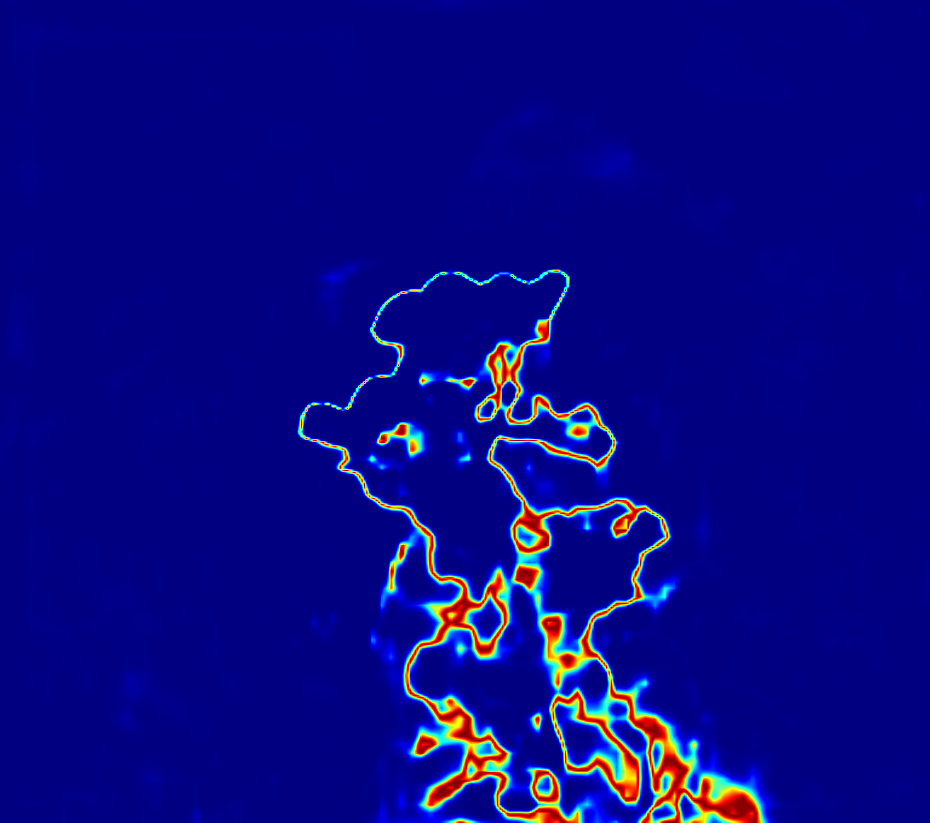}}\\
  \multicolumn{1}{c}{\footnotesize{Image}}&\multicolumn{1}{c}{\footnotesize{GT}}&\multicolumn{2}{c}{\footnotesize{GTSOD \cite{zhang2021learning_nips}}}&\multicolumn{2}{c}{\footnotesize{$\text{TIGAN}$}}\\
  \end{tabular}
  \end{center}
%   \vspace{-5pt}
  \caption{\footnotesize{Performance comparison with existing generative SOD model GTSOD \cite{zhang2021learning_nips}, where prediction in each block is the model prediction and the corresponding uncertainty map.}
  }
\label{fig:generative_performance_comparison}
\end{figure}
\begin{table}[t!]
  \centering
  \scriptsize
  \renewcommand{\arraystretch}{1.2}
  \renewcommand{\tabcolsep}{2.1mm}
  \caption{\footnotesize{\Rev{Calibration degree of conventional uncertainty estimation techniques 
  % \ie~MC-dropout~\cite{gal2016dropout}, 
  and the proposed inferential GAN with transformer backbone.}
  }}
  \begin{tabular}{l|cccccc}
  \hline
   & DUTS &ECSSD &DUT &HKU-IS &PASCAL-S &SOD     \\
   & \cite{imagesaliency} &\cite{yan2013hierarchical} &\cite{Manifold-Ranking:CVPR-2013} &\cite{li2015visual} &\cite{pascal_s_dataset} &\cite{sod_dataset}     \\
  \hline
    GTSOD~\cite{zhang2021learning_nips} &.0388 &.0214 &.0461 &.0231 &\underline{.0476} &.0552     \\
    MCD   &.0391 &.0230 &.0468 &.0234 &.0498 &.0581    \\
    TGAN  &.0382 &.0227 &.0464 &.0227 &.0488 &.0570    \\
    TCVAE &.0377 &.0220 &.0441 &.0229 &.0497 &.0560    \\
    TABP  &\underline{.0365} &\underline{.0204} &\underline{.0430} &\underline{.0211} &.0487 &\underline{.0545}    \\
    TIGAN &\textbf{.0353} &\textbf{.0198} &\textbf{.0400} &\textbf{.0198} &\textbf{.0467} &\textbf{.0519}    \\
   \hline 
% \bottomrule
  \end{tabular}
  \label{tab:mc_dropout_analysis}
%   \vspace{-5mm}
\end{table}

\begin{table*}[t!]
  \centering
  \scriptsize
  \renewcommand{\arraystretch}{1.2}
  \renewcommand{\tabcolsep}{0.705mm}
  \caption{\footnotesize{Depth contribution analysis for RGB-D SOD. Given the RGB image based model (\enquote{*\_RGB}), we first adapt it for RGB-D saliency detection with early-fusion (\enquote{*Early}). Then, the auxiliary depth module is attached to \enquote{*Early} to analyze the depth contribution.
%   .  where we conduct experiments  \YC{Detailed caption is needed.}
  }}
  \begin{tabular}{l|cccc|cccc|cccc|cccc|cccc|cccc}
  \hline
% \toprule
  &\multicolumn{4}{c|}{NJU2K \cite{NJU2000}}&\multicolumn{4}{c|}{SSB \cite{niu2012leveraging}}&\multicolumn{4}{c|}{DES \cite{cheng2014depth}}&\multicolumn{4}{c|}{NLPR \cite{peng2014rgbd}}&\multicolumn{4}{c|}{LFSD \cite{li2014saliency}}&\multicolumn{4}{c}{SIP \cite{sip_dataset}} \\
    Method & $S_{\alpha}\uparrow$&$F_{\beta}\uparrow$&$E_{\xi}\uparrow$&$\mathcal{M}\downarrow$& $S_{\alpha}\uparrow$&$F_{\beta}\uparrow$&$E_{\xi}\uparrow$&$\mathcal{M}\downarrow$& $S_{\alpha}\uparrow$&$F_{\beta}\uparrow$&$E_{\xi}\uparrow$&$\mathcal{M}\downarrow$& $S_{\alpha}\uparrow$&$F_{\beta}\uparrow$&$E_{\xi}\uparrow$&$\mathcal{M}\downarrow$& $S_{\alpha}\uparrow$&$F_{\beta}\uparrow$&$E_{\xi}\uparrow$&$\mathcal{M}\downarrow$& $S_{\alpha}\uparrow$&$F_{\beta}\uparrow$&$E_{\xi}\uparrow$&$\mathcal{M}\downarrow$ \\ \hline
    $\text{CB\_{RGB}}$ &.906 &.891 &.936 &.038 &.903 &.882 &.933 &.039 &.908 &.890 &.936 &.025 &.916 &.891 &.948 &.025 &.802 &.777 &.833 &.105 &.874 &.862 &.913 &.052  \\
    $\text{CEarly}$ &.912 &.901 &.941 &.036 &.903 &.881 &.934 &.038 &.932 &.921 &.964 &.018 &.914 &.886 &.945 &.026 &.830 &.804 &.858 &.082 &.878 &.864 &.914 &.050  \\
    $\text{CADE}$ &.911 &.902 &.939 &.036 &.902 &.877 &.936 &.039 &.935 &.922 &.968 &.018 &.922 &.896 &.951 &.025 &.860 &.848 &.899 &.069 &.888 &.880 &.925 &.045  \\ \hline
    % $\text{CADE}$ &.905 &.889 &.935 &.040 &.904 &.881 &.935 &.039 &.908 &.889 &.935 &.026 &.917 &.891 &.952 &.025 &.805 &.782 &.837 &.100 &.877 &.861 &.919 &.050  \\ \hline
   $\text{TB\_{RGB}}$ &\underline{.924} &\textbf{.919} &\textbf{.955} &\underline{.029}  &\textbf{.922} &\textbf{.906} &\textbf{.952} &\textbf{.030} &.918 &.908 &.943 &.022 &.931 &\underline{.914} &\underline{.962} &.021 &.869 &.856 &.899 &.067 &.895 &\textbf{.898} &.935 &.042 \\
   $\text{TEarly}$ &\textbf{.925} &\underline{.917} &\textbf{.955} &\textbf{.028} &\underline{.911} &\underline{.890} &\underline{.948} &\underline{.033} &\underline{.938} &\underline{.926} &\underline{.974} &\underline{.016} &\textbf{.935} &\textbf{.916} &.865 &\underline{.019} &\underline{.875} &\underline{.862} &\underline{.903} &\underline{.060} &\underline{.897} &\underline{.895} &\underline{.938} &\underline{.039} \\
%   $\text{TADE}$ &.922 &.915 &.953 &.030 &.921 &.905 &.952 &.030 &.921 &.910 &.950 &.021 &.934 &.915 &.964 &.019 &.872 &.857 &.898 &.066 &.893 &.894 &.934 &.042  \\
   $\text{TADE}$ &\textbf{.925} &\underline{.917} &\underline{.953} &\underline{.029} &\underline{.911} &\underline{.890} &.946 &.034 &\textbf{.944} &\textbf{.930} &\textbf{.977} &\textbf{.015} &\underline{.934} &.913 &\textbf{.965} &\textbf{.018} &\textbf{.879} &\textbf{.869} &\textbf{.910} &\textbf{.056} &\textbf{.902} &\underline{.895} &\textbf{.939} &\textbf{.038}  \\
   \hline
% \bottomrule
  \end{tabular}
  \label{tab:rgbd_sod_analysis}
%   \vspace{-5mm}
\end{table*}

\begin{figure}[tp]
%  \vspace{-5mm}
  \begin{center}
  \begin{tabular}{c@{ } c@{ } c@{ }}
  {\includegraphics[width=0.40\linewidth]{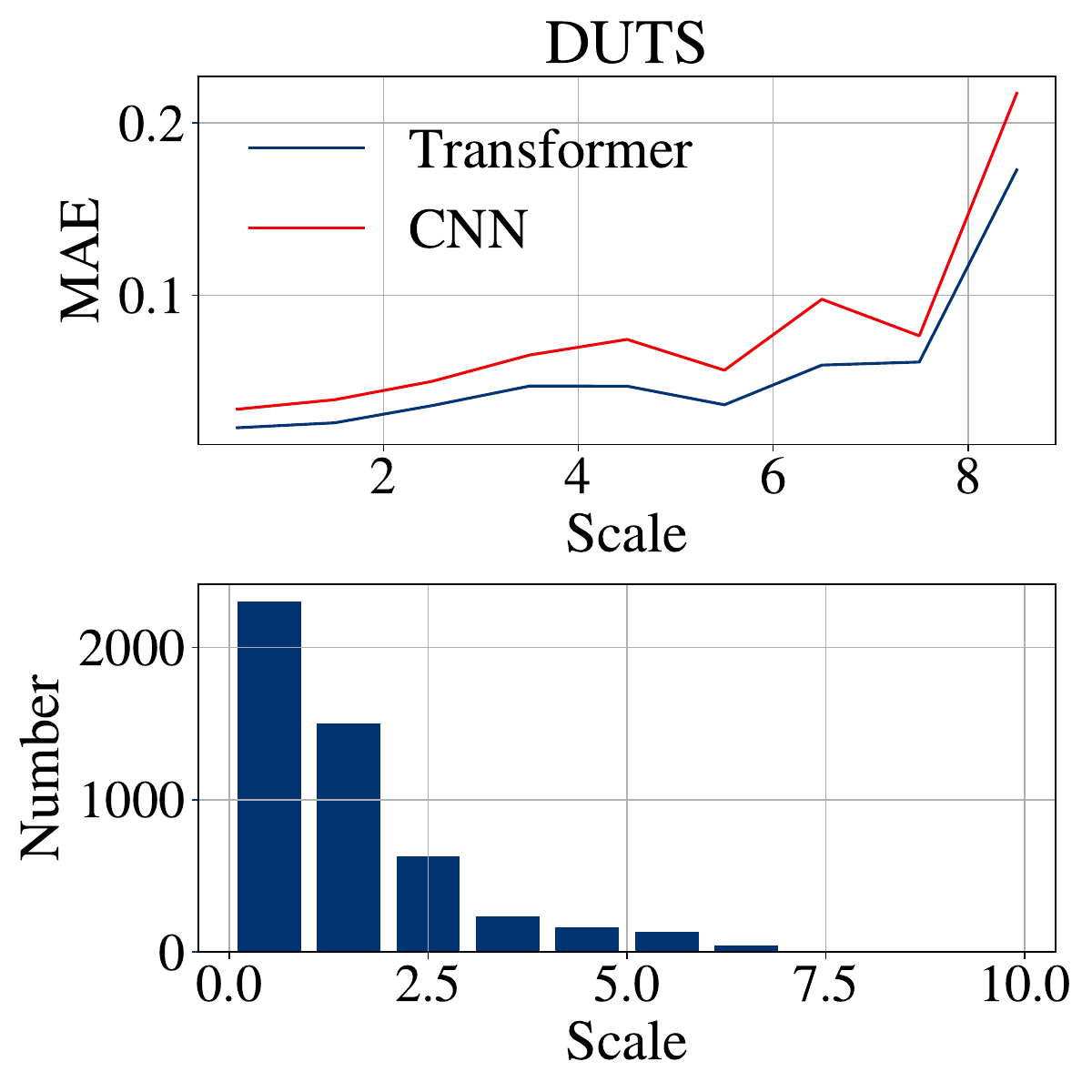}} &
  {\includegraphics[width=0.40\linewidth]{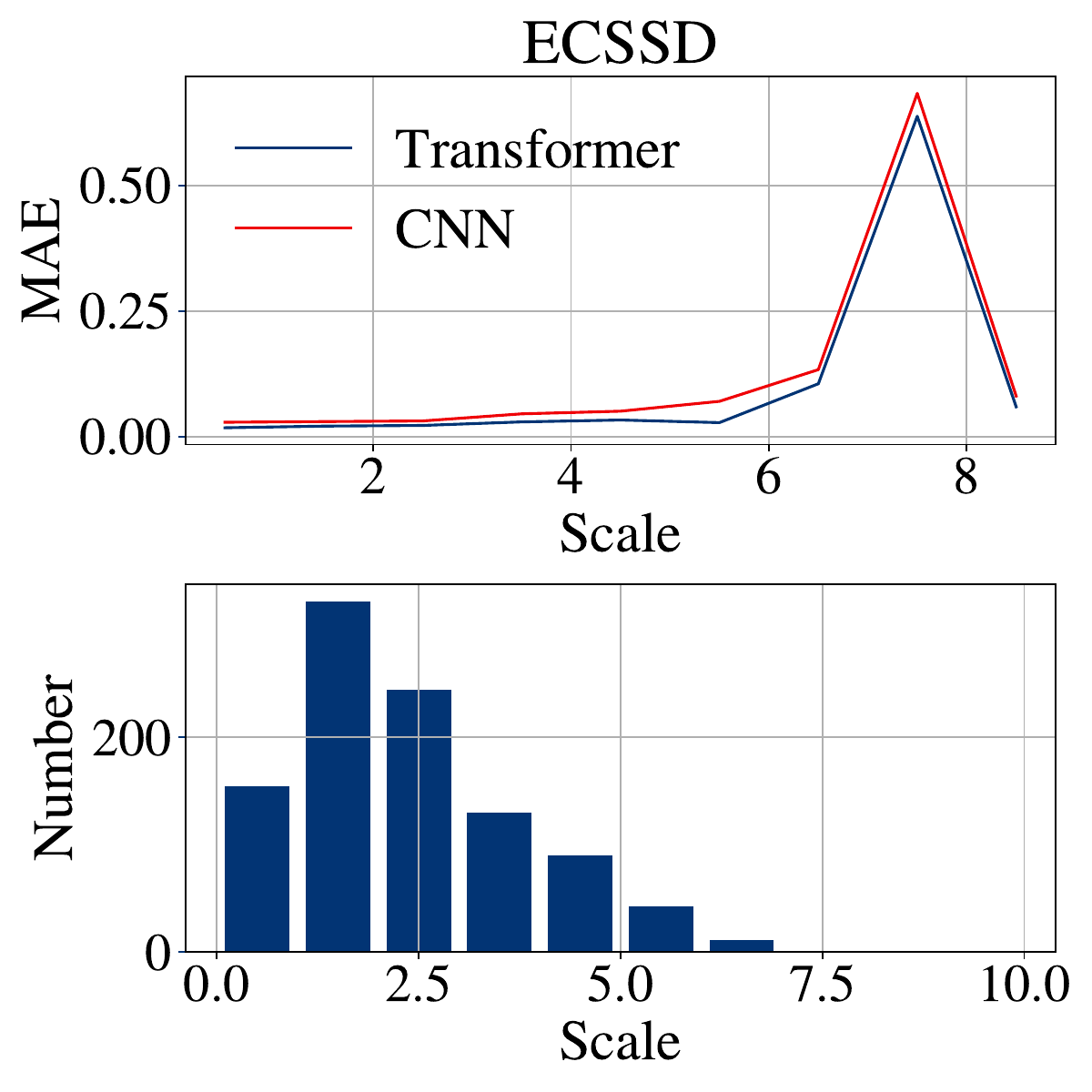}} \\
  {\includegraphics[width=0.40\linewidth]{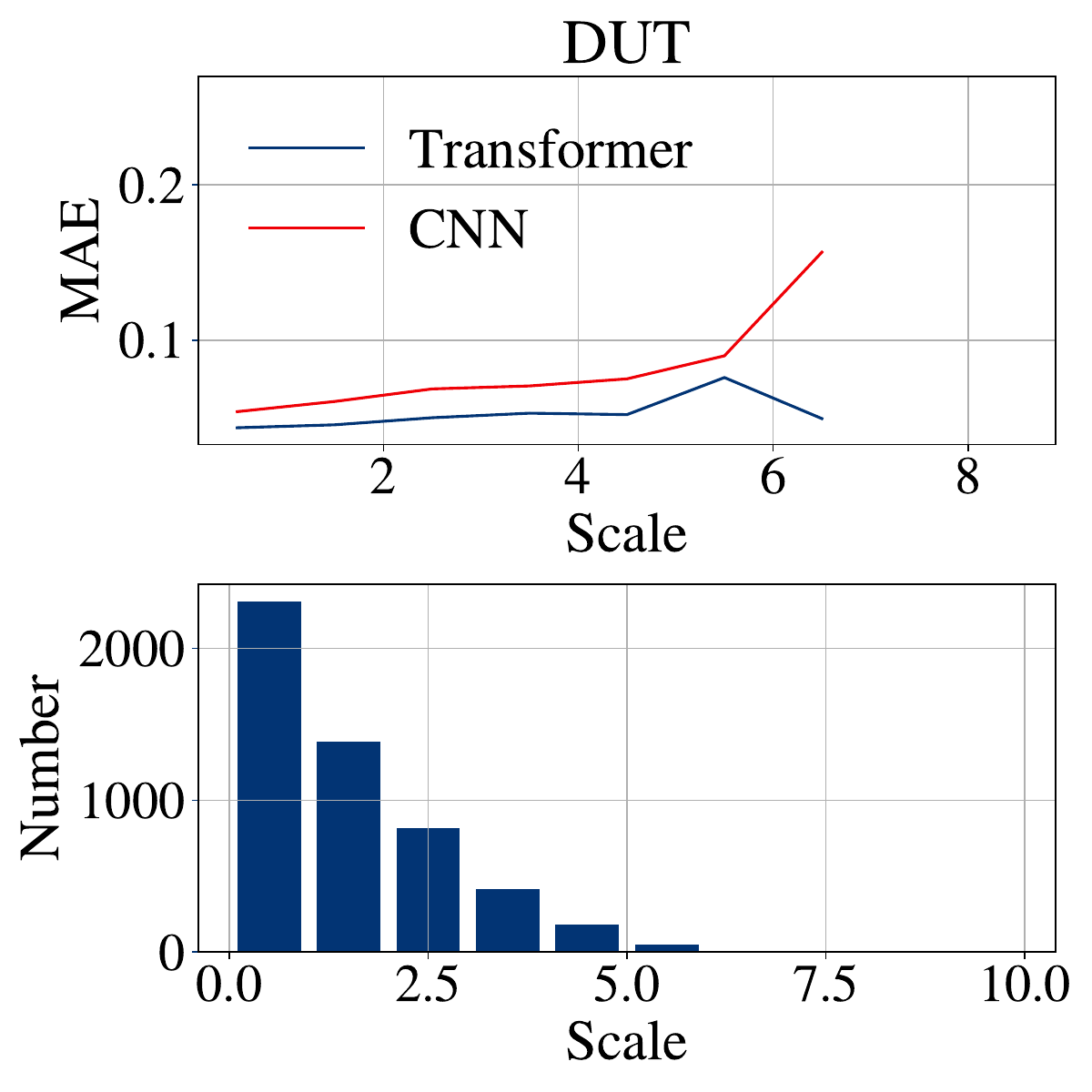}} &
  {\includegraphics[width=0.40\linewidth]{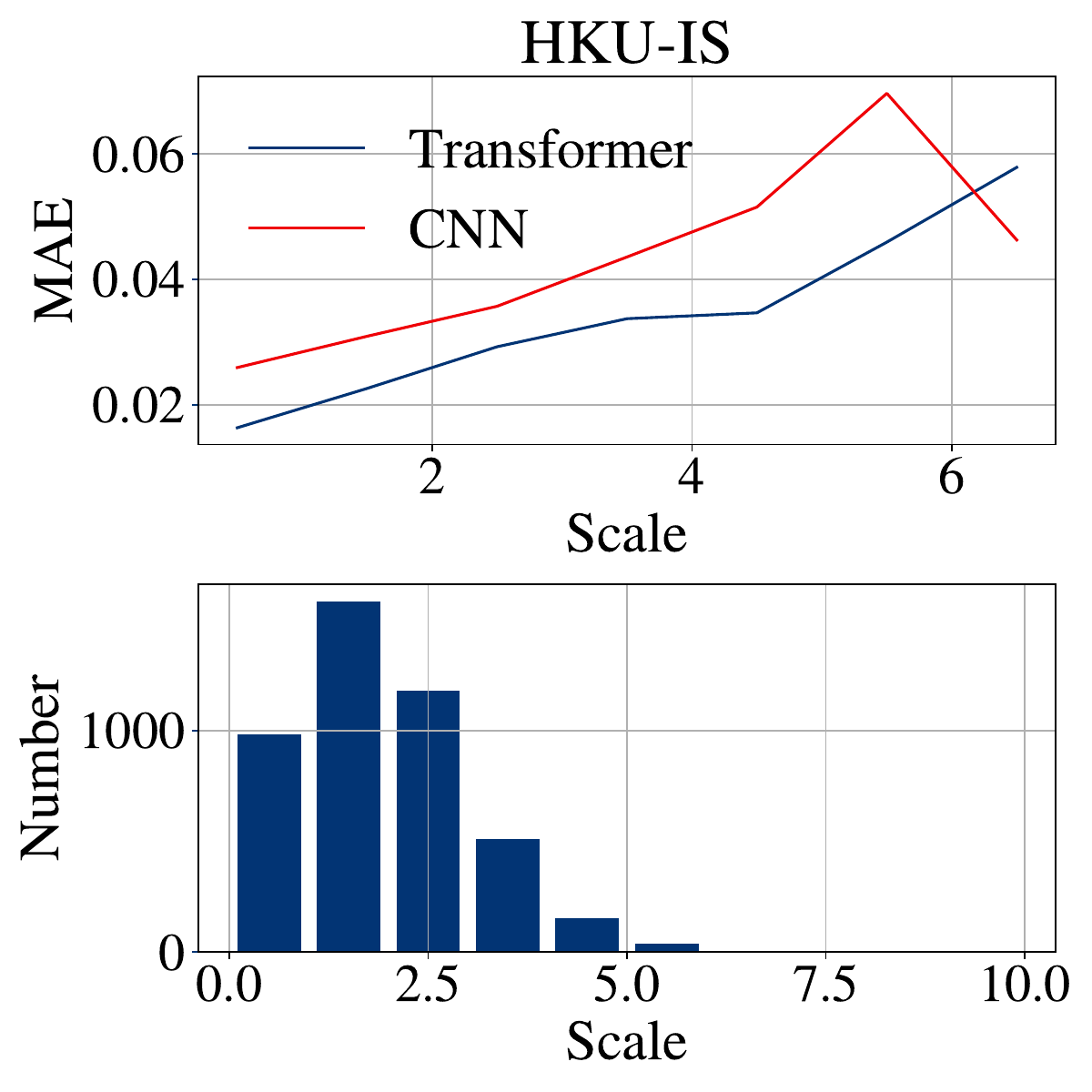}} \\
  {\includegraphics[width=0.40\linewidth]{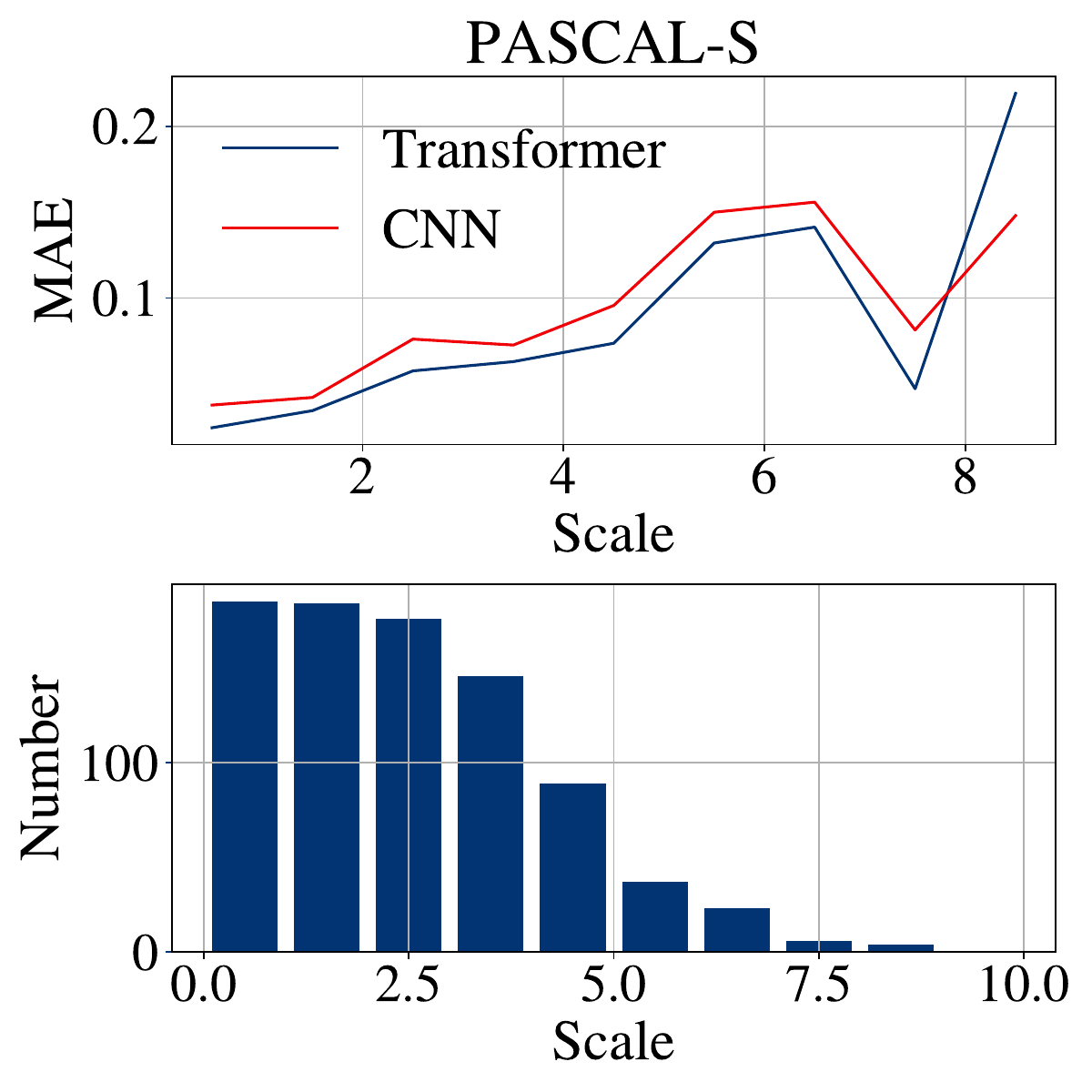}} &
  {\includegraphics[width=0.40\linewidth]{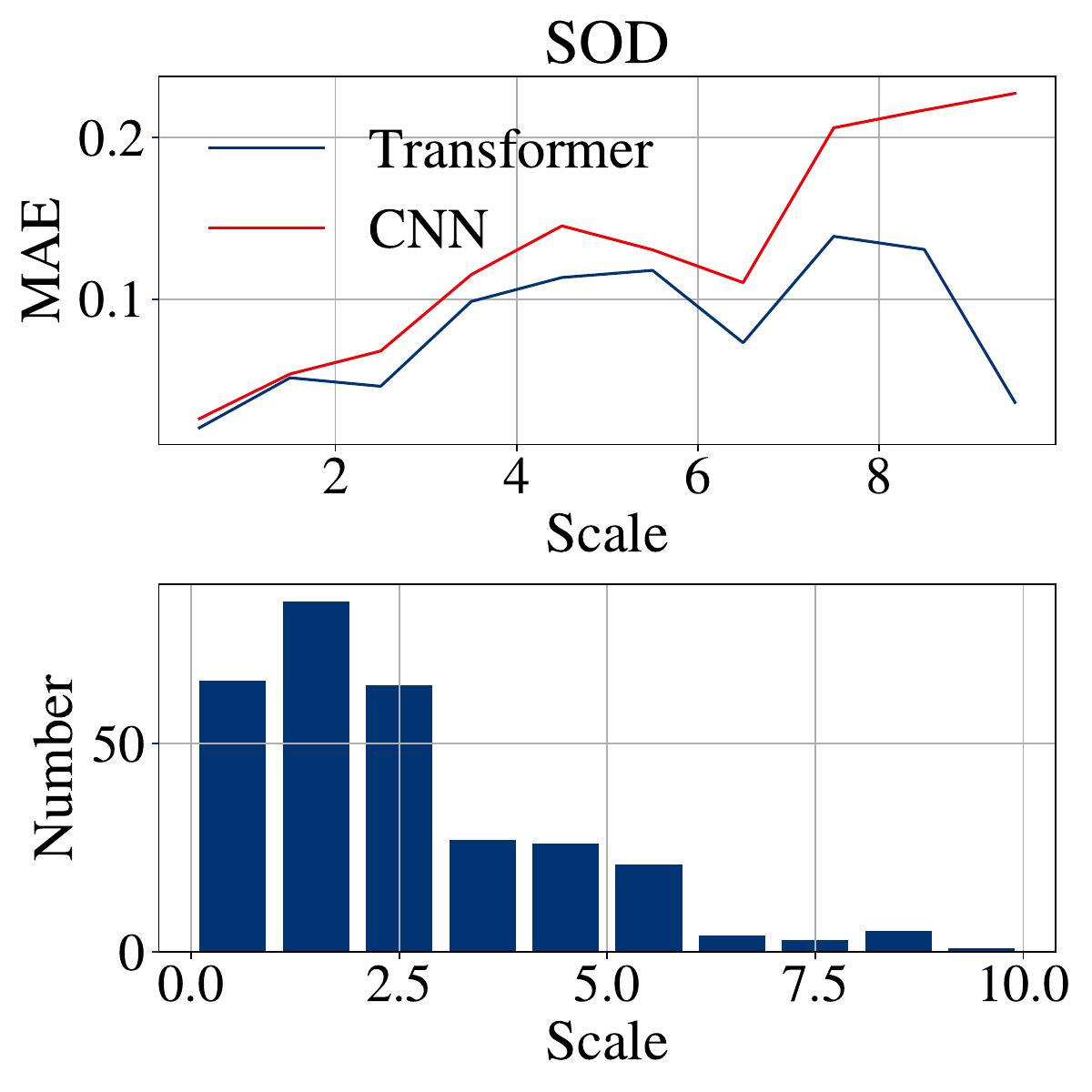}} \\
%   \footnotesize{DUT}&\footnotesize{DUTS}&\footnotesize{ECSSD}&\footnotesize{HKU-IS}&\footnotesize{PASCAL}&\footnotesize{SOD}
  \end{tabular}
  \end{center}
%   \vspace{-5pt}
  \caption{\footnotesize{Model ($\text{B'\_{cnn}}$ and $\text{B'\_{tr}}$ in Table \ref{tab:fully_rgb_sod_expeiments}) performance (the top curve of each block) \wrt~salient foreground size distribution (the bottom bar of each block) on six testing datasets.}
%   , where the top figure shows the performance of the CNN backbone-based and transformer backbone-based models, and the bottom figure shows the salient foreground size distribution.}
  }
\label{fig:performance_wrt_salient_object_size}
\end{figure}

% \subsubsection{Effectiveness of Transformer Backbone for Large Salient Object Detection}
\noindent\textbf{Transformer backbone for large salient object detection:} \Rev{As discussed, the larger receptive field of transformer backbone makes it ideal for the context based task, \ie~salient object detection. We then
% We want to identify samples where
% % \sout{that} \NB{where} 
% the transformer backbone outperforms the CNN backbone. To do so,
% % As an important factor for model performance, 
% we 
visualize the performance (we use mean absolute error (MAE) here as it's easy to implement and share the similar performance comparison trend with other measures)} of CNN backbone ($\text{B'\_{cnn}}$) and transformer backbone ($\text{B'\_{tr}}$) \wrt~size of the salient foreground in Fig.~\ref{fig:performance_wrt_salient_object_size}. Specifically, we uniformly group the scale of salient foreground to 10 bins and compute the mean performance of each backbone based model. Fig.~\ref{fig:performance_wrt_salient_object_size} shows that the transformer backbone based model outperforms the 
% \NB{the} 
CNN backbone based model almost consistently across all scales. Specifically, the performance gap for the largest foreground scale (on the
% \NB{the} 
DUT \cite{Manifold-Ranking:CVPR-2013} and SOD datasets \cite{sod_dataset}) is the most significant compared with other scales, which explains the superiority of the transformer for large salient object detection. There also exist scales for the HKU-IS~\cite{li2015visual} and PASCAL-S~\cite{pascal_s_dataset} datasets when the
% \NB{the} 
transformer backbone based model fails to outperform the CNN backbone based model, which are
% \NB{are} 
mainly due to the \enquote{double-edged sword} effect of the transformer's larger receptive field, which will be explained in Sec.~\ref{sec:model_discussion}. Images in ECSSD dataset~\cite{yan2013hierarchical} are relatively simpler compared with other datasets, and the foreground objects are distributed compactly around the image center, leading to less significant performance gain with the
% \NB{the} 
transformer backbone. 
% which clearly shows that the transformer backbone outperforms the CNN backbone mainly on images with larger salient foreground. 

% Further, we introduce auxiliary depth estimation module to solve the \enquote{distribution gap} issue within the existing RGB-D saliency detection datasets, and show model performance in Table \ref{tab:rgbd_sod_analysis}.

% \subsubsection{Depth Fusion Analysis}
% For the above experiments, we work on RGB saliency detection, and we then analyze how the transformer backbone works for RGB-D saliency detection. As we discussed before, the depth domain gap may hinder the performance of existing RGB-D saliency detection models. We then propose auxiliary depth estimation module to solve the \enquote{distribution gap} issue within the existing RGB-D saliency detection datasets, and show model performance in Table \ref{tab:rgbd_sod_analysis}. The auxiliary depth estimation module has the same structure as our saliency decoder. Within this framework, the final loss function has extra depth related loss: 
% \begin{equation}
%     \label{depth_loss}
%     \mathcal{L}_{depth}=\alpha(\beta*\mathcal{L}_{ssim}+(1-\beta)*\mathcal{L}_1),
% \end{equation}
% where $\alpha=0.1$ is used to control the contribution of the auxiliary depth estimation module, and following the conventional setting, we set $\beta=0.85$ in this paper. $\mathcal{L}_{ssim}$ is the SSIM loss function and $\mathcal{L}_1$ is the L1 loss.

% \subsubsection{Effectiveness of Auxiliary Depth Estimation}
\noindent\Rev{\textbf{Auxiliary depth module for depth domain gap modeling:}
We discuss this in Sec.~\ref{sec:intro} that the inconsistent depth distribution (see Table~\ref{tab:existing_rgbd_dataset} and Fig.~\ref{fig:depth_global_contrast}) leads to a domain gap between the training datasets and testing datasets for
% different levels of depth contribution for
RGB-D salient object detection. To solve the issue, we present an auxiliary depth module to fully explore the depth contribution and fix the depth domain gap issue via self-supervised learning. To evaluate the effectiveness of the proposed auxiliary depth module, we design three SOD models, namely the pure RGB image based model (\enquote{CB\_{RGB}} and \enquote{TB\_{RGB}}), early fusion model (\enquote{CEarly} and \enquote{TEarly}) and early fusion model with auxiliary depth module (\enquote{CADE} and \enquote{TADE}), where \enquote{C*} represents
% is the 
CNN backbone (ResNet50 \cite{resnet}),
% based model 
and \enquote{T*} is the transformer backbone.
% based model.
The performance of each model is shown in Table \ref{tab:rgbd_sod_analysis}.}
% We show model performance for RGB-D saliency detection in Table \ref{tab:rgbd_sod_analysis}, where \enquote{CB\_{RGB}}, \enquote{CEarly}, \enquote{CADE} are CNN-based models trained with RGB image only, the early fusion model and early fusion model with the auxiliary depth estimation module. \enquote{TB\_{RGB}}, \enquote{TEarly}, \enquote{TADE} are the corresponding transformer backbone counterparts. 

\begin{table*}[t!]
  \centering
  \scriptsize
  \renewcommand{\arraystretch}{1.1}
  \renewcommand{\tabcolsep}{0.735mm}
  \caption{\footnotesize{Reliable fully-supervised RGB SOD models, where we present performance of stochastic saliency prediction models via GAN, CVAE, ABP as well as the proposed iGAN. Performance of the baseline deterministic models ($\text{B'\_{cnn}}$ and $\text{B'\_{tr}}$ in Table \ref{tab:fully_rgb_sod_expeiments}) are listed for easier reference.
%   \YC{Detailed caption is needed.}
  }}
  \begin{tabular}{l|cccc|cccc|cccc|cccc|cccc|cccc}
  \hline
% \toprule
  &\multicolumn{4}{c|}{DUTS \cite{imagesaliency}}&\multicolumn{4}{c|}{ECSSD \cite{yan2013hierarchical}}&\multicolumn{4}{c|}{DUT \cite{Manifold-Ranking:CVPR-2013}}&\multicolumn{4}{c|}{HKU-IS \cite{li2015visual}}&\multicolumn{4}{c|}{PASCAL-S \cite{pascal_s_dataset}}&\multicolumn{4}{c}{SOD \cite{sod_dataset}} \\
    Method & $S_{\alpha}\uparrow$&$F_{\beta}\uparrow$&$E_{\xi}\uparrow$&$\mathcal{M}\downarrow$& $S_{\alpha}\uparrow$&$F_{\beta}\uparrow$&$E_{\xi}\uparrow$&$\mathcal{M}\downarrow$& $S_{\alpha}\uparrow$&$F_{\beta}\uparrow$&$E_{\xi}\uparrow$&$\mathcal{M}\downarrow$& $S_{\alpha}\uparrow$&$F_{\beta}\uparrow$&$E_{\xi}\uparrow$&$\mathcal{M}\downarrow$& $S_{\alpha}\uparrow$&$F_{\beta}\uparrow$&$E_{\xi}\uparrow$&$\mathcal{M}\downarrow$& $S_{\alpha}\uparrow$&$F_{\beta}\uparrow$&$E_{\xi}\uparrow$&$\mathcal{M}\downarrow$ \\ \hline
   $\text{CGAN}$ &.881 &.839 &.917 &.036 &.919 &.916 &.945 &.036 &.818 &.734 &.845 &.056 &.909 &.898 &.945 &.031 &.857 &.845 &.899 &.064 &.818 &.807 &.846 &.078   \\
   $\text{CCVAE}$ &.877 &.833 &.911 &.040 &.922 &.920 &.949 &.034 &.817 &.735 &.845 &.063 &.910 &.900 &.947 &.031 &.855 &.842 &.897 &.066 &.830 &.822 &.866 &.073  \\
   $\text{CABP}$ &.828 &.757 &.859 &.058 &.887 &.877 &.913 &.055 &.778 &.670 &.801 &.078 &.878 &.855 &.913 &.047 &.810 &.782 &.845 &.094 &.773 &.744 &.799 &.102   \\
   $\text{CIGAN}$ &.876 &.820 &.906 &.042 &.923 &.913 &.945 &.037 &.823 &.733 &.848 &.061 &.911 &.892 &.943 &.034 &.856 &.836 &.893 &.068 &.833 &.816 &.862 &.075   \\ 
   $\text{B'\_{cnn}}$ &.882 &.840 &.916 &.037 &.922 &.919 &.947 &.035 &.823 &.742 &.851 &.057 &.912 &.901 &.947 &.030 &.855 &.841 &.896 &.065 &.832 &.825 &.863 &.073  \\ \hline
   $\text{TGAN}$ &.907 &.877 &.944 &.029 &.939 &\underline{.938} &.964 &\underline{.025} &.852 &.789 &.882 &.051 &\underline{.927} &.920 &\underline{.963} &\underline{.024} &\underline{.878} &\textbf{.872} &\textbf{.918} &\textbf{.053} &.855 &.849 &.894 &.061   \\
   $\text{TCVAE}$ &.908 &\underline{.879} &\underline{.945} &\underline{.028} &.940 &\textbf{.940} &\textbf{.966} &\textbf{.024} &.857 &.796 &.890 &.048 &\underline{.927} &\textbf{.922} &\textbf{.964} &\underline{.024} &.876 &\underline{.871} &\textbf{.918} &\underline{.054} &.858 &\underline{.854} &\textbf{.898} &\underline{.060}   \\
   $\text{TABP}$ &\underline{.910} &.878 &.944 &\underline{.028} &\textbf{.942} &\textbf{.940} &\textbf{.966} &\textbf{.024} &\underline{.860} &\underline{.799} &\underline{.891} &.048 &\textbf{.929} &\textbf{.922} &\textbf{.964} &\underline{.024} &\textbf{.879} &.870 &\textbf{.918} &\underline{.054} &\underline{.860} &\textbf{.858} &\underline{.897} &.061   \\
   $\text{TIGAN}$ &.909 &.873 &.941 &\underline{.028} &\underline{.941} &.936 &.964 &\underline{.025} &\textbf{.861} &.796 &.890 &\underline{.047} &\textbf{.929} &.918 &.962 &.025 &\textbf{.879} &.869 &.916 &\underline{.054} &\textbf{.861} &\underline{.854} &.894 &\underline{.060}   \\
   $\text{B'\_{tr}}$ &\textbf{.911} &\textbf{.882} &\textbf{.947} &\textbf{.026} &.939 &\textbf{.940} &\underline{.965} &\textbf{.024} &\underline{.860} &\textbf{.801} &\textbf{.894} &\textbf{.045} &\underline{.927} &\underline{.921} &\textbf{.964} &\textbf{.023} &.876 &\textbf{.872} &\underline{.917} &\textbf{.053} &.858 &.853 &\underline{.897} &\textbf{.059} \\
   \hline
  \end{tabular}
  \label{tab:reliable_rgb_sod}
\end{table*}

\begin{table*}[t!]
  \centering
  \scriptsize
  \renewcommand{\arraystretch}{1.2}
  \renewcommand{\tabcolsep}{0.7351mm}
  \caption{\footnotesize{Reliable fully-supervised RGB-D SOD models, where we present performance of stochastic saliency prediction models via GAN, CVAE, ABP as well as the proposed iGAN. Performance of the baseline models ($\text{CADE}$ and $\text{TADE}$ in Table \ref{tab:rgbd_sod_analysis}) are listed for easier reference.
%   \YC{Detailed caption is needed.}
  }}
  \begin{tabular}{l|cccc|cccc|cccc|cccc|cccc|cccc}
  \hline
% \toprule
  &\multicolumn{4}{c|}{NJU2K \cite{NJU2000}}&\multicolumn{4}{c|}{SSB \cite{niu2012leveraging}}&\multicolumn{4}{c|}{DES \cite{cheng2014depth}}&\multicolumn{4}{c|}{NLPR \cite{peng2014rgbd}}&\multicolumn{4}{c|}{LFSD \cite{li2014saliency}}&\multicolumn{4}{c}{SIP \cite{sip_dataset}} \\
    Method & $S_{\alpha}\uparrow$&$F_{\beta}\uparrow$&$E_{\xi}\uparrow$&$\mathcal{M}\downarrow$& $S_{\alpha}\uparrow$&$F_{\beta}\uparrow$&$E_{\xi}\uparrow$&$\mathcal{M}\downarrow$& $S_{\alpha}\uparrow$&$F_{\beta}\uparrow$&$E_{\xi}\uparrow$&$\mathcal{M}\downarrow$& $S_{\alpha}\uparrow$&$F_{\beta}\uparrow$&$E_{\xi}\uparrow$&$\mathcal{M}\downarrow$& $S_{\alpha}\uparrow$&$F_{\beta}\uparrow$&$E_{\xi}\uparrow$&$\mathcal{M}\downarrow$& $S_{\alpha}\uparrow$&$F_{\beta}\uparrow$&$E_{\xi}\uparrow$&$\mathcal{M}\downarrow$ \\ \hline
    $\text{CGAN}$ &.914 &.905 &.943 &.035 &.904 &.881 &.937 &.039 &.929 &.917 &.957 &.019 &.924 &.899 &.954 &.023 &.849 &.826 &.884 &.074 &.885 &.875 &.921 &.047   \\
    $\text{CCVAE}$ &.906 &.894 &.937 &.039 &.896 &.871 &.934 &.041 &.940 &.923 &.975 &.017 &.916 &.891 &.951 &.026 &.841 &.825 &.881 &.075 &.887 &.878 &.927 &.045   \\
   $\text{CABP}$ &.916 &.903 &.941 &.034 &.905 &.878 &.935 &.039 &.941 &.928 &.972 &.017 &.921 &.891 &.949 &.025 &.845&.828 &.876&.077 &.888 &.876 &.922 &.046 \\
   $\text{CIGAN}$ &.914 &.900 &.939 &.036 &.903 &.876 &.934 &.040 &.937 &.921 &.970 &.018 &.922 &.890 &.952 &.025 &.851 &.832 &.889 &.075 &.884 &.870 &.917 &.049   \\ 
%   $\text{CGAN}$ &.909 &.896 &.938 &.038 &.903 &.880 &.934 &.040 &.931 &.919 &.963 &.019 &.914 &.889 &.948 &.026 &.829 &.809 &.860 &.084 &.880 &.867 &.917 &.050   \\
%   $\text{CABP}$ &.907 &.891 &.934 &.040 &.904 &.877 &.931 &.041 &.925 &.907 &.949 &.022 &.913 &.881 &.942 &.027 &.824 &.795 &.847 &.089 &.872 &.852 &.906 &.056  \\
%   $\text{CIGAN}$ &.910 &.892 &.934 &.040 &.904 &.874 &.930 &.042 &.932 &.913 &.960 &.021 &.911 &.874 &.940 &.029 &.832 &.806 &.858 &.087 &.876 &.855 &.910 &.054   \\ 
   $\text{CADE}$ &.911 &.902 &.939 &.036 &.902 &.877 &.936 &.039 &.935 &.922 &.968 &.018 &.922 &.896 &.951 &.025 &.860 &.848 &.899 &.069 &.888 &.880 &.925 &.045  \\ \hline
   $\text{TGAN}$ &\textbf{.928} &\underline{.921} &\textbf{.956} &\textbf{.027} &.911 &.890 &\underline{.946} &\textbf{.034} &.941 &\textbf{.931} &\underline{.975} &\textbf{.015} &\underline{.934} &\textbf{.915} &\underline{.964} &\underline{.019} &.875 &.860 &.903 &.060 &.900 &\underline{.901} &.939 &\underline{.038}   \\
   $\text{TCVAE}$ &\textbf{.928} &\textbf{.922} &\textbf{.956} &\underline{.028} &.911 &.889 &.944 &\underline{.035} &.941 &.929 &.973 &\underline{.016} &\textbf{.935} &\textbf{.915} &\underline{.964} &.020 &\underline{.879} &.863 &.905 &.060 &.903 &\textbf{.904} &\underline{.941} &\underline{.038}  \\
   $\text{TABP}$ &\underline{.927} &.917 &\underline{.954} &.029 &\underline{.913} &\underline{.891} &\textbf{.947} &\textbf{.034} &\underline{.943} &.929 &.974 &\textbf{.015} &.933 &.911 &.962 &.020 &.870 &.852 &.900 &.065 &\underline{.904} &.900 &\textbf{.942} &\textbf{.037}   \\
   $\text{TIGAN}$ &\textbf{.928} &.919 &\textbf{.956} &\underline{.028} &\textbf{.915} &\textbf{.893} &\textbf{.947} &\textbf{.034} &.940 &.929 &.970 &\underline{.016} &.932 &.911 &.961 &.020 &\textbf{.884} &\underline{.868} &\textbf{.911} &\underline{.057} &\textbf{.905} &\underline{.901} &\underline{.941} &\textbf{.037}  \\
%     $\text{TGAN}$ &.925 &.918 &.955 &.029 &.921 &.905 &.952 &.030 &.918 &.908 &.943 &.022 &.932 &.915 &.963 &.020 &.871 &.856 &.898 &.068 &.895 &.899 &.937 &.040   \\
%   $\text{TABP}$ &.921 &.908 &.948 &.032 &.910 &.884 &.942 &.037 &.936 &.916 &.966 &.018 &.927 &.904 &.957 &.022 &.888 &.878 &.917 &.055 &.895 &.894 &.929 &.043   \\
%   $\text{TIGAN}$ &.922 &.909 &.948 &.032 &.910 &.884 &.942 &.037 &.933 &.913 &.964 &.019 &.927 &.902 &.956 &.023 &.884 &.871 &.912 &.056 &.893 &.890 &.929 &.044   \\
   $\text{TADE}$ &.925 &.917 &.953 &.029 &.911 &.890 &\underline{.946} &\textbf{.034} &\textbf{.944} &\underline{.930} &\textbf{.977} &\textbf{.015} &\underline{.934} &\underline{.913} &\textbf{.965} &\textbf{.018} &\underline{.879} &\textbf{.869} &\underline{.910} &\textbf{.056} &.902 &.895 &.939 &\underline{.038}  \\
   \hline 
% \bottomrule
  \end{tabular}
  \label{tab:reliable_rgbd_sod}
%   \vspace{-5mm}
\end{table*}

\Rev{Firstly, we observe the improved performance of early fusion models compared with training only with RGB images, indicating the benefits of depth for SOD. Further, as shown in Fig.~\ref{fig:depth_global_contrast}, salient foreground depth contrast of the LFSD dataset \cite{li2014saliency} is most different from the corresponding RGB image salient foreground contrast. In this way, we claim that the auxiliary depth module should contribute the most, which is consistent with our experiments. Note that, the proposed auxiliary depth module aims to further explore the depth contribution, especially for depth data that distributes differently from the training dataset.
% shares more complementary information with the RGB image (see Section \ref{sec:intro} and Fig.~\ref{fig:depth_global_contrast}).
}

\subsubsection{Reliable Saliency Model}
% A reliable model should be aware of its prediction, and m
Model reliability is an important factor for measuring accountability for decisions before deployment, and a reliable model should be aware of its predictions. In this paper, we introduce the iGAN for reliable saliency detection with an image conditioned latent prior.
% to borrow the benefit of both GAN \cite{gan_raw} for it's accurate prediction and ABP \cite{ABP} for it's reliable prediction.
In addition to the proposed iGAN, we also design GAN-based \cite{gan_raw}, CVAE-based~\cite{VAE_Kingma,cvae} and ABP-based \cite{ABP} generative models for RGB SOD and RGB-D SOD. The performance is shown in Table \ref{tab:reliable_rgb_sod} and Table \ref{tab:reliable_rgbd_sod} respectively, where \enquote{CGAN}, \enquote{CCVAE}, \enquote{CABP} and \enquote{CIGAN} are the stochastic models
% convolutional neural network based model 
based on GAN, CVAE, ABP, and the proposed inferential GAN respectively with CNN backbone, and \enquote{TGAN}, \enquote{TCVAE}, \enquote{TABP} and \enquote{TIGAN} are the transformer counterparts. 
% Note that, for the CVAE \cite{cvae} based models, we follow the same prior and posterior distribution model designing as in~\cite{zhang2021_ucnet}.

\begin{figure*}[tp]
%  \vspace{-5mm}
  \begin{center}
  \setlength\tabcolsep{2pt}
  \begin{tabular}{*{2}{p{0.092\linewidth}<{\centering}} | *{2}{p{0.092\linewidth}<{\centering}} | *{2}{p{0.092\linewidth}<{\centering}} | *{2}{p{0.092\linewidth}<{\centering}} | *{2}{p{0.092\linewidth}<{\centering}}}
  % \begin{tabular}{c@{ } c@{ } c@{ } c@{ } c@{ } c@{ }  c@{ } c@{ } c@{ } c@{ }}
  {\includegraphics[width=\linewidth]{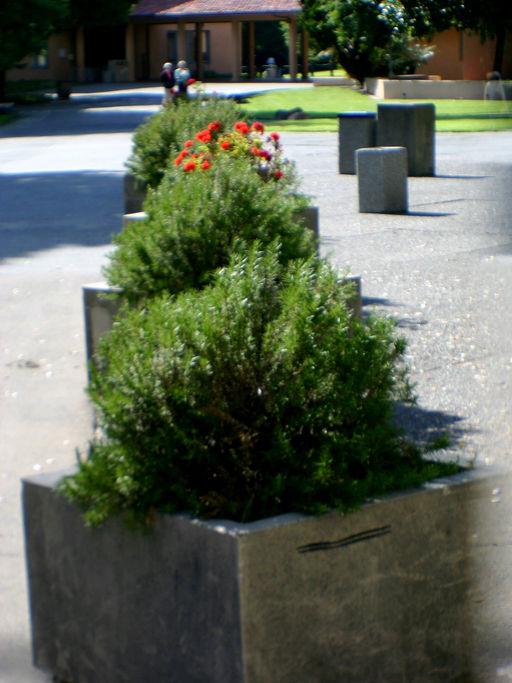}} &
  {\includegraphics[width=\linewidth]{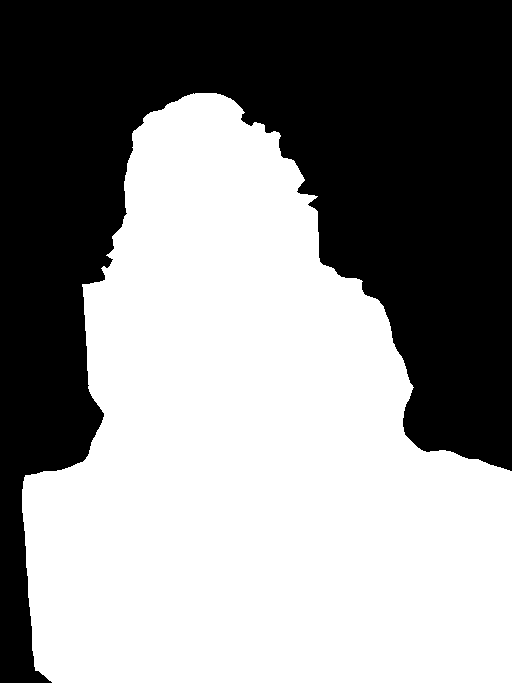}}  &
  {\includegraphics[width=\linewidth]{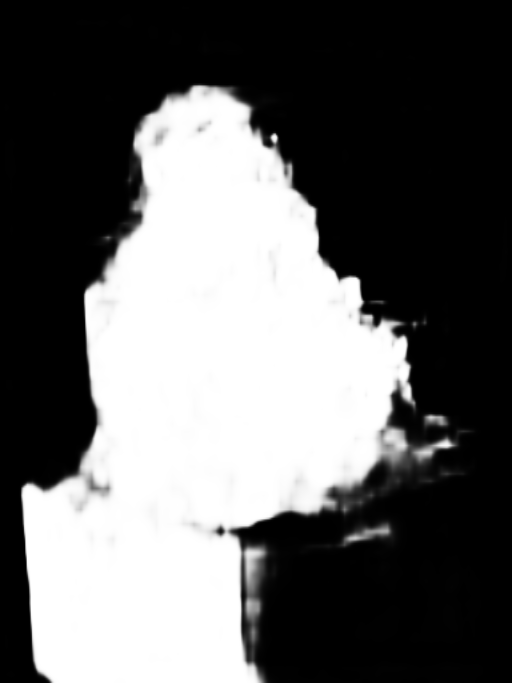}} &
  {\includegraphics[width=\linewidth]{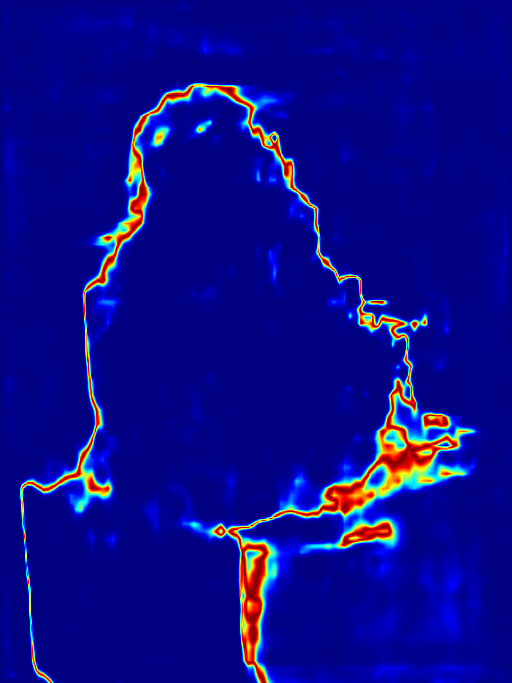}} &
  {\includegraphics[width=\linewidth]{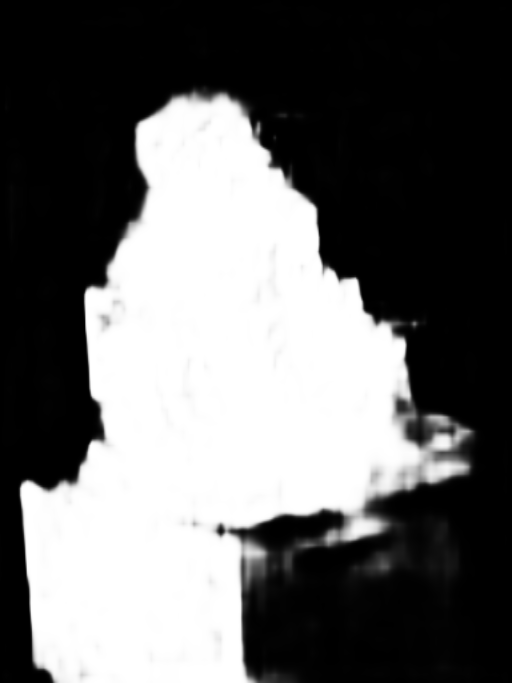}} &
  {\includegraphics[width=\linewidth]{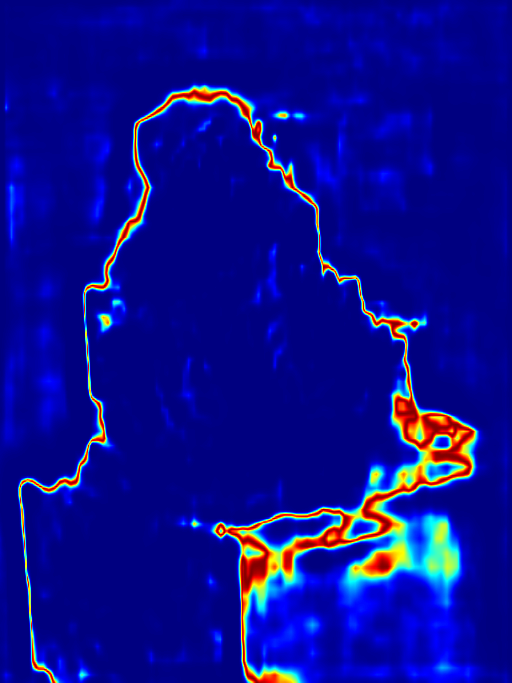}}  &
  {\includegraphics[width=\linewidth]{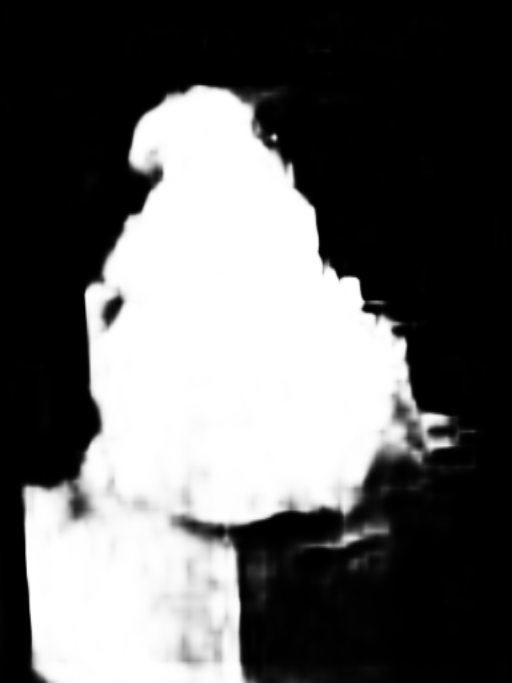}} &
  {\includegraphics[width=\linewidth]{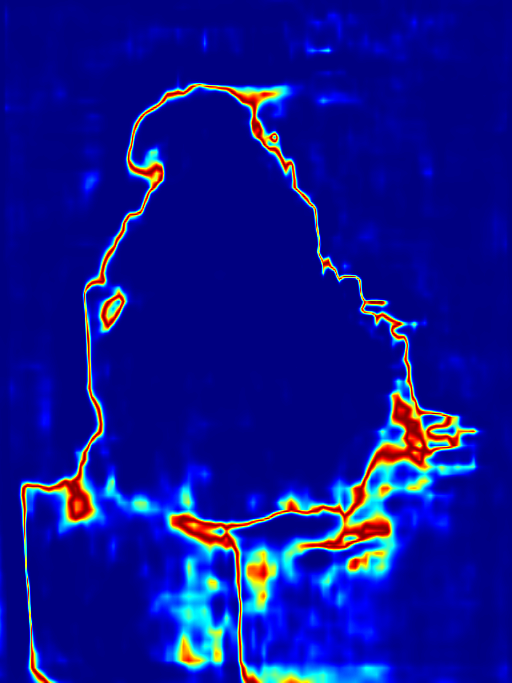}} &
  {\includegraphics[width=\linewidth]{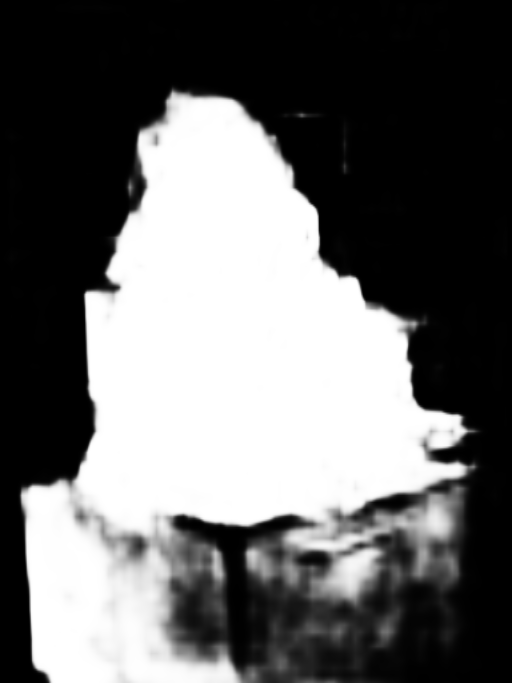}} &
  {\includegraphics[width=\linewidth]{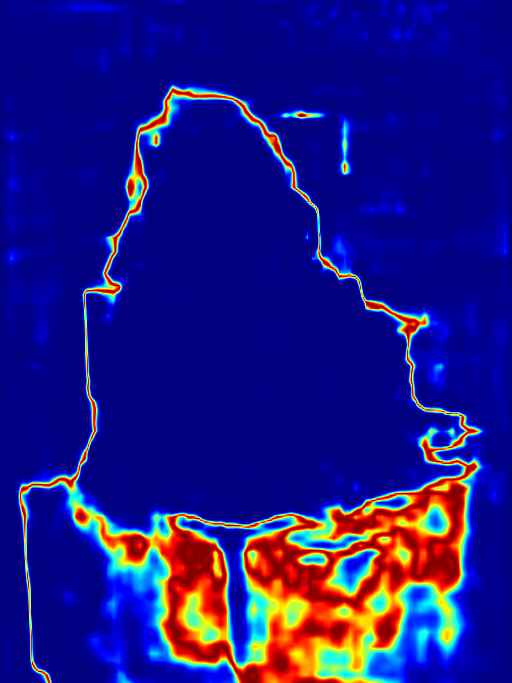}}\\
  {\includegraphics[width=\linewidth]{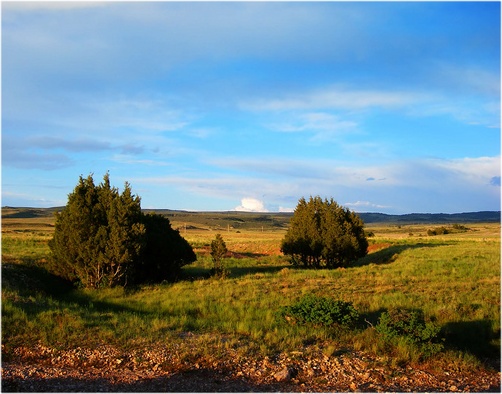}}   &
  {\includegraphics[width=\linewidth]{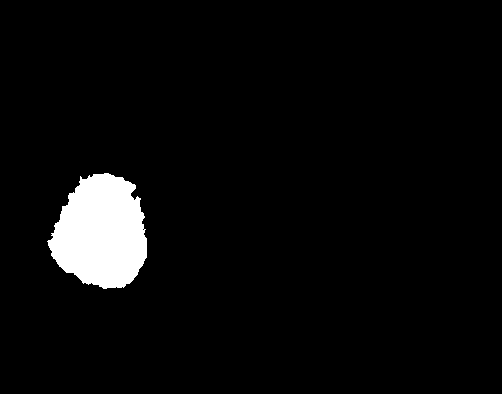}}  &
  {\includegraphics[width=\linewidth]{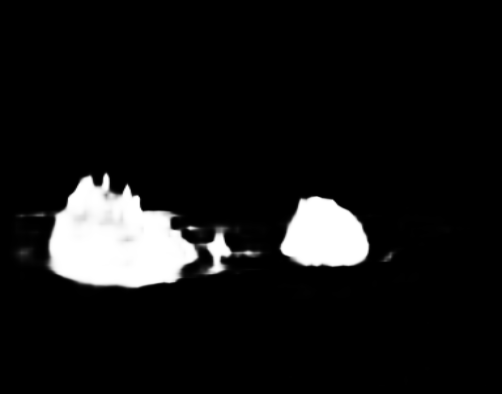}} &
  {\includegraphics[width=\linewidth]{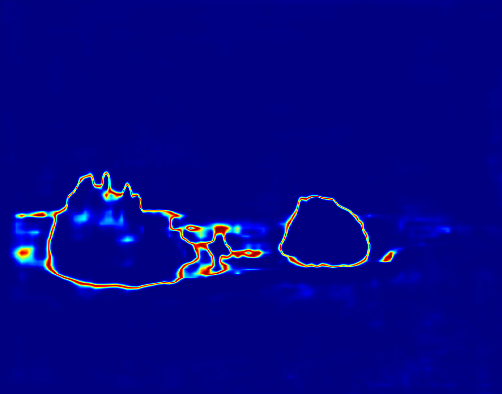}}  &
  {\includegraphics[width=\linewidth]{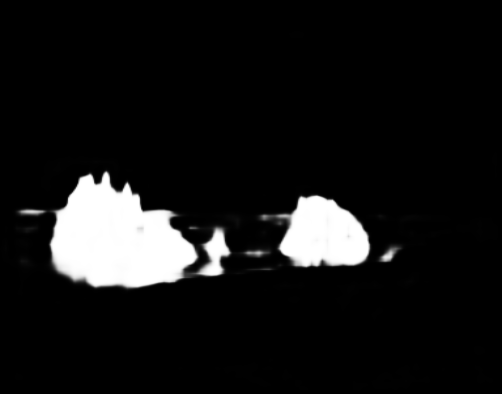}} &
  {\includegraphics[width=\linewidth]{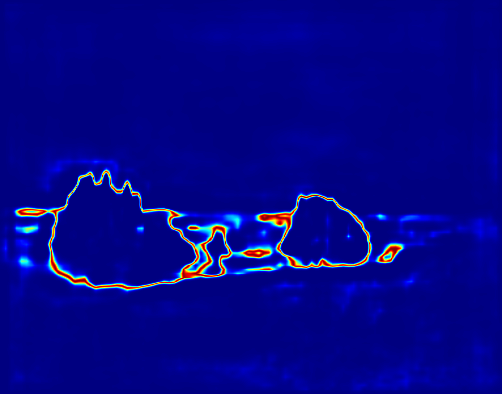}} &
  {\includegraphics[width=\linewidth]{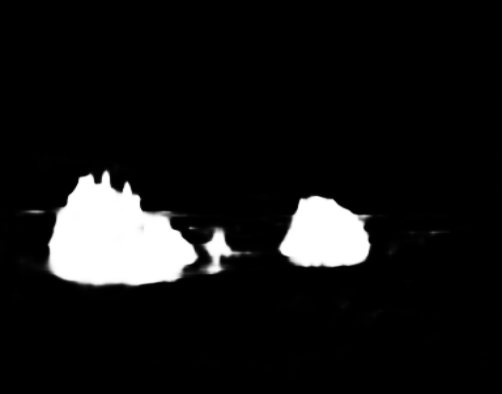}} &
  {\includegraphics[width=\linewidth]{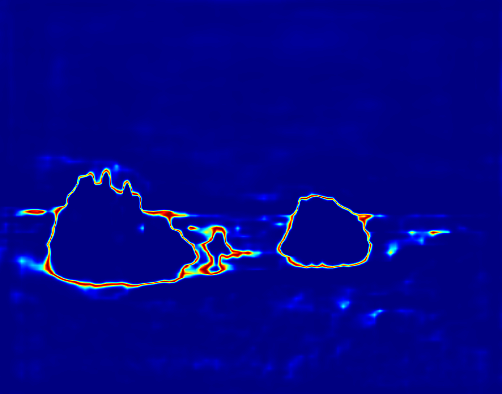}}  &
  {\includegraphics[width=\linewidth]{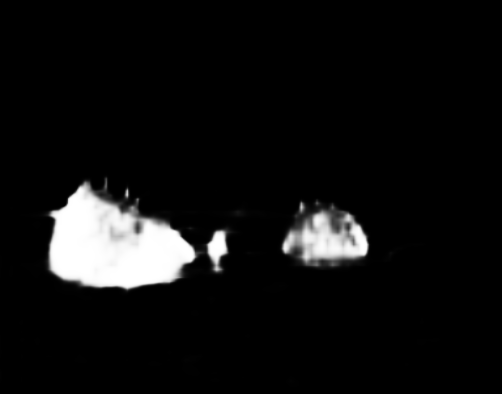}} &
  {\includegraphics[width=\linewidth]{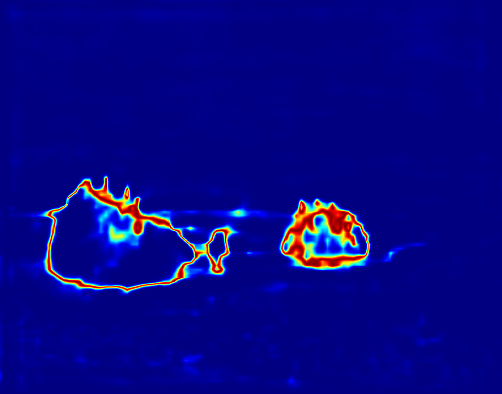}}\\
    \multicolumn{1}{c}{\footnotesize{Image}}&\multicolumn{1}{c}{\footnotesize{GT}}&\multicolumn{2}{c}{\footnotesize{$\text{TGAN}$}}&\multicolumn{2}{c}{\footnotesize{$\text{TCVAE}$}}&\multicolumn{2}{c}{\footnotesize{$\text{TABP}$}}&\multicolumn{2}{c}{\footnotesize{$\text{TIGAN}$}}
  \end{tabular}
  \end{center}
%   \vspace{-5pt}
  \caption{\footnotesize{Predictions from different generative models in Table \ref{tab:reliable_rgb_sod}, where we randomly sample $T=10$ times and obtain the entropy of mean prediction as predictive uncertainty \cite{kendall2017uncertainties}. Note that the predictions within each method block are saliency prediction and uncertainty respectively.}
  }
\label{fig:visualization_reliable_sod_methods}
\end{figure*}

\begin{table*}[t!]
  \centering
  \scriptsize
  \renewcommand{\arraystretch}{1.1}
  \renewcommand{\tabcolsep}{0.651mm}
  \caption{\footnotesize{Weakly-supervised SOD models analysis with both the CNN and transformer backbone, where models in the top block are the CNN based models with different weakly supervised loss functions, and models in the bottom block are the corresponding transformer based counterparts.}}
  \begin{tabular}{l|cccc|cccc|cccc|cccc|cccc|cccc}
  \hline
% \toprule
  &\multicolumn{4}{c|}{DUTS \cite{imagesaliency}}&\multicolumn{4}{c|}{ECSSD \cite{yan2013hierarchical}}&\multicolumn{4}{c|}{DUT \cite{Manifold-Ranking:CVPR-2013}}&\multicolumn{4}{c|}{HKU-IS \cite{li2015visual}}&\multicolumn{4}{c|}{PASCAL-S \cite{pascal_s_dataset}}&\multicolumn{4}{c}{SOD \cite{sod_dataset}} \\
    Method & $S_{\alpha}\uparrow$&$F_{\beta}\uparrow$&$E_{\xi}\uparrow$&$\mathcal{M}\downarrow$& $S_{\alpha}\uparrow$&$F_{\beta}\uparrow$&$E_{\xi}\uparrow$&$\mathcal{M}\downarrow$& $S_{\alpha}\uparrow$&$F_{\beta}\uparrow$&$E_{\xi}\uparrow$&$\mathcal{M}\downarrow$& $S_{\alpha}\uparrow$&$F_{\beta}\uparrow$&$E_{\xi}\uparrow$&$\mathcal{M}\downarrow$& $S_{\alpha}\uparrow$&$F_{\beta}\uparrow$&$E_{\xi}\uparrow$&$\mathcal{M}\downarrow$& $S_{\alpha}\uparrow$&$F_{\beta}\uparrow$&$E_{\xi}\uparrow$&$\mathcal{M}\downarrow$ \\ \hline
   $\text{BCNN}$ &.754 &.641 &.806 &.088 &.822 &.771 &.868 &.087 &.723 &.588 &.758 &.107 &.807 &.739 &.864 &.080 &.760 &.703 &.812 &.117 &.745 &.692 &.801 &.117  \\
   $\text{BCNN\_S}$ &.805 &.724 &.858 &.069 &.869 &.846 &.913 &.060 &.771 &.664 &.806 &.088 &.859 &.822 &.913 &.055 &.802 &.768 &.855 &.095 &.780 &.750 &.841 &.098  \\ 
   $\text{BCNN\_G}$ &.835 &.781 &.884 &.053 &\underline{.895} &.889 &.934 &.046 &.796 &.708 &.831 &.068 &.883 &.871 &.932 &.041 &.825 &.807 &.874 &.080 &.797 &.787 &.843 &.870  \\ 
   $\text{BCNN\_{self}}$&.787 &.682 &.834 &.077 &.851 &.803 &.888 &.075 &.761 &.634 &.793 &.089 &.833 &.767 &.880 &.072 &.796 &.743 &.839 &.100 &.766 &.716 &.811 &.111  \\
   $\text{BCNN\_{weak}}$ &.839 &.792 &.892 &.050 &\underline{.895} &.892 &.934 &.044 &.800 &.719 &.835 &.063 &.886 &.877 &.935 &.039 &.829 &.813 &.880 &.078 &.788 &.778 &.834 &.088   \\ \hline
   $\text{BT}$ &.748 &.636 &.798 &.081 &.834 &.786 &.885 &.071 &.728 &.600 &.765 &.099 &.802 &.734 &.861 &.077 &.777 &.727 &.830 &.100 &.756 &.710 &.820 &.108  \\
   $\text{BT\_S}$ &.829 &.767 &.893 &.053 &.886 &.878 &.938 &.045 &.807 &.726 &.854 &.069 &.869 &.848 &.932 &.046 &.825 &.810 &.884 &.078 &.802 &.791 &.869 &.085  \\ 
   $\text{BT\_G}$ &\underline{.857} &\underline{.808} &\underline{.916} &\underline{.043} &\textbf{.908} &\underline{.903} &\underline{.951} &\underline{.036} &\underline{.825} &\underline{.751} &\underline{.870} &\underline{.059} &\underline{.896} &\underline{.883} &\underline{.949} &\underline{.035} &\underline{.843} &\underline{.831} &\underline{.898} &\underline{.069} &\underline{.819} &\underline{.813} &\underline{.877} &\underline{.077}  \\ 
   $\text{BT\_{self}}$ &.791 &.683 &.840 &.070 &.854 &.799 &.893 &.069 &.773 &.649 &.809 &.083 &.832 &.763 &.882 &.070 &.802 &.748 &.846 &.094 &.781 &.732 &.836 &.101  \\ 
   $\text{BT\_{weak}}$ &\textbf{.858} &\textbf{.815} &\textbf{.917} &\textbf{.042} &\textbf{.908} &\textbf{.907} &\textbf{.952} &\textbf{.035} &\textbf{.835} &\textbf{.770} &\textbf{.879} &\textbf{.054} &\textbf{.898} &\textbf{.891} &\textbf{.952} &\textbf{.034} &\textbf{.848} &\textbf{.843} &\textbf{.904} &\textbf{.065} &\textbf{.821} &\textbf{.819} &\textbf{.880} &\textbf{.075}   \\ 
 \hline
  \end{tabular}
  \label{tab:weakly_supervised_sod}
\end{table*}

For easier reference, we also include the baseline models $\text{B'\_{cnn}}$ and $\text{B'\_{tr}}$ from Table \ref{tab:fully_rgb_sod_expeiments} in Table \ref{tab:reliable_rgb_sod}, and the CNN and transformer backbone based stochastic models are built upon the two baseline models respectively. The stochastic RGB-D SOD models in Table \ref{tab:reliable_rgbd_sod} are based on the corresponding RGB-D SOD models with auxiliary depth module as shown in Table \ref{tab:rgbd_sod_analysis}. We show the performance of \enquote{CADE} and \enquote{TADE} for easier reference. Table \ref{tab:reliable_rgb_sod} and Table \ref{tab:reliable_rgbd_sod} show that the four types of generative models can achieve comparable deterministic performance (compared with the corresponding deterministic baseline models) for both RGB image based SOD and RGB-D image pair based SOD. As the goal of a generative model is to obtain stochastic predictions for the model explanation, the deterministic performance may be slightly The proposed inferential GAN, \eg~\enquote{CABP} for RGB SOD. The main reason lies in two parts. First, the hyper-parameters within the inference model in Eq.~\eqref{langevin_dynamics} need to be tuned to effectively explore the latent space. Second, the final performances of those generative models are obtained by performing multiple iterations (10 iterations in this paper) of forward passes during testing, and the performance of the mean prediction is then reported, which varies with different iterations of sampling. In this paper, we focus on the new generative model for accurate and reliable SOD,
% as we have already obtain comparable deterministic performance, 
we leave the generative model hyper-parameter tuning for future work.

Besides the deterministic performance, the main advantage of the generative model is its ability for stochastic predictions, making it possible to estimate predictive uncertainty \cite{kendall2017uncertainties} for model reliability estimation. In Fig.~\ref{fig:visualization_reliable_sod_methods},
% coAs the proposed inferential GAN aims
% \NB{aims} \sout{is} 
% to achieve reliable saliency detection with reasonable uncertainty maps, we then 
we visualize the uncertainty maps of each generative model. In this paper, the \enquote{uncertainty} refers to predictive uncertainty \cite{uncertainty_decomposition,kendall2017uncertainties}, which is the total uncertainty, including both data uncertainty and model uncertainty. Given the mean predictions after multiple forward passes during testing, the predictive uncertainty is defined as the entropy of the mean prediction. A reliable model should be aware of its prediction, leading to a reasonable uncertainty model to explain model prediction. Fig.~\ref{fig:visualization_reliable_sod_methods} shows that the uncertainty map from the proposed inferential GAN explains better model prediction, \Rev{highlighting the hard samples caused by training/testing discrepancy. Especially, for the
% sample in the 
$1^{st}$ sample
% row, 
the bottom region is relatively low-contrast, which is different from the high-contrast foreground in the training dataset. The ground truth of the $2^{nd}$
% row 
image is biased, focusing only on a compact foreground region, where the less compact region is discarded. All four latent variable models can discover the discarded less compact region, where the uncertainty map of \enquote{TIGAN} is more informative in explaining the less accurate predictions
(see Table~\ref{tab:mc_dropout_analysis} for extensive comparison).}
% which highlights both the hard samples (data uncertainty, image in the $1^{st}$ row) and out-of-distribution samples (model uncertainty, image in the $2^{nd}$ row).

\begin{figure*}[tp]
%  \vspace{-5mm}
   \begin{center}
  \setlength\tabcolsep{2pt}
  \begin{tabular}{*{2}{p{0.075\textwidth}<{\centering}} | *{5}{p{0.075\textwidth}<{\centering}} | *{5}{p{0.075\textwidth}<{\centering}}}
  {\includegraphics[width=\linewidth]{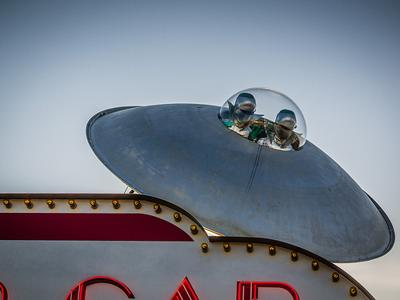}}
  &{\includegraphics[width=\linewidth]{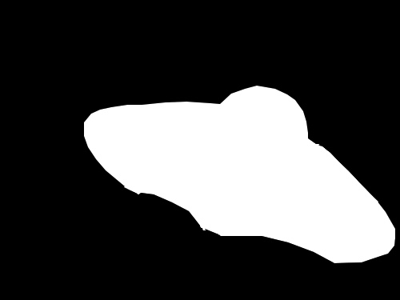}} 
  &{\includegraphics[width=\linewidth]{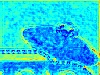}} 
  &{\includegraphics[width=\linewidth]{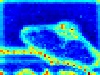}} 
  &{\includegraphics[width=\linewidth]{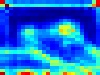}} 
  &{\includegraphics[width=\linewidth]{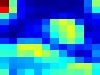}} 
  &{\includegraphics[width=\linewidth]{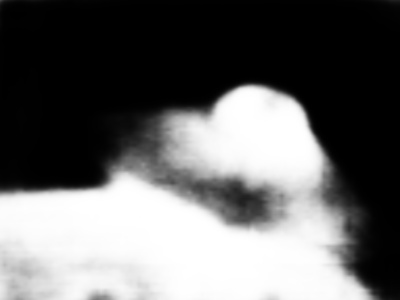}}  
  &{\includegraphics[width=\linewidth]{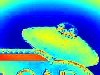}} 
  &{\includegraphics[width=\linewidth]{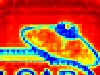}} 
  &{\includegraphics[width=\linewidth]{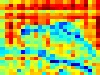}} 
  &{\includegraphics[width=\linewidth]{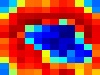}} 
  &{\includegraphics[width=\linewidth]{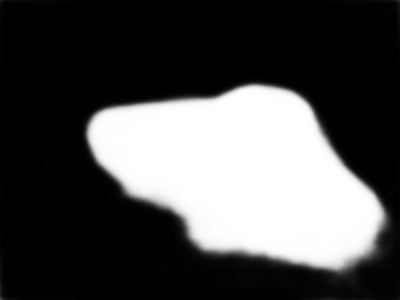}}  \\
  {\includegraphics[width=\linewidth]{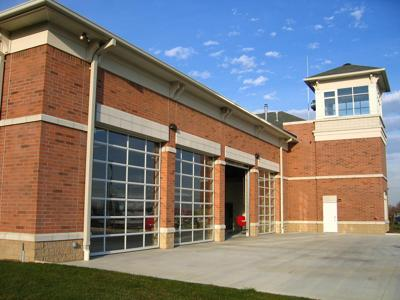}} 
  &{\includegraphics[width=\linewidth]{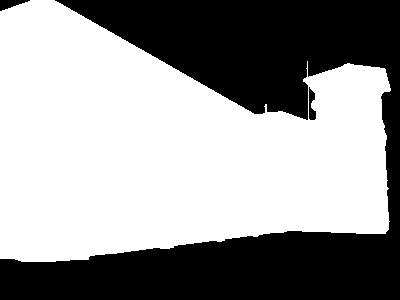}} 
  &{\includegraphics[width=\linewidth]{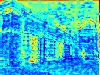}} 
  &{\includegraphics[width=\linewidth]{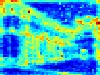}} 
  &{\includegraphics[width=\linewidth]{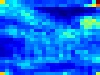}} 
  &{\includegraphics[width=\linewidth]{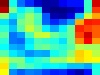}} 
  &{\includegraphics[width=\linewidth]{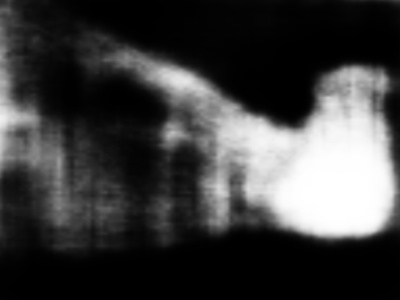}} 
  &{\includegraphics[width=\linewidth]{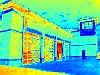}} 
  &{\includegraphics[width=\linewidth]{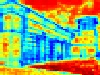}} 
  &{\includegraphics[width=\linewidth]{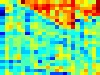}} 
  &{\includegraphics[width=\linewidth]{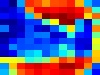}} 
  &{\includegraphics[width=\linewidth]{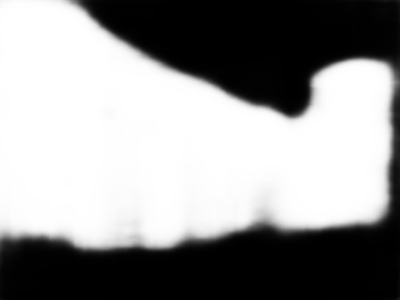}} \\
  \multicolumn{1}{c}{\footnotesize{Image}}&\multicolumn{1}{c}{\footnotesize{GT}}&\multicolumn{5}{c}{\footnotesize{CNN}}&\multicolumn{5}{c}{\footnotesize{Transformer}}\\ 
   % \begin{tabular}{c@{ }}
   % {\includegraphics[width=0.95\linewidth]{imgs/scribble-weak.pdf}} \\
   \end{tabular}
   \end{center}
%   \vspace{-15pt}
   \caption{\footnotesize{Features of CNN (ResNet50 \cite{resnet}) and transformer backbone (Swin \cite{liu2021swin}) for weakly-supervised SOD using only partial cross-entropy loss.}
   }
\label{fig:weak_feature_visualization}
\end{figure*}

% \section{Weakly Supervised Salient Object Detection}
% \section{Weakly-/semi-supervised Salient Object Detection}

\subsection{Accurate and Reliable Weakly-supervised Salient Object Detection}
\label{sec:accurate_reliable_weak_sod}

Different from pixel-wise annotation based fully supervised SOD,
% which relies on pixel-wise annotations, 
weakly supervised SOD models learn saliency from cheap annotations, \eg,~scribble annotations \cite{jing2020weakly}, image-level labels \cite{imagesaliency}.
In this paper, we investigate the superiority of the
% \NB{the} 
transformer backbone for weakly supervised SOD with scribble supervision~\cite{jing2020weakly}.

% \subsection{Task Analysis}
% \noindent\textbf{Weakly-supervised Salient Object Detection via Scribble Annotation:} 

% \noindent\textbf{Semi-supervised Salient Object Detection:} For the semi-supervised learning setting, we have labeled set and unlabeled set. The main focus of semi-supervised learning is then to effectively use the unlabeled set for model updating. Two main directions are widely studied for semi-supervised learning, namely pseudo-labeling and consistency loss. The former generates pseudo labels for the unlabeled set, which is then added to the labeled to set to update model parameters. The latter aims to design some self-supervised loss functions to achieve transformation robust predictions. In this paper, we investigate both strategies for semi-supervised salient object detection with both CNN backbone and transformer backbone.

\subsubsection{Weakly-supervised Transformer}
% The generative models for the
% % \NB{the} 
% fully-supervised setting are
% % \NB{setting are} \sout{is} 
% straight-forward, as we can simply take the inferred latent variable as part of the input, and the discriminator can directly estimate real/fake of its input (the ground truth/model prediction).
% Compared with fully supervised learning, t
The main difficulty of learning from weak annotations is the missing structure information, which cannot be recovered without extra structure-aware regularizers.
% In Fig.~\ref{fig:scribble_annotation_visualization}, we visualize the weak label (scribble annotation) and full label (pixel-wise annotation), which clearly indicates the missing structure attributes of the weak annotation. 
In this way, the main focus of designing models to learn from weak annotations is to recover the missing structure information. Specifically, we investigate three widely used strategies, namely smoothness loss~\cite{smoothness_loss}, gated CRF loss~\cite{obukhov2019gated}, and data-augmentation based consistency loss.
% to force the model to achieve transformation robust predictions.
% self-supervised loss. 
The first and second ones aim to recover the structure of prediction, and the third one aims to achieve transformation robust prediction, serving as an internal data augmentation trick. To test how the model performs with different loss functions within both the CNN framework and the transformer framework, we design models with each type of loss function within both the CNN backbone and transformer backbone and report the performance in Table \ref{tab:weakly_supervised_sod}.

\begin{figure*}[tp]
%  \vspace{-5mm}
  \begin{center}
  \begin{tabular}{c@{ } c@{ } c@{ } c@{ } c@{ } c@{ } c@{ } c@{ } c@{ }}
  {\includegraphics[width=0.104\linewidth]{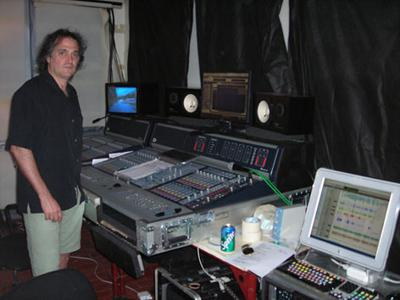}} &
  {\includegraphics[width=0.104\linewidth]{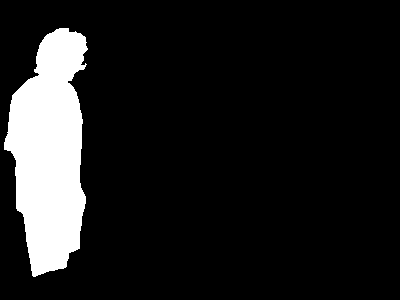}} &
  {\includegraphics[width=0.104\linewidth]{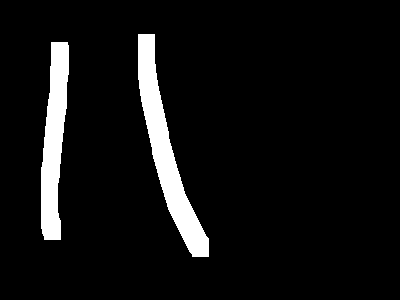}} &
  {\includegraphics[width=0.104\linewidth]{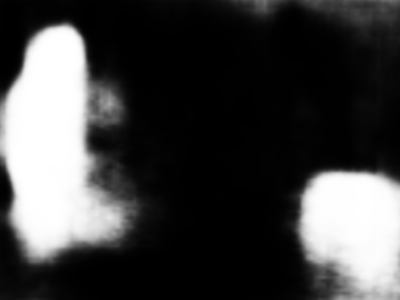}} &
  {\includegraphics[width=0.104\linewidth]{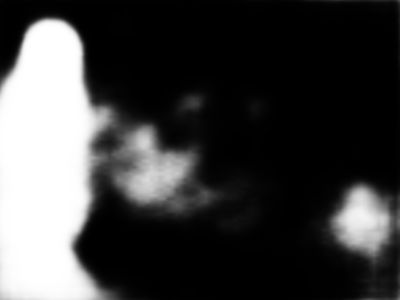}} &
  {\includegraphics[width=0.104\linewidth]{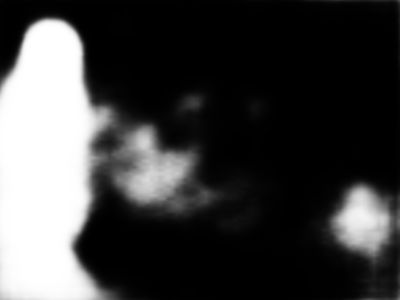}} &
  {\includegraphics[width=0.104\linewidth]{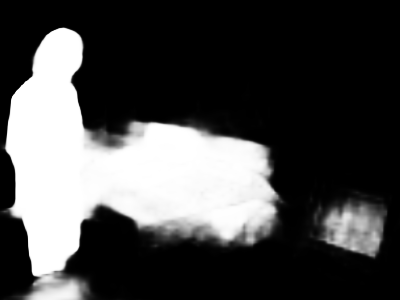}} &
  {\includegraphics[width=0.104\linewidth]{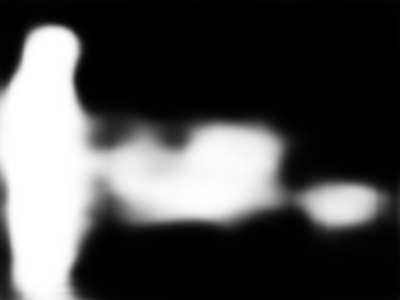}} &
  {\includegraphics[width=0.104\linewidth]{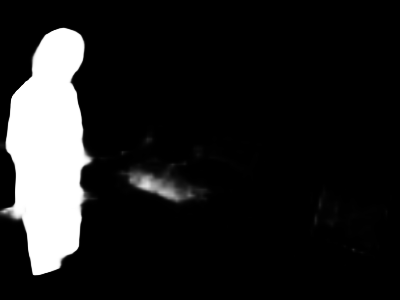}} \\
  % {\includegraphics[width=0.104\linewidth]{figures/scribble_show/img/ILSVRC2012_test_00020285.png}} &
  % {\includegraphics[width=0.104\linewidth]{figures/scribble_show/gt/ILSVRC2012_test_00020285.png}} &
  % {\includegraphics[width=0.104\linewidth]{figures/scribble_show/mask/ILSVRC2012_test_00020285.png}} &
  % {\includegraphics[width=0.104\linewidth]{figures/scribble_show/cnn_weak/ILSVRC2012_test_00020285.png}} &
  % {\includegraphics[width=0.104\linewidth]{figures/scribble_show/bt_path/ILSVRC2012_test_00020285.png}} &
  % {\includegraphics[width=0.104\linewidth]{figures/scribble_show/bt_s_path/ILSVRC2012_test_00020285.png}} &
  % {\includegraphics[width=0.104\linewidth]{figures/scribble_show/bt_g_path/ILSVRC2012_test_00020285.png}} &
  % {\includegraphics[width=0.104\linewidth]{figures/scribble_show/bt_self_path/ILSVRC2012_test_00020285.png}} &
  % {\includegraphics[width=0.104\linewidth]{figures/scribble_show/bt_full_path/ILSVRC2012_test_00020285.png}} \\
  \footnotesize{Image}&\footnotesize{Full}&\footnotesize{Weak}&\footnotesize{$\text{BCNN}$}&\footnotesize{$\text{BT}$}&\footnotesize{$\text{BT\_S}$}&\footnotesize{$\text{BT\_G}$}&\footnotesize{$\text{BT\_{self}}$}&\footnotesize{$\text{BT\_{weak}}$}\\
  \end{tabular}
  \end{center}
%   \vspace{-5pt}
  \caption{\footnotesize{Weakly supervised saliency model predictions with different loss functions. Detailed performance of each model is illustrated in Table~\ref{tab:weakly_supervised_sod}.}
  }
\label{fig:weak_supervised_prediction_analysis}
\end{figure*}

\noindent\textbf{Implementation details:} To train the weakly-supervised transformer for SOD, similar to \cite{jing2020weakly}, we adopt extra smoothness loss \cite{smoothness_loss} $\mathcal{L}_{sm}$, the gated CRF loss \cite{obukhov2019gated} $\mathcal{L}_{gcrf}$ and the data-augmentation based self-supervised learning strategy \cite{structure_consistency_scribble} $\mathcal{L}_{ss}$ to recover the missing structure information in scribble annotation. The smoothness loss aims to produce a saliency map with edges well-aligned with the input image. The gated CRF loss introduces pairwise constraints to produce a saliency map with spatial consistency. The self-supervised learning strategy aims
% is proven effective in weakly-supervised learning, aiming 
to achieve transformation robust predictions. For the Swin transformer backbone \cite{liu2021swin}, as it can only take fixed-size input, we perform image rotation instead of image scaling to achieve data augmentation.
% Following \cite{structure_consistency_scribble},
Specifically, we define the self-supervised loss as a weighted sum of the structural similarity index measure \cite{Godard_2017_CVPR} and $L_1$ loss,
% For the self-supervised loss, we resize the image $x$ to half of its original scale to obtain prediction $c_h$. We then define the self-supervised loss as a weighted sum of the structural similarity index measure and L1 loss following \cite{structure_consistency_scribble}, 
which is defined as:
\begin{equation}
    \label{self-supervised-loss}
    \mathcal{L}_\text{ss} = \alpha*SSIM(s,s^t)+(1-\alpha)*\mathcal{L}_1(s,s^t),
\end{equation}
where $s=f_\theta(T(x))$ is the output of the generator with the rotated image\footnote{In this paper, we perform image rotation instead of image scaling due to the fixed input size of Swin transformer~\cite{liu2021swin}.} $T(x)$ as input, and $s^t=T(f_\theta(x))$ is the rotated prediction with original image $x$ as input. We randomly pick the rotation $T(.)$ from $\{\pi,1/2\pi,-1/2\pi\}$ in our experiments.
% to serve as the transformation operation $T(.)$.
$\alpha$ is used to balance the two types of loss functions and we set
% set the rotation to rate as 
$\alpha=0.85$ in our experiments following~\cite{structure_consistency_scribble}.
% Accordingly, $s_l^\alpha$
% % \fdp{$s_l^\alpha$} 
% is re-scaled
% % \MYX{rotted?} 
% model prediction with the original image $x$ as input.
Further, given the scribble annotation, we adopt the partial cross-entropy loss $\mathcal{L}_\text{pce}$ to constrain predictions on the scribble region. In this way, we define the loss function for the weakly-supervised model as:
\begin{equation}
    \label{weak_loss}
    \mathcal{L}_\text{weak} = \mathcal{L}_\text{pce} + \lambda_1*\mathcal{L}_\text{sm} + \lambda_2*\mathcal{L}_\text{gcrf} + \lambda_3*\mathcal{L}_\text{ss}.
\end{equation}
With grid search, we set $\lambda_1 = 0.3$,  $\lambda_2 = 1.0$ and $\lambda_3 = 1.2$.

\begin{table*}[t!]
  \centering
  \scriptsize
  \renewcommand{\arraystretch}{1.2}
  \renewcommand{\tabcolsep}{0.65mm}
  \caption{\footnotesize{Reliable weakly-supervised RGB SOD models, where the deterministic models (\enquote{$\text{*\_{weak}}$}) is trained
  % with is the proposed deterministic weakly-supervised SOD model with 
  loss function in Eq.~\eqref{weak_loss}.}}
  \begin{tabular}{l|cccc|cccc|cccc|cccc|cccc|cccc}
  \hline
% \toprule
  &\multicolumn{4}{c|}{DUTS \cite{imagesaliency}}&\multicolumn{4}{c|}{ECSSD \cite{yan2013hierarchical}}&\multicolumn{4}{c|}{DUT \cite{Manifold-Ranking:CVPR-2013}}&\multicolumn{4}{c|}{HKU-IS \cite{li2015visual}}&\multicolumn{4}{c|}{PASCAL-S \cite{pascal_s_dataset}}&\multicolumn{4}{c}{SOD \cite{sod_dataset}} \\
    Method & $S_{\alpha}\uparrow$&$F_{\beta}\uparrow$&$E_{\xi}\uparrow$&$\mathcal{M}\downarrow$& $S_{\alpha}\uparrow$&$F_{\beta}\uparrow$&$E_{\xi}\uparrow$&$\mathcal{M}\downarrow$& $S_{\alpha}\uparrow$&$F_{\beta}\uparrow$&$E_{\xi}\uparrow$&$\mathcal{M}\downarrow$& $S_{\alpha}\uparrow$&$F_{\beta}\uparrow$&$E_{\xi}\uparrow$&$\mathcal{M}\downarrow$& $S_{\alpha}\uparrow$&$F_{\beta}\uparrow$&$E_{\xi}\uparrow$&$\mathcal{M}\downarrow$& $S_{\alpha}\uparrow$&$F_{\beta}\uparrow$&$E_{\xi}\uparrow$&$\mathcal{M}\downarrow$ \\ \hline
   $\text{CGAN}$ &.834 &.785 &.887 &.053 &.891 &.888 &.932 &.046 &.792 &.706 &.826 &.069 &.881 &.873 &.933 &.041 &.828 &.814 &.880 &.079 &.798 &.797 &.852 &.082   \\
   $\text{CCVAE}$ &.832 &.781 &.882 &.055 &.894 &.889 &.934 &.045 &.793 &.708 &.827 &.070 &.883 &.872 &.933 &.041 &.822 &.806 &.876 &.081 &.797 &.792 &.851 &.085   \\
   $\text{CABP}$ &.838 &.790 &.893 &.051 &.894 &.890 &.935 &.044 &.801 &.718 &.838 &.064 &.887 &.878 &.937 &.039 &.828 &.813 &.882 &.078 &.794 &.791 &.848 &.084   \\
   $\text{CIGAN}$ &.834 &.779 &.887 &.056 &.896 &.890 &.938 &.044 &.799 &.713 &.838 &.070 &.886 &.873 &.938 &.039 &.827 &.810 &.880 &.079 &.800 &.793 &.855 &.083   \\ 
   $\text{BCNN\_{weak}}$ &.839 &.792 &.892 &.050 &.895 &.892 &.934 &.044 &.800 &.719 &.835 &.063 &.886 &.877 &.935 &.039 &.829 &.813 &.880 &.078 &.788 &.778 &.834 &.088   \\ \hline
   $\text{TGAN}$ &\underline{.856} &.813 &\textbf{.918} &\underline{.043} &.906 &.905 &.950 &.037 &.824 &.753 &.868 &.060 &\underline{.895} &.886 &.949 &\underline{.035} &\textbf{.848} &.840 &\underline{.905} &\underline{.065} &\underline{.819} &\underline{.816} &\underline{.878} &\underline{.076}   \\
   $\text{TCVAE}$ &.855 &.813 &.916 &\underline{.043} &\underline{.907} &\underline{.906} &.950 &\underline{.036} &.825 &.757 &.872 &.059 &.894 &\underline{.887} &.949 &\underline{.035} &.843 &.837 &.900 &.067 &.814 &.813 &.873 &.079   \\
   $\text{TABP}$ &.854 &.812 &\underline{.917} &\underline{.043} &.905 &.905 &\underline{.951} &\underline{.036} &\underline{.827} &.759 &\underline{.875} &\underline{.058} &.893 &\underline{.887} &\underline{.950} &\underline{.035} &\underline{.847} &\textbf{.844} &\textbf{.906} &\textbf{.064} &.810 &.810 &.868 &.082   \\
   $\text{TIGAN}$ &.855 &\underline{.814} &\textbf{.918} &\underline{.043} &.905 &.905 &.950 &.037 &.826 &\underline{.760} &.874 &\underline{.058} &.893 &\underline{.887} &.949 &\underline{.035} &.844 &.839 &.902 &.066 &.811 &.810 &.872 &.082   \\ 
   $\text{BT\_{weak}}$ &\textbf{.858} &\textbf{.815} &\underline{.917} &\textbf{.042} &\textbf{.908} &\textbf{.907} &\textbf{.952} &\textbf{.035} &\textbf{.835} &\textbf{.770} &\textbf{.879} &\textbf{.054} &\textbf{.898} &\textbf{.891} &\textbf{.952} &\textbf{.034} &\textbf{.848} &\underline{.843} &.904 &\underline{.065} &\textbf{.821} &\textbf{.819} &\textbf{.880} &\textbf{.075}   \\  \hline
  \end{tabular}
  \label{tab:reliable_weakly_sod}
\end{table*}

\noindent\textbf{CNN backbone vs Transformer backbone:}
% We observe clear object structure in
% \NB{in?} \sout{of} 
% the transformer backbone within the lower level features as shown in Fig.~\ref{fig:structure_comparison}.
% , which indicates better structure preserving property of transformer backbone even before fine-tuning it for SOD. 
We design two different models with ResNet50~\cite{resnet} and Swin transformer~\cite{liu2021swin} as the backbone. The decoder part is the same as $\text{B'\_{tr}}$ in Table \ref{tab:fully_rgb_sod_expeiments}. We train the two models with only partial cross-entropy loss, and we show their corresponding results in Table \ref{tab:weakly_supervised_sod} \enquote{BCNN} and \enquote{BT} respectively. The significantly improved performance of \enquote{BT} compared with \enquote{BCNN} shows the effectiveness of the transformer backbone for weakly-supervised SOD. We also visualize the features of the two trained models in Fig.~\ref{fig:weak_feature_visualization}, where heat maps are the features and gray maps are the predictions. We observe clear structure information in \enquote{Transformer}, which explains the superior performance of the transformer for weakly-supervised learning via supervisions with less structure information \cite{naseer_IntriguingProperties_Arxiv_2021}.

\noindent\textbf{Weakly-supervised loss analysis:}
Besides the partial cross-entropy loss, we use three extra loss functions for weakly-supervised SOD, namely smoothness loss to constrain the predictions to be well aligned with the image edges, gated CRF loss to regularize the pairwise term predictions which aim to produce similar predictions for spatially similar pixels, and a self-supervised loss to effectively learn from less
% \sout{fewer} \NB{less} 
supervision data with consistency loss, \eg, rotation-invariant predictions. We then carry out extra experiments to verify the effectiveness of each loss function and show the results in Table \ref{tab:weakly_supervised_sod}. \enquote{$\text{BCNN\_S}$}, \enquote{$\text{BCNN\_G}$} and \enquote{$\text{BCNN\_{self}}$} indicate baseline model (\enquote{BCNN}) training with extra smoothness loss \enquote{$\mathcal{L}_\text{sm}$}, gated CRF loss \enquote{$\mathcal{L}_\text{gcrf}$} and self-supervised loss \enquote{$\mathcal{L}_\text{ss}$}. \enquote{$\text{BT\_S}$}, \enquote{$\text{BT\_G}$} and \enquote{$\text{BT\_{self}}$} are the corresponding transformer backbone counterparts.
% Note that, all the experiments in this section are built upon the base model in our main paper Table 3. \enquote{Base} is the base performance with only the partial cross-entropy loss. 
% \enquote{SSl}, \enquote{Smooth} and \enquote{GCRF} are models of adding self-supervised loss, smoothness loss and gated CRF loss to the \enquote{Base} respectively. 
We observe an improved performance of each extra loss function, which explains their effectiveness. Further, we find that the smoothness and gated CRF achieve more performance gain than the self-supervised loss, which mainly comes from their effective structure modeling ability.
% of each of the two loss functions. 
The improved performance of $\text{BCNN\_{weak}}$ and $\text{BT\_{weak}}$ with the weighted loss function in Eq.~\eqref{weak_loss} compared with the corresponding models with individual loss functions verifies the effectiveness of the proposed weighted weakly supervised loss function. We also show predictions of the weakly supervised models in Fig.~\ref{fig:weak_supervised_prediction_analysis}. It is clear that both the base model with only partial cross-entropy loss (\enquote{$\text{BCNN}$} and \enquote{$\text{BT}$}) and the model with extra self-supervised loss (\enquote{$\text{BT\_{self}}$}) fail to accurately localize object boundaries, leading to blurred predictions. The main reason is the absence of structure constraints. The smoothness loss and the gated CRF loss work better in modeling the structure information, leading to more accurate predictions, especially along object boundaries, and models (\enquote{$\text{BCNN\_{weak}}$} and \enquote{$\text{BT\_{weak}}$}) with our final loss function (Eq.~\eqref{weak_loss}) achieve the best performance.

\begin{table*}[t!]
  \centering
  \scriptsize
  \renewcommand{\arraystretch}{1.0}
  \renewcommand{\tabcolsep}{0.69mm}
  \caption{\footnotesize{Model analysis related experiments, where we discuss model performance with respect to model optimizer (\enquote{$\text{B'\_{SGD}}$} and \enquote{$\text{B'\_{trSGD}}$}), initialization weights (\enquote{$\text{B'\_{R}}$}, $\text{B'\_{trR}}$ and $\text{B'\_{tr22K}}$) and different transformer backbones (\enquote{$\text{B'\_{ViT}}$}).
%   \YC{Detailed caption is preferred.}
%   investigation on backbone networks and initialization weights for the backbones.
  }}
%   \caption{\footnotesize{Performance comparison with different backbone networks for RGB SOD.}}
  \begin{tabular}{l|cccc|cccc|cccc|cccc|cccc|cccc}
  \hline
% \toprule
  &\multicolumn{4}{c|}{DUTS \cite{imagesaliency}}&\multicolumn{4}{c|}{ECSSD \cite{yan2013hierarchical}}&\multicolumn{4}{c|}{DUT \cite{Manifold-Ranking:CVPR-2013}}&\multicolumn{4}{c|}{HKU-IS \cite{li2015visual}}&\multicolumn{4}{c|}{PASCAL-S \cite{pascal_s_dataset}}&\multicolumn{4}{c}{SOD \cite{sod_dataset}} \\
    Method & $S_{\alpha}\uparrow$&$F_{\beta}\uparrow$&$E_{\xi}\uparrow$&$\mathcal{M}\downarrow$& $S_{\alpha}\uparrow$&$F_{\beta}\uparrow$&$E_{\xi}\uparrow$&$\mathcal{M}\downarrow$& $S_{\alpha}\uparrow$&$F_{\beta}\uparrow$&$E_{\xi}\uparrow$&$\mathcal{M}\downarrow$& $S_{\alpha}\uparrow$&$F_{\beta}\uparrow$&$E_{\xi}\uparrow$&$\mathcal{M}\downarrow$& $S_{\alpha}\uparrow$&$F_{\beta}\uparrow$&$E_{\xi}\uparrow$&$\mathcal{M}\downarrow$& $S_{\alpha}\uparrow$&$F_{\beta}\uparrow$&$E_{\xi}\uparrow$&$\mathcal{M}\downarrow$ \\ \hline
%   ResNet50  &.876 &.826 &.910 &.041 &.918 &.910 &.944 &.038 &.820 &.733 &.846 &.062 &.909 &.894 &.945 &.033 &.856 &.840 &.895 &.067 &.827 &.813 &.863 &.073   \\
  $\text{B'\_{cnn}}$ &.882 &.840 &.916 &.037 &.922 &.919 &.947 &.035 &.823 &.742 &.851 &.057 &.912 &.901 &.947 &.030 &.855 &.841 &.896 &.065 &.832 &.825 &.863 &.073  \\
  $\text{B'\_{SGD}}$ &.876 &.826 &.910 &.041 &.918 &.910 &.944 &.038 &.820 &.733 &.846 &.062 &.909 &.894 &.945 &.033 &.856 &.840 &.895 &.067 &.827 &.813 &.863 &.077   \\
  $\text{B'\_{R}}$  &.745 &.623 &.773 &.110 &.832 &.789 &.853 &.093 &.738 &.605 &.762 &.114 &.820 &.760 &.858 &.084 &.752 &.697 &.777 &.140 &.725 &.673 &.758 &.145   \\
%   $\text{B'\_{DS}}$  &.880 &.839 &.918 &.038 &.924 &.922 &.952 &.033 &.822 &.743 &.853 &.058 &.912 &.903 &.952 &.030 &.855 &.845 &.900 &.065 &.830 &.828 &.867 &.71   \\
    \hline
    % $\text{B'\_{tr}}$ &.911 &.882 &.947 &.026 &.939 &.940 &.965 &.024 &.860 &.801 &.894 &.045 &.927 &.921 &.964 &.023 &.876 &.872 &.917 &.053 &.858 &.853 &.897 &.059   \\
    $\text{B'\_{tr}}$ &.911 &.882 &.947 &.026 &.939 &.940 &\underline{.965} &\underline{.024} &.860 &.801 &.894 &.045 &.927 &.921 &.964 &.023 &.876 &.872 &.917 &\underline{.053} &\underline{.858} &.853 &\underline{.897} &\textbf{.059} \\
    $\text{B'\_{trSGD}}$ &.899 &.861 &.936 &.031 &.928 &.923 &.954 &.032 &.854 &.786 &.886 &.046 &.921 &.909 &.956 &.029 &.867 &.856 &.905 &.061 &.833 &.818 &.862 &.075   \\
    $\text{B'\_{trR}}$ &.768 &.667 &.804 &.097 &.848 &.819 &.874 &.082 &.754 &.637 &.784 &.105 &.843 &.803 &.884 &.070 &.760 &.715 &.794 &.134 &.730 &.689 &.763 &.142    \\
    $\text{B'\_{tr22K}}$ &\underline{.918} &\underline{.891} &\underline{.952} &\underline{.025} &\textbf{.944} &\underline{.943} &\textbf{.967} &\textbf{.022} &\underline{.869} &\underline{.814} &\underline{.902} &\underline{.044} &\underline{.933} &\underline{.928} &\underline{.968} &\underline{.022} &\textbf{.885} &\underline{.881} &\underline{.925} &\textbf{.050} &\textbf{.863} &\textbf{.863} &\textbf{.900} &\textbf{.059}   \\
    $\text{B'\_{ViT}}$ &\textbf{.922} &\textbf{.899} &\textbf{.955} &\textbf{.023} &\underline{.943} &\textbf{.945} &\textbf{.967} &\textbf{.022} &\textbf{.874} &\textbf{.824} &\textbf{.906} &\textbf{.043} &\textbf{.934} &\textbf{.931} &\textbf{.969} &\textbf{.021} &\underline{.884} &\textbf{.884} &\textbf{.926} &\textbf{.050} &\underline{.858} &\underline{.859} &.895 &\underline{.065}   \\
    % $\text{B'\_{trDS}}$ &.908 &.882 &.947 &.027 &.938 &.939 &.964 &.024 &.859 &.802 &.895 &.047 &.929 &.924 &.966 &.023 &.876 &.874 &.918 &.054 &.856 &.858 &.897 &.062   \\
     \hline
  \end{tabular}
  \label{tab:model_analysis}
%   \vspace{-5mm}
\end{table*}

\begin{table*}[t!]
  \centering
  \scriptsize
  \renewcommand{\arraystretch}{1.2}
  \renewcommand{\tabcolsep}{0.67mm}
  \caption{\footnotesize{Replacing the CNN backbone of existing RGB SOD models with a transformer backbone.}}
  \begin{tabular}{l|cccc|cccc|cccc|cccc|cccc|cccc}
  \hline
% \toprule
   &\multicolumn{4}{c|}{DUTS \cite{imagesaliency}}&\multicolumn{4}{c|}{ECSSD \cite{yan2013hierarchical}}&\multicolumn{4}{c|}{DUT \cite{Manifold-Ranking:CVPR-2013}}&\multicolumn{4}{c|}{HKU-IS \cite{li2015visual}}&\multicolumn{4}{c|}{PASCAL-S \cite{pascal_s_dataset}}&\multicolumn{4}{c}{SOD \cite{sod_dataset}} \\
    Method & $S_{\alpha}\uparrow$&$F_{\beta}\uparrow$&$E_{\xi}\uparrow$&$\mathcal{M}\downarrow$& $S_{\alpha}\uparrow$&$F_{\beta}\uparrow$&$E_{\xi}\uparrow$&$\mathcal{M}\downarrow$& $S_{\alpha}\uparrow$&$F_{\beta}\uparrow$&$E_{\xi}\uparrow$&$\mathcal{M}\downarrow$& $S_{\alpha}\uparrow$&$F_{\beta}\uparrow$&$E_{\xi}\uparrow$&$\mathcal{M}\downarrow$& $S_{\alpha}\uparrow$&$F_{\beta}\uparrow$&$E_{\xi}\uparrow$&$\mathcal{M}\downarrow$& $S_{\alpha}\uparrow$&$F_{\beta}\uparrow$&$E_{\xi}\uparrow$&$\mathcal{M}\downarrow$ \\ \hline
    SCRN \cite{scrn_sal} & .885 & .833 & .900 & .040 & .920 & .910 & .933 & .041 & .837 & .749 & .847 & .056 & .916 & .894 & .935 & .034 & .869 & .833 & .892 & .063 & .817 & .790 & .829 & .087\\ 
   F3Net \cite{wei2020f3net} & .888 & .852 & .920 & .035 & .919 & .921 & .943 & .036 & .839 &  .766 & .864 & .053 & .917 & .910 & .952 & .028 & .861 & .835 & .898 & .062 & .824 & .814 & .850 & .077\\
%   ITSD \cite{zhou2020interactive} & .886 & .841 & .917 & .039 & .920 & .916 & .943 & .037 & .842 & .767 & .867 & .056 & .921 & .906 & .950 & .030 & .860 & .830 & .894 & .066 & .836 & .829 & .867 & .076\\ 
   \hline
   SCRN* \cite{scrn_sal}  &.908 &.873 &.937 &.029 &\underline{.937} &.935 &\underline{.960} &.027 &\textbf{.864} &.800 &.890 &\underline{.044} &\textbf{.930} &.919 &\underline{.958} &\underline{.026} &\underline{.875} &\underline{.870} &.911 &.057 &.843 &.836 &.865 &.069 \\
   F3Net* \cite{wei2020f3net} &\textbf{.913} &\textbf{.891} &\textbf{.950} &\textbf{.024} &\textbf{.939} &\textbf{.941} &\textbf{.965} &\textbf{.023} &\underline{.860} &\textbf{.802} &\underline{.891} &\textbf{.042} &\underline{.928} &\textbf{.925} &\textbf{.964} &\textbf{.023} &.872 &\textbf{.872} &\underline{.916} &\underline{.055} &\underline{.848} &\underline{.849} &\underline{.881} &\underline{.065} \\ \hline
%   ITSD* \cite{zhou2020interactive} &. &. &. &. &. &. &. &. &. &. &. &. &. &. &. &. &. &. &. &. &. &. &. &. \\ \hline
   $\text{B'\_{tr}}$  &\underline{.911} &\underline{.882} &\underline{.947} &\underline{.026} &\textbf{.939} &\underline{.940} &\textbf{.965} &\underline{.024} &\underline{.860} &\underline{.801} &\textbf{.894} &.045 &.927 &\underline{.921} &\textbf{.964} &\textbf{.023} &\textbf{.876} &\textbf{.872} &\textbf{.917} &\textbf{.053} &\textbf{.858} &\textbf{.853} &\textbf{.897} &\textbf{.059}  \\
   \hline 
% \bottomrule
  \end{tabular}
  \label{tab:benchmark_rgb_models_ours}
%   \vspace{-5mm}
\end{table*}

\subsubsection{Reliable Weakly-supervised Transformer}
The generative models for the
% \NB{the} 
fully-supervised setting are
% \NB{setting are} \sout{is} 
straightforward, as we can simply take the inferred latent variable as part of the input, and the discriminator can directly estimate the real/fake of its input (the ground truth/model prediction). In this section, we apply our proposed inferential GAN (iGAN) to weakly-supervised RGB SOD. We also design the GAN~\cite{gan_raw}, CVAE \cite{VAE_Kingma,cvae}, and ABP~\cite{ABP} based generative models for comparison. Similarly, we design the generative model within both the CNN backbone and the transformer backbone.
% as well as GAN-based~\cite{gan_raw} and ABP-based~\cite{ABP} generative models to explore the reliability of the model for weakly-supervised RGB salient object detection.
The performance is shown in Table~\ref{tab:reliable_weakly_sod}, where \enquote{CGAN}, \enquote{CCVAE}, \enquote{CABP} and \enquote{CIGAN} are
% "CGAN", "CABP" and "CIGAN" 
the stochastic models based on GAN, CVAE, ABP, and the proposed inferential GAN respectively with CNN backbone, and \enquote{TGAN}, \enquote{TCVAE}, \enquote{TABP} and \enquote{TIGAN} are the corresponding transformer counterparts. Same as the fully-supervised reliable models in Table~\ref{tab:reliable_rgb_sod} and Table \ref{tab:reliable_rgbd_sod}, the prior and posterior distribution models of the CVAE based models are designed following~\cite{zhang2021_ucnet}.

\noindent\textbf{Implementation details:}
For the learning process of the reliable weakly-supervised transformer, we apply the same pipeline of the fully supervised method. Considering there is only scribble ground-truth as supervision in training the discriminator, we use partial cross-entropy loss for training the discriminator within the GAN and our proposed iGAN based models, and the adversarial loss $\mathcal{L}_\text{adv}$ is also partial cross-entropy loss.
% and  and partial mean-square error to measure distance between model output and scribble ground-truth for our proposed GAN and IGAN models during training.

\noindent\textbf{Performance analysis:}
For easier reference, we also list the baseline models \enquote{BCNN\_weak} and \enquote{BT\_weak} from Table~\ref{tab:weakly_supervised_sod} and Table~\ref{tab:reliable_weakly_sod}, and the CNN and transformer backbone based stochastic models are built upon the two baseline models respectively. Same as the results of fully supervised SOD models in Table~\ref{tab:reliable_rgb_sod}, the three types of generative models for weakly-supervised SOD can also achieve comparable deterministic performance.

\subsection{Discussions}
\label{sec:model_discussion}
% \subsection{Model Analysis}
% In this section, w
We further analyze our transformer backbone-based models in details. Unless otherwise stated, the experiments are based on the fully supervised deterministic RGB SOD (\enquote{$\text{B'\_{tr}}$} in Table \ref{tab:fully_rgb_sod_expeiments}).
% We highlight the result in \textcolor{blue}{blue} if it's better than our final result.

\noindent\textbf{Model performance \wrt~optimizer:} We observe that
% the transformer backbone is sensitive to the
% % \NB{the} 
% optimizer, and the
% \NB{the} 
the AdamW optimizer is more suitable to train the transformer backbone (Swin~\cite{liu2021swin} in particular) based framework compared with SGD.
% based network is sensitive to the initial learning rate. We then analyze the loss convergence of the transformer backbone-based network as shown in Fig.~\ref{fig:LossFigure}, which clearly shows that a larger learning rate may lead to significant instability in training. Further, we want to analyze how conventional backbone and transformer backbone perform with different optimizers. 
To explain this,
% we change our backbone network to ResNet50 \cite{resnet}, and obtain model performance as shown in Table \ref{tab:model_analysis} \enquote{ResNet50}, which uses Adam as optimizer. Then, 
we train $\text{B'\_{cnn}}$
% (CNN backbone model) 
and $\text{B'\_{tr}}$
% (transformer backbone model) 
with SGD as optimizer, leading to $\text{B'\_{SGD}}$ and $\text{B'\_{trSGD}}$ respectively in Table \ref{tab:model_analysis}.
% the model with SGD as optimizer, which is \enquote{ResNet50\_SGD}. 
We observe that for both the CNN and transformer backbone based networks,
% and transformer backbone based network,
% ($\text{B'\_{cnn}}$), 
the SGD optimizer usually achieves worse performance compared with the
% ($\text{B'\_{SGD}}$) and 
AdamW optimizer.
% We further visualize the loss convergence curve of transformer backbone based models in Fig.~\ref{fig:LossFigure}. 
Note that models with the two types of optimizers share the same initial learning rate, and \enquote{SGD} in this paper is SGD with a momentum of 0.9. We find that after the first epoch, the AdamW optimizer based model jumps directly to a minimum of smaller loss compared with SGD, and later, the loss decrease behaviors of both models are similar. We also tried different learning rate configurations for models with the two types of the optimizers, and the performance of SGD based model is still bad. We further explore whether the AdamW converges faster than SGD. However, even when we train more epochs for SGD based model, the conclusion is still similar.
% ($\text{B'\_{cnn}}$) achieve similar performance. However, for the transformer backbone-based network, the Adam optimizer (\enquote{Ours\_F}) achieves significantly better performance than the model with SGD optimizer (\enquote{Ours\_SGD}). 
We will investigate it further to extensively explain the different model behaviors with various types of optimizers.

\noindent\textbf{The importance of initialization weights:}
% \noindent\textbf{The importance of transformer network as backbone:}
% We
% % train with same decoder, but 
% change our transformer backbone to
% % VGG16 \cite{vgg_network} and 
% ResNet50 \cite{resnet}, and show the model performance in Table \ref{tab:model_analysis} as
% % \enquote{VGG16} and 
% \enquote{ResNet50}.
% % respectively in the \enquote{Different Backbones} section.
% The significant performance improvement of \enquote{Ours\_F} explains the effectiveness of transformer backbone for SOD.
% our transformer backbone based network (\enquote{Ours}) further explains the superior performance of our network. 
For both CNN and transformer backbone, we initialize them with the image classification model trained on the ImageNet-1K \cite{imagenet_1k} dataset. To test how the initialization weights contribute to the model performance, we randomly initialize the two models ($\text{B'\_{cnn}}$ and $\text{B'\_{tr}}$ in Table \ref{tab:fully_rgb_sod_expeiments})
% in Table \ref{tab:backbone_sod} 
and obtain model performance as
% model of random initialization in Table \ref{tab:backbone_sod} as 
$\text{B'\_{R}}$ and $\text{B'\_{trR}}$ in Table \ref{tab:model_analysis}. We observe the worse performance of both $\text{B'\_{R}}$ and $\text{B'\_{trR}}$, which further illustrates the necessity of fine-tuning the backbone models for SOD. We also initialize our transformer backbone with parameters pre-trained on the ImageNet-22K dataset and show the result as $\text{B'\_{tr22K}}$ in Table \ref{tab:model_analysis}. The better performance of $\text{B'\_{tr22K}}$ compared with $\text{B'\_{tr}}$
% initialized with parameters pre-trained on ImageNet-1K
again explains the importance of the initialization weights.

\noindent\textbf{Different transformer backbones analysis:}
Following our pipeline, we change the Swin transformer backbone \cite{liu2021swin} to the ViT backbone \cite{dosovitskiy_ViT_ICLR_2021,dpt_transformer}, and achieve $\text{B'\_{ViT}}$ in Table \ref{tab:model_analysis}.
% respectively, where \enquote{Base\_DPT} has the same network structure as \enquote{Base} in Table \ref{tab:model_analysis} except that we change the backbone to DPT \cite{dpt_transformer}, and \enquote{DPT} has all our proposed components, \eg~deep supervision and difficulty-aware learning. 
Note that, the ViT backbone we used in Table \ref{tab:model_analysis} is initialized with weights trained on ImageNet-22K. The comparable performance of $\text{B'\_{ViT}}$ compared with $\text{B'\_{tr22K}}$ explains that the two types of backbones both work for SOD.
% better performance of $\text{B'\_{ViT}}$
% \enquote{DPT} 
% compared with $\text{B'\_{tr}}$ may comes from either the backbone or the larger training dataset for the initialization weights.
% for the initialized weights.
% illustrates the effectiveness of the two strategies. Although the DPT backbone \cite{dpt_transformer} based model (\enquote{DPT}) outperforms the Swin backbone \cite{liu2021swin} based model (\enquote{Ours\_F}), our model with Swin backbone initialized with ImageNet-22K  (\enquote{Ours\_22k}) achieve comparable performance as \enquote{DPT}, which 
% further explains the importance of initial weights for the backbone network.
% illustrates the effectiveness of the Swin backbone for salient object detection.

\begin{figure}[!ht]
\begin{center}
    \begin{tabular}{c@{ } c@{ } c@{ } c@{ } c@{ } c@{ }}
    \includegraphics[width=0.45\linewidth]{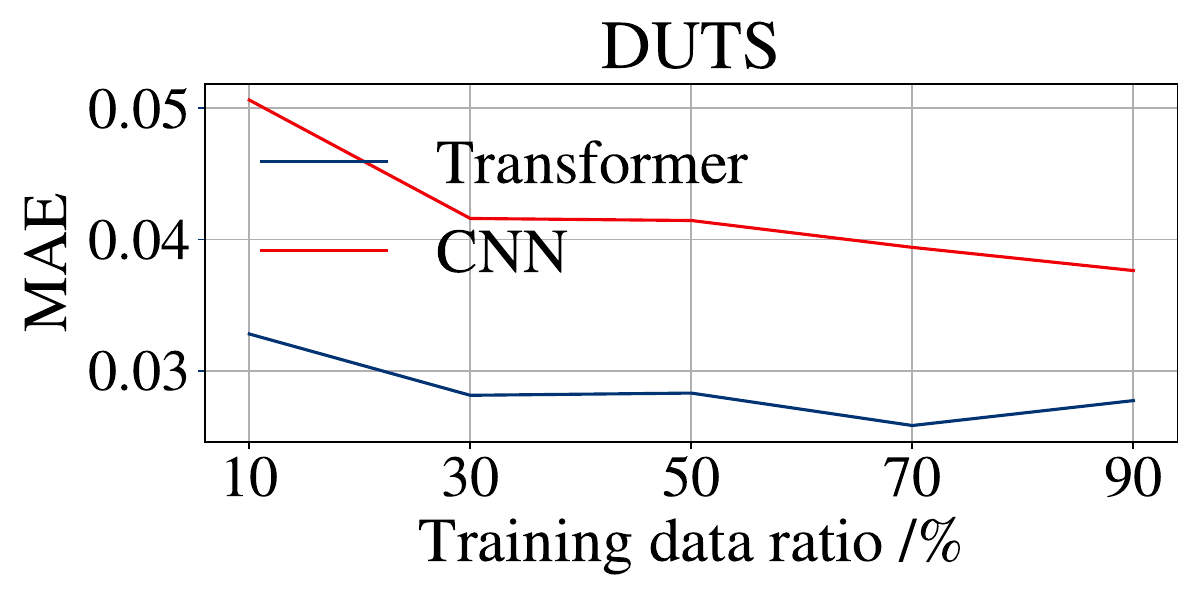} &
    \includegraphics[width=0.45\linewidth]{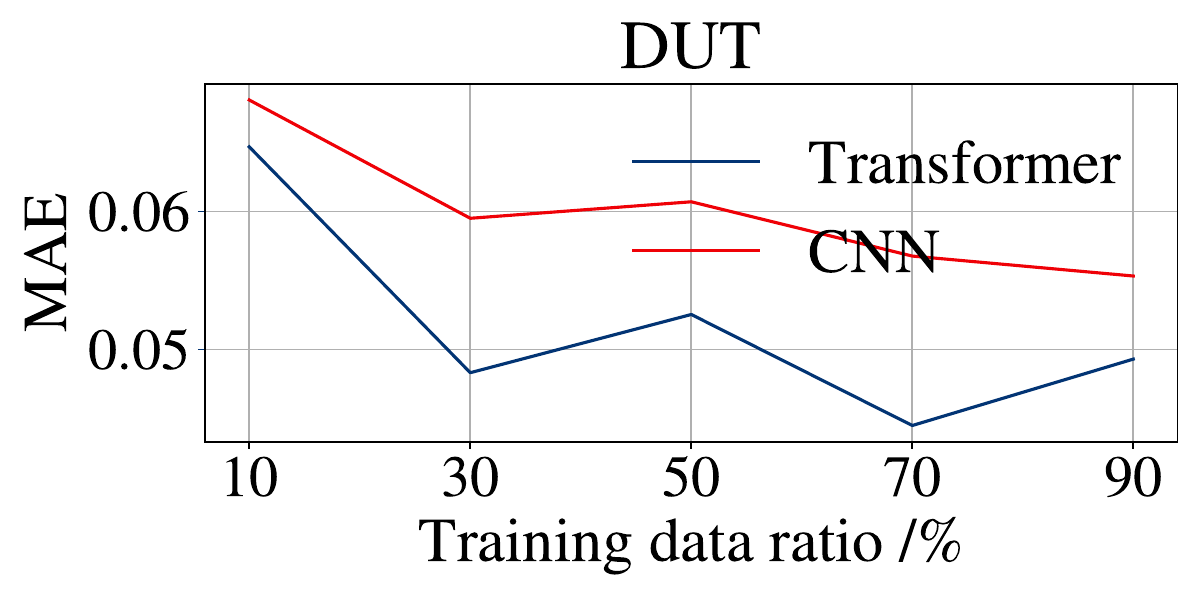} \\
    \includegraphics[width=0.45\linewidth]{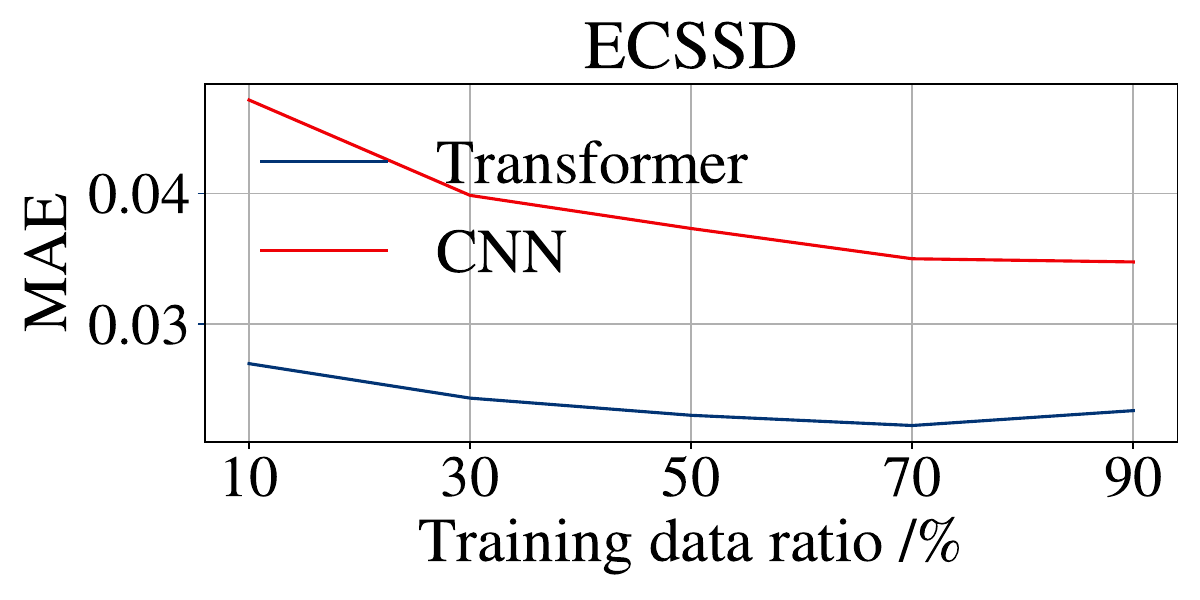} &
    \includegraphics[width=0.45\linewidth]{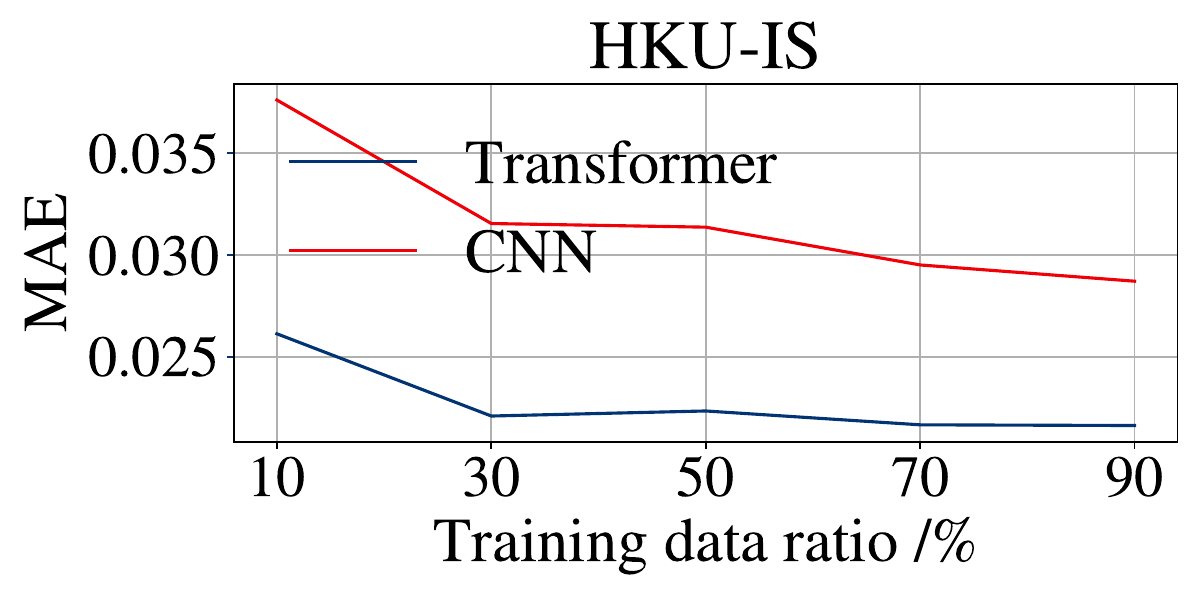} \\
    \includegraphics[width=0.45\linewidth]{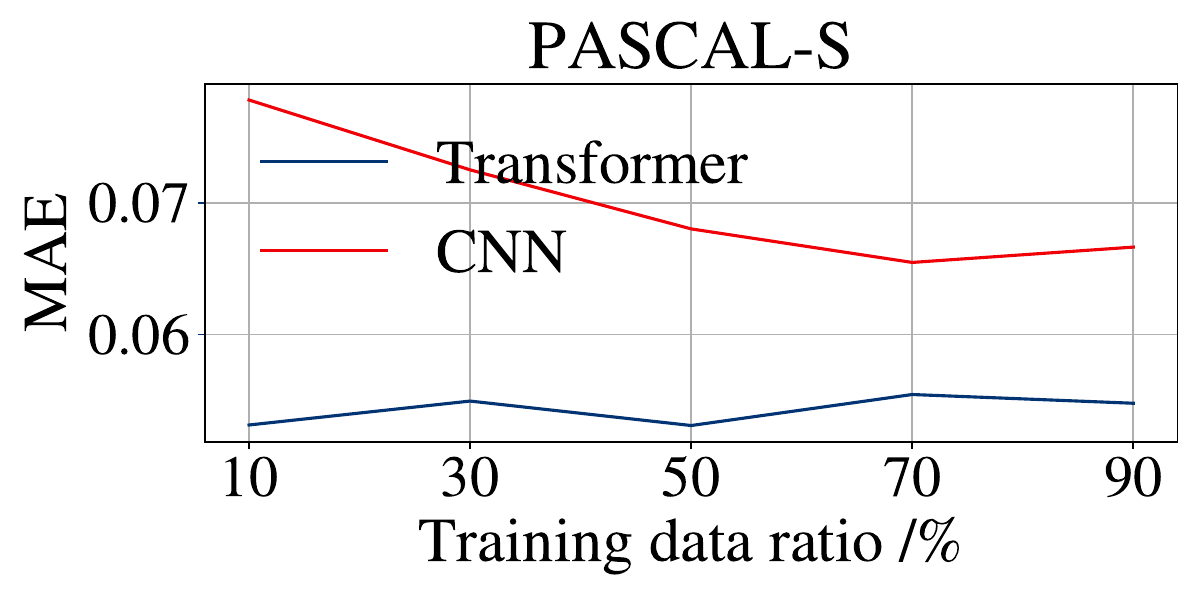} &
    \includegraphics[width=0.45\linewidth]{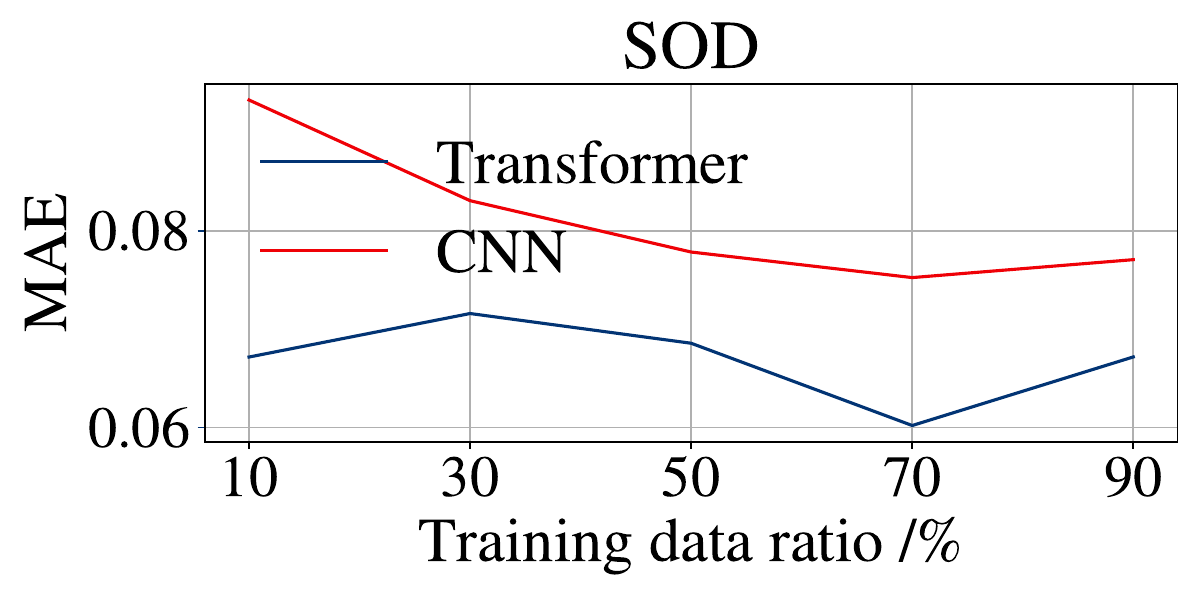} \\
    \end{tabular}
\end{center}
\caption{Model performance of the CNN backbone and the transformer backbone
% based salient object detection network
\wrt different training dataset sizes on six testing datasets.}
    \label{fig:performance_wrt_training_size}
    % \vspace{-4mm}
\end{figure}

\noindent\textbf{Model performance \wrt~training datasets scales:} As the two types of backbones
% (CNN and transformer) 
have significantly different numbers of model parameters, leading to different model capacities. We aim to analyze how model capacity is sensitive to scales of the training dataset.
We then train our transformer backbone networks ($\text{B'\_{tr}}$) and CNN backbone based model ($\text{B'\_{cnn}}$) in Table \ref{tab:fully_rgb_sod_expeiments}) with different sizes
% \NB{sizes} \sout{numbers} 
of training datasets, which are 10\%, 30\%, 50\%, 70\%, 90\% of the entire training dataset respectively, and report the model performance in Fig.~\ref{fig:performance_wrt_training_size}. The consistently better performance of the transformer backbone-based model with regard to different numbers of training examples
% \NB{examples?}dataset 
explains its effectiveness. Meanwhile, we observe that the model performance is not always increasing with a larger training dataset, which inspires us to work on an active learning-based transformer network to actively select representative samples for model training.

\begin{table*}[t!]
  \centering
  \scriptsize
  \renewcommand{\arraystretch}{1.1}
  \renewcommand{\tabcolsep}{0.715mm}
  \caption{\footnotesize{\Rev{Model robustness to adversarial attack and defense.}}}
  \begin{tabular}{l|cccc|cccc|cccc|cccc|cccc|cccc}
  \hline
% \toprule
  &\multicolumn{4}{c|}{DUTS \cite{imagesaliency}}&\multicolumn{4}{c|}{ECSSD \cite{yan2013hierarchical}}&\multicolumn{4}{c|}{DUT \cite{Manifold-Ranking:CVPR-2013}}&\multicolumn{4}{c|}{HKU-IS \cite{li2015visual}}&\multicolumn{4}{c|}{PASCAL-S \cite{pascal_s_dataset}}&\multicolumn{4}{c}{SOD \cite{sod_dataset}} \\
    Method & $S_{\alpha}\uparrow$&$F_{\beta}\uparrow$&$E_{\xi}\uparrow$&$\mathcal{M}\downarrow$& $S_{\alpha}\uparrow$&$F_{\beta}\uparrow$&$E_{\xi}\uparrow$&$\mathcal{M}\downarrow$& $S_{\alpha}\uparrow$&$F_{\beta}\uparrow$&$E_{\xi}\uparrow$&$\mathcal{M}\downarrow$& $S_{\alpha}\uparrow$&$F_{\beta}\uparrow$&$E_{\xi}\uparrow$&$\mathcal{M}\downarrow$& $S_{\alpha}\uparrow$&$F_{\beta}\uparrow$&$E_{\xi}\uparrow$&$\mathcal{M}\downarrow$& $S_{\alpha}\uparrow$&$F_{\beta}\uparrow$&$E_{\xi}\uparrow$&$\mathcal{M}\downarrow$ \\ \hline
    $\text{B'\_{cnn}}$ &.882 &.840 &.916 &.037 &.922 &.919 &.947 &.035 &.823 &.742 &.851 &.057 &.912 &.901 &.947 &.030 &.855 &.841 &.896 &.065 &.832 &.825 &.863 &.073  \\
    $\text{CIGAN}$ &.876 &.820 &.906 &.042 &.923 &.913 &.945 &.037 &.823 &.733 &.848 &.061 &.911 &.892 &.943 &.034 &.856 &.836 &.893 &.068 &.833 &.816 &.862 &.075   \\ 
    $\text{B'\_{tr}}$ &{.911} &{.882} &{.947} &{.026} &{.939} &{.940} &{.965} &{.024} &.860 &{.801} &{.894} &{.045} &{.927} &{.921} &{.964} &{.023} &{.876} &{.872} &{.917} &{.053} &.858 &.853 &{.897} &{.059} \\
    $\text{TIGAN}$ &{.909} &.873 &.941 &{.028} &{.941} &.936 &{.964} &{.025} &{.861} &.796 &.890 &{.047} &{.929} &{.918} &{.962} &.025 &{.879} &{.869} &{.916} &{.054} &{.861} &{.854} &.894 &.060   \\ \hline
  \multicolumn{24}{c}{Performing FGSM~\cite{goodfellow2014explaining} Attack} \\ \hline
  $\text{AB'\_{cnn}}$ &.782 &.709 &.824 &.082 &.837 &.829 &.874 &.078 &.714 &.594 &.744 &.113 &.838 &.818 &.889 &.061 &.771 &.747 &.819 &.113 &.713 &.680 &.747 &.134 \\
  $\text{ACIGAN}$ &.786 &.692 &.813 &.092 &.858 &.833 &.883 &.076 &.730 &.601 &.753 &.119 &.853 &.814 &.887 &.065 &.789 &.758 &.824 &.113 &.737 &.690 &.774 &.136 \\
  $\text{AB'\_{tr}}$ &.802 &.744 &.872 &.065 &.844 &.840 &.898 &.068 &.746 &.652 &.808 &.093 &.840 &.823 &.910 &.056 &.779 &.759 &.842 &.100 &.746 &.734 &.807 &.113 \\
  $\text{ATIGAN}$ &.881 &.837 &.921 &.040 &.916 &.912 &.946 &.037 &.841 &.773 &.879 &.056 &.909 &.897 &.950 &.033 &.856 &.851 &.899 &.067 &.848 &.847 &.891 &.067 \\ \hline
  \multicolumn{24}{c}{Performing Adversarial Training~\cite{madry2018towards} based Defense} \\ \hline
  $\text{DB'\_{cnn}}$ &.822 &.761 &.862 &.063 &.872 &.868 &.906 &.059 &.756 &.649 &.786 &.088 &.870 &.855 &.916 &.047 &.806 &.787 &.850 &.093 &.765 &.740 &.798 &.105 \\
  $\text{DCIGAN}$ &.838 &.775 &.872 &.059 &.891 &.881 &.918 &.053 &.781 &.676 &.808 &.084 &.886 &.866 &.922 &.044 &.826 &.806 &.867 &.088 &.783 &.759 &.819 &.104 \\
  $\text{DB'\_{tr}}$ &.861 &.822 &.909 &.045 &.902 &.905 &.939 &.041 &.802 &.726 &.845 &.070 &.893 &.887 &.941 &.036 &.827 &.819 &.877 &.078 &.797 &.793 &.842 &.090 \\
  $\text{DTIGAN}$ &.906 &{.874} &{.942} &.030 &{.939} &{.938} &{.964} &{.025} &{.864} &{.806} &{.898} &.048 &{.929} &{.921} &{.964} &{.024} &.873 &{.869} &.915 &.056 &{.866} &{.866} &{.903} &{.058} \\ 
     \hline
  \end{tabular}
  \label{tab:model_analysis_attack_defense}
%   \vspace{-5mm}
\end{table*}

\begin{figure*}[tp]
%  \vspace{-5mm}
  \begin{center}
  \setlength\tabcolsep{2pt}
  \begin{tabular}{*{2}{p{0.092\textwidth}<{\centering}} | *{4}{p{0.092\textwidth}<{\centering}} | *{4}{p{0.092\textwidth}<{\centering}}}
  {\includegraphics[width=\linewidth]{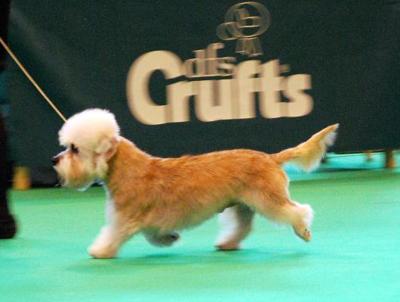}}
  &{\includegraphics[width=\linewidth]{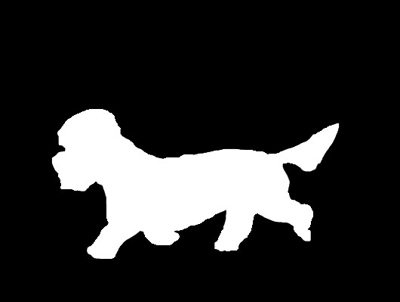}}
  &{\includegraphics[width=\linewidth]{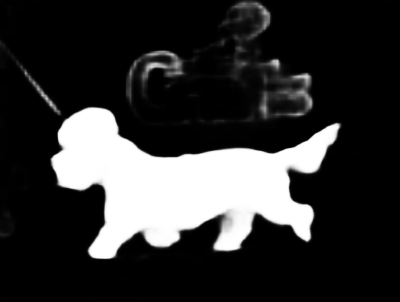}}
  &{\includegraphics[width=\linewidth]{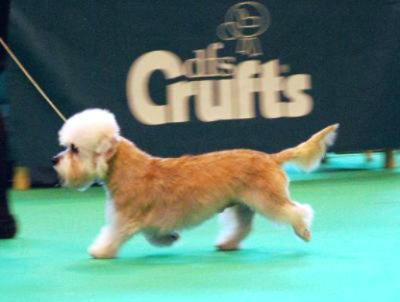}}
  &{\includegraphics[width=\linewidth]{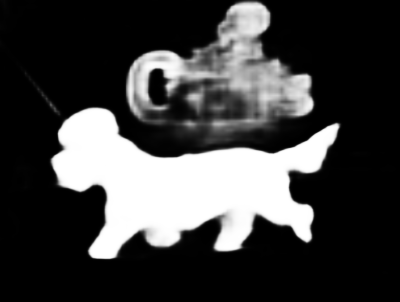}}
  &{\includegraphics[width=\linewidth]{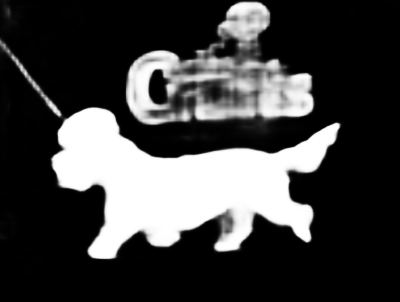}}
  &{\includegraphics[width=\linewidth]{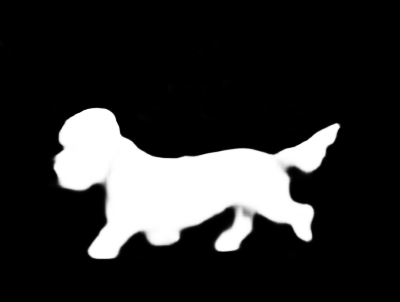}}
  &{\includegraphics[width=\linewidth]{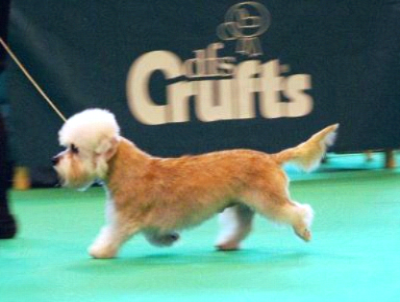}}
  &{\includegraphics[width=\linewidth]{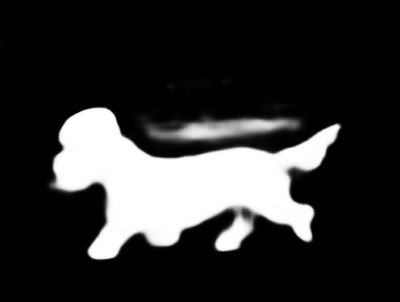}}
  &{\includegraphics[width=\linewidth]{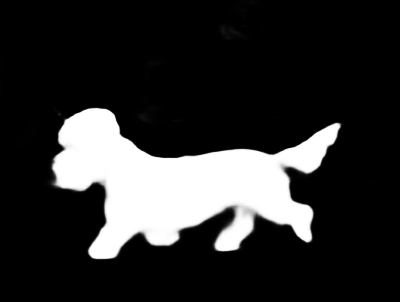}}\\
  % {\includegraphics[width=\linewidth]{adv_vis/sun_bxivvymiinsxtren/0.jpg}}
  % &{\includegraphics[width=\linewidth]{adv_vis/sun_bxivvymiinsxtren/1.png}}
  % &{\includegraphics[width=\linewidth]{adv_vis/sun_bxivvymiinsxtren/2.png}}
  % &{\includegraphics[width=\linewidth]{adv_vis/sun_bxivvymiinsxtren/3.png}}
  % &{\includegraphics[width=\linewidth]{adv_vis/sun_bxivvymiinsxtren/4.png}}
  % &{\includegraphics[width=\linewidth]{adv_vis/sun_bxivvymiinsxtren/5.png}}
  % &{\includegraphics[width=\linewidth]{adv_vis/sun_bxivvymiinsxtren/6.png}}
  % &{\includegraphics[width=\linewidth]{adv_vis/sun_bxivvymiinsxtren/7.png}}
  % &{\includegraphics[width=\linewidth]{adv_vis/sun_bxivvymiinsxtren/8.png}}
  % &{\includegraphics[width=\linewidth]{adv_vis/sun_bxivvymiinsxtren/9.png}}\\
%   \footnotesize{$x$}&\footnotesize{$y$}&\footnotesize{$s$}&\footnotesize{$x_{adv}$}&\footnotesize{$s_{adv}$}&\footnotesize{$s_{def}$}\\
\multicolumn{1}{c}{\footnotesize{$x$}}&\multicolumn{1}{c}{\footnotesize{$y$}}&\multicolumn{4}{c}{\footnotesize{$\text{CIGAN}$}}&\multicolumn{4}{c}{\footnotesize{$\text{TIGAN}$}}\\ 
  \end{tabular}
  \end{center}
%   \vspace{-5pt}
  \caption{\footnotesize{\Rev{Model robustness to adversarial attack, \ie~FGSM~\cite{goodfellow2014explaining} attack, where the samples (from left to right) within each method ($\text{CIGAN}$ and $\text{TIGAN}$) are the prediction of $x$, the adversarial sample $x_{adv}$, its prediction $s_{adv}$, and the prediction of $x_{adv}$ after adversarial training.}}
  }
\label{fig:adversarial_attack_analysis}
\end{figure*}

\noindent\textbf{Model performance with different decoders:} To test how the transformer encoder performs with different decoders, we change the backbone of existing SOD models (SCRN~\cite{scrn_sal} and F3Net~\cite{wei2020f3net}) to transformer backbone \cite{liu2021swin}, and show their performance in Table~\ref{tab:benchmark_rgb_models_ours}, where \enquote{*} is the transformer backbone based counterpart. Table~\ref{tab:benchmark_rgb_models_ours} shows that the transformer backbone can indeed improve the performance of existing SOD models. However, we observe similar performance of model with our decoder ($\text{B'\_{tr}}$) (around 1M parameters) compared with other complicated decoders (more than 20M for both SCRN~\cite{scrn_sal} and F3Net~\cite{wei2020f3net}). The Swin backbone model~\cite{liu2021swin} has around 85M parameters, and its high capacity poses challenges to the decoder design. We argue that the transformer-compatible decoder should be investigated to further explore the contribution of transformer backbones. 
% \YC{decoder?}.

\noindent\Rev{\textbf{Model performance with conventional uncertainty estimation techniques:} A systematic way to deal with model uncertainty is via Bayesian statistics~\cite{Bayesian_Uncertainty,bayesian_conv, generalized_bnn,kendall2017uncertainties,uncertainty_decomposition}. Bayesian Neural Networks aim to learn a distribution over each of the network parameters by placing a prior probability distribution over network weights, \ie~$p(\theta|D)$.
% , which is set as a
% constraint
% to regularize the distribution of the model parameters and achieve stochastic prediction.   
According to the Bayesian rule, the posterior over model parameters $p(\theta|x,y)$ (or $p(\theta|D)$) can be achieved as:
\begin{equation}
% \begin{aligned}
     p(\theta|x,y)=\frac{p(x,y|\theta)p(\theta)}{p(x,y)}
     =\frac{p(x,y|\theta)p(\theta)}{\int p(x,y|\theta)p(\theta)d\theta}.
% \end{aligned}
\end{equation}
The marginalization over $\theta$ to calculate $p(x,y)$ in the denominator is intractable. $p(\theta|x,y)$ is then not available in closed-form, making it computationally intractable to calculate the exact Bayesian posterior. More efforts have been put into developing approximations of Bayesian Neural Network that can work in practice, including Variational Inference (VI)~\cite{Jordan99anintroduction,Graphical_Models_VI,lee2022gibbs} and Markov Chain Monte Carlo (MCMC)~\cite{mcmc_sampling}. \cite{gal2016dropout} shows that a neural network of arbitrary depth and non-linearity, with dropout applied before every weighted layer, is mathematically equivalent to an approximation to the probabilistic deep Gaussian process~\cite{deep_gaussian_process} (GP). Based on it,~\cite{gal2016dropout} further shows that the dropout objective minimizes the KL divergence between an approximate distribution and the posterior of a deep Gaussian process. In this paper, we apply MC-dropout~\cite{gal2016dropout} as a free-lunch model uncertainty estimation technique to our baseline model $\text{B'\_{tr}}$ in Table~\ref{tab:reliable_rgb_sod} with dropout rate 0.3, and show its performance in Table~\ref{tab:mc_dropout_analysis}, which again verify the effectiveness of our model in achieving better model calibration.}

\noindent\Rev{\textbf{Robustness to adversarial attack:} Deep neural network based models are known to suffer from adversarial examples. With small perturbations, model predictions can be changed drastically~\cite{szegedy2013intriguing}. Common defense methods for adversarial attacks include adversarial training~\cite{madry2018towards}, certified robustness~\cite{wong2018provable}, \etc In this paper, we investigate model robustness with respect to adversarial attack. Specifically, we discuss FGSM~\cite{goodfellow2014explaining}, a gradient based attack, and perform adversarial training~\cite{madry2018towards} to achieve model defense.}

\begin{figure}[tp]
%  \vspace{-5mm}
  \begin{center}
  \begin{tabular}{c@{ } c@{ } c@{ } c@{ } c@{ } c@{ }}
  {\includegraphics[width=0.15\linewidth]{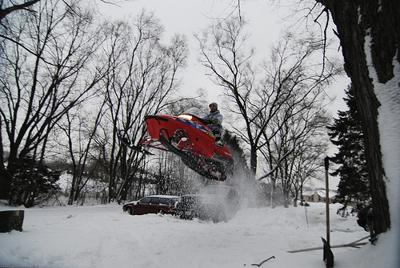}} &
  {\includegraphics[width=0.15\linewidth]{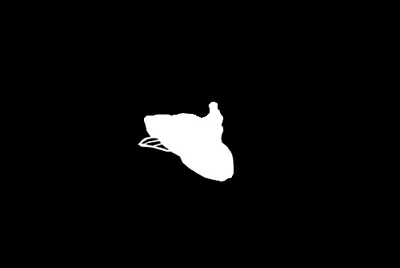}} &
  {\includegraphics[width=0.15\linewidth]{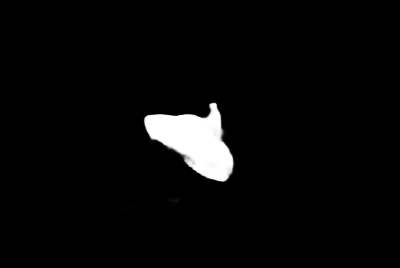}} &
  {\includegraphics[width=0.15\linewidth]{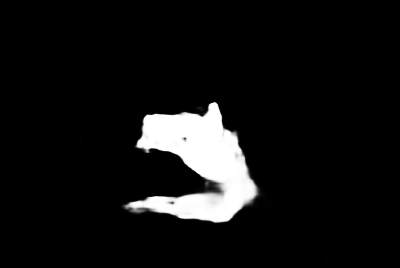}}
  &
  {\includegraphics[width=0.15\linewidth]{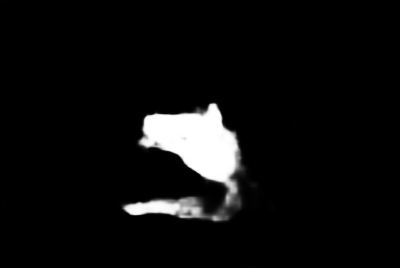}} &
  {\includegraphics[width=0.15\linewidth]{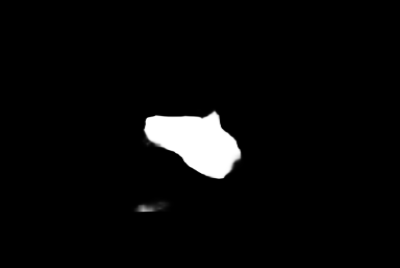}}\\
  {\includegraphics[width=0.15\linewidth]{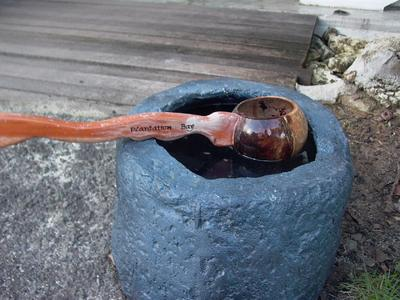}} &
  {\includegraphics[width=0.15\linewidth]{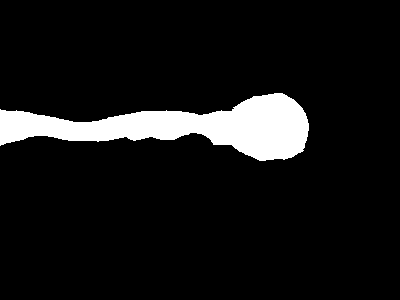}} &
  {\includegraphics[width=0.15\linewidth]{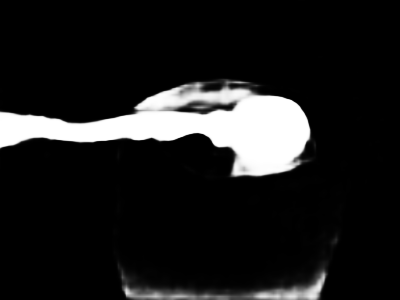}} &
  {\includegraphics[width=0.15\linewidth]{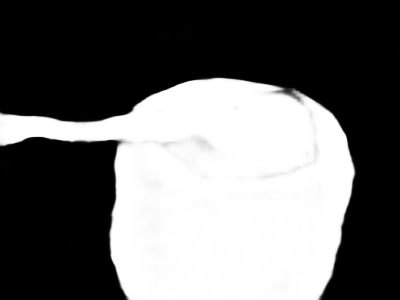}}
  &
  {\includegraphics[width=0.15\linewidth]{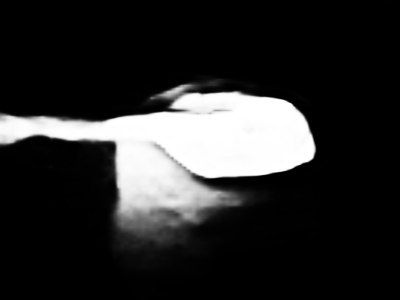}} &
  {\includegraphics[width=0.15\linewidth]{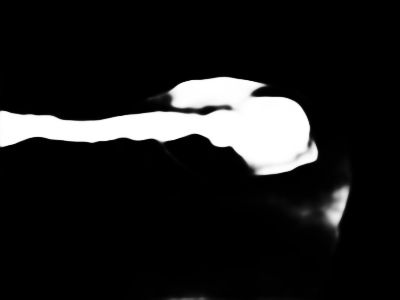}}\\
%   {\includegraphics[width=0.23\linewidth]{figures/swin_fail/rgb_dir/sun_bezpaipecjkglcax.png}} &
%   {\includegraphics[width=0.23\linewidth]{figures/swin_fail/gt_dir/sun_bezpaipecjkglcax.png}} &
%   {\includegraphics[width=0.23\linewidth]{figures/swin_fail/cnn_good_dir/sun_bezpaipecjkglcax.png}} &
%   {\includegraphics[width=0.23\linewidth]{figures/swin_fail/swin_bad_dir/sun_bezpaipecjkglcax.png}}\\
  \footnotesize{Image}&\footnotesize{GT}&\footnotesize{$\text{B'\_{cnn}}$ }&\footnotesize{$\text{B'\_{tr}}$ }&\footnotesize{CIGAN}&\footnotesize{CGAN}\\
  \end{tabular}
  \end{center}
%   \vspace{-5pt}
  \caption{\footnotesize{Failure cases of the transformer backbone compared with the CNN backbone ($\text{B'\_{cnn}}$ and $\text{B'\_{tr}}$), and iGAN compared with CGAN within the CNN backbone (CIGAN and CGAN) for RGB SOD.}
  }
\label{fig:failure_case_transformer}
\end{figure}

\Rev{FGSM~\cite{goodfellow2014explaining} attack only needs to do backprop once to get the gradient of classification loss with respect to the input $x$, and the adversarial sample $x_{adv}$ can be generated via:
% The equation to generate the FGSM attack is shown as:
\begin{equation}
\label{fgsm_attack}
    x_{adv} = x + \varepsilon \operatorname{sign}(\nabla_x \mathcal{L}(\theta,x,y)),
\end{equation}
where $\mathcal{L}(\theta,x,y))$ is the classification loss, and the sign function is used to achieve faster convergence. Correspondingly, the adversarial training based defense is achieved via training the model with adversarial sample pair $(x_{adv},y)$, leading to a new objective:
\begin{equation}
\label{eq:minmax}
\min_\theta\mathbb{E}_{(x,y)\sim\mathcal{D}}\left[\max_{\delta\in 
    \mathcal{S}}
    \mathcal{L}(\theta,x+\delta,y)\right] \; ,
\end{equation}
where $\mathcal{S}$ is the candidate adversarial attacks. In practice, a more efficient way to achieve adversarial training based defense is thorough joint training with both clean sample $x$ and adversarial sample $x_{adv}$ with weighted loss:
\begin{equation}
\label{eq:minmax_practice}
\min_\theta (\alpha\mathcal{L}(\theta,x,y)+(1-\alpha)\mathcal{L}(\theta,x_{adv},y))\; ,
\end{equation}
where $\alpha=0.5$ is used to control the balance of accurate prediction ($\mathcal{L}(\theta,x,y)$) and model robustness ($\mathcal{L}(\theta,x_{adv},y)$).}

\Rev{In this paper, we perform adversarial attack with FGSM~\cite{goodfellow2014explaining} and defense with adversarial training in Eq.~\eqref{eq:minmax_practice} to both the baseline models ($\text{B'\_{cnn}}$ and $\text{B'\_{tr}}$ in Table~\ref{tab:reliable_rgb_sod}) and the proposed iGAN based frameworks ($\text{CIGAN}$ and $\text{TIGAN}$ in Table~\ref{tab:reliable_rgb_sod}). Specifically, based on the above models, we set $\varepsilon=8/255$ in 
Eq.~\eqref{fgsm_attack} following~\cite{goodfellow2014explaining} to generate adversarial samples $x_{adv}$ of each training image, which will be used to train the above models again together with the clean sample $x$ to achieve the defense process. We report model performance in Table~\ref{tab:model_analysis_attack_defense}. Note that the performance of attacked models is obtained by performing FGSM~\cite{goodfellow2014explaining} attack on the testing samples, where the adversarial testing samples are fed to the specific model to generate the predictions. We show in Fig.~\ref{fig:adversarial_attack_analysis} the clean sample $x$ and its prediction $s$, the adversarial sample $x_{adv}$ and its prediction $s_{adv}$, and the prediction of $x$ and $x_{adv}$ after the defense.
Fig.~\ref{fig:adversarial_attack_analysis} and Table~\ref{tab:model_analysis_attack_defense} show that invisible adversarial attack~\cite{goodfellow2014explaining} can cause significant performance degradation, which can be partially solved with the adversarial training strategies~\cite{madry2018towards}. We also observe the slightly robust performance of the proposed iGAN frameworks ($\text{CIGAN}$ and $\text{TIGAN}$) compared with the baseline models, which further explain the robustness of the proposed generative model.}

\noindent\textbf{Failure Case Analysis:}
To further investigate the limitations of both the transformer backbone and the iGAN framework, we look deeper in Table~\ref{tab:reliable_rgb_sod} and Table~\ref{tab:reliable_rgbd_sod}, and the predictions of each related model. We find out two main issues: 1) the transformer backbone does not always perform superior to the CNN backbone; 2) our iGAN model can lead to over-smoothed predictions compared with CGAN due to the diverse generation process.
% not always performs superior compared with the CGAN counterparts (see \enquote{CGAN} and \enquote{CIGAN} for the CNN backbone based models, and \enquote{TGAN} and \enquote{TIGAN} for the transformer backbone based models in Table~\ref{tab:reliable_rgb_sod} and Table~\ref{tab:reliable_rgbd_sod}). We then analyze: 1) when our transformer backbone based model fails, and 2) when the iGAN framework fails to outperform the CGAN model. 

For the former, to exclude the influence of the iGAN framework, we compare the predictions of the CNN backbone ($\text{B'\_{cnn}}$) and transformer backbone ($\text{B'\_{tr}}$) for RGB SOD, and show samples in Fig.~\ref{fig:failure_case_transformer}.
% and Fig.~\ref{fig:failure_case_igan_transformer}, 
% where Fig.~\ref{fig:failure_case_transformer} 
% which compares the prediction of the CNN backbone ($\text{B'\_{cnn}}$) and transformer backbone ($\text{B'\_{tr}}$) for RGB SOD.
% and in Fig.~\ref{fig:failure_case_igan_transformer}, we compare the proposed generative model with the deterministic transformer backbone based framework.
% , where \enquote{ResNet50} is the ResNet50 model for salient object detection in Table \ref{tab:model_analysis}. 
We observe more false positives within $\text{B'\_{tr}}$,
% the main less accurate predictions are those with too many false positives, 
which can be explained as the \enquote{double-edged sword} of the transformer backbone. On the one hand, the larger receptive field of the transformer makes it superior in localizing the larger salient foreground. On the other hand,
% due to its larger receptive field, 
the less salient objects that expand for a larger region can be falsely detected as positive foreground. We argue that salient object ranking \cite{amirul2018revisiting} can be beneficial in identifying the less salient regions by providing extra saliency degree evaluation. \Rev{For the latter, as the latent variable $z$ within iGAN is conditioned on input $x$, leading to informative latent space, where the predictive distribution $p(y^*|x^*,\theta,z)$ for input $x^*$ has larger variance compared with the CGAN based framework. In this case, the averaged prediction is
% deterministic prediction is obtained by averaging the diverse predictions, leading to 
over-smoothed. Similarly, this issue can also be fixed with saliency ranking~\cite{amirul2018revisiting}.}
% , avoiding the false positive detection due to .
% This can be caused by at least two reasons. Firstly, the generated binary ground truth might be biased, which focuses mostly on person (images in the $2^{nd}$, $3^{rd}$, $4^{th}$ rows). Secondly, the subtle contrast difference for the foreground regions is difficult to be perceived by the deep model (image in the $1^{st}$ row),
% which makes salient object ranking \cite{amirul2018revisiting} an interesting direction to work on. 
% We will work on it in the future.

\section{Conclusion and Future Work}
In this paper,
% we have studied the generative transformer for accurate and reliable salient object detection.
% % context information learning problem under the transformer backbone for salient object detection (SOD).
% %we adopt transformer backbone \cite{liu2021swin} for salient object detection (SOD) to effectively exploit the context information.
% Specifically, 
we proposed an inferential GAN within the transformer framework for both fully and weakly supervised SOD. Different from typical GANs that define the prior distribution as a standard normal distribution, we inferred the latent variable via Langevin Dynamics \cite{mcmc_langevin}, a gradient based MCMC, leading to the image-conditioned prior distribution.  
% a unified framework for fully-supervised RGB image-based SOD, RGB-D image pair-based SOD and weakly-supervised RGB image-based SOD.
% , and achieve three new benchmark for each task. 
Through extensive experiments, we observed that a larger receptive field of the transformer leads to its better performance on images with larger salient objects (see Fig.~\ref{fig:performance_wrt_salient_object_size}). However, we also found the double-edged sword effect of the larger receptive field that leads to serious false positives (see Fig.~\ref{fig:failure_case_transformer}). 
% \sout{Besides it} \NB{Further}
Further, for RGB-D SOD,\Rev{we found that the various depth sensors lead to a domain gap between the training dataset and the testing dataset (see Table~\ref{tab:existing_rgbd_dataset} and Fig.~\ref{fig:depth_global_contrast}).
% contrast distributions lead to different depth contribution for RGB-D SOD. 
We then presented auxiliary depth module, leading to consistent depth contribution (see Table \ref{tab:rgbd_sod_analysis})}. For weakly supervised SOD, we observed that the accurate structure information encoded in the transformer backbone as shown in Fig.~\ref{fig:weak_feature_visualization} makes it powerful in generating structure-preserving predictions (see Table \ref{tab:weakly_supervised_sod}).
Extensive experimental results demonstrate the superiority of our transformer backbone-based generative network, achieving new benchmarks with reliable uncertainty maps. 

Our proposed generative model aims to estimate the reliability of saliency prediction with uncertainty maps, \Rev{which also show superiority in achieving robust models (see Table~\ref{tab:model_analysis_attack_defense} and Fig.~\ref{fig:adversarial_attack_analysis}) and well calibrated models (see Table~\ref{tab:mc_dropout_analysis})}.
% As there exists no proper evaluation metrictab:mc_dropout_analysiss for uncertainty map quality estimation, we mainly visualized the uncertainty maps of each generative model.
% \NB{Our} \sout{The} 
Our future work includes two main parts. Firstly, we will apply the produced uncertainty map (see Fig.~\ref{fig:visualization_reliable_sod_methods}
% and Fig.~\ref{fig:viaulization_weakly_supervised_reliable}
) to the saliency generator for effective hard negative mining. In this way, the uncertainty map can not only explain model predictions but also serve as an important prior for effective model learning.
Secondly,
% uncertainty map quality estimation methods are needed to quantitatively analyze the quality of uncertainty maps. 
% Lastly,
we have several hyper-parameters within the inference model, and we observe that our model performance can be influenced by them. We plan to further investigate model performance \wrt those hyper-parameters.

\section*{Acknowledgment}
This work was partly supported by the National Natural Science Foundation of China (62271410, 61871325).

{\small
\bibliographystyle{ieeetr}
\bibliography{Tran_Reference}

\begin{thebibliography}{100}

\bibitem{zhuge2023salient}
M.~Zhuge, D.-P. Fan, N.~Liu, D.~Zhang, D.~Xu, and L.~Shao, ``Salient object
  detection via integrity learning,'' {\em IEEE TPAMI}, 2023.

\bibitem{scrn_sal}
Z.~Wu, L.~Su, and Q.~Huang, ``Stacked cross refinement network for edge-aware
  salient object detection,'' in {\em IEEE ICCV}, 2019.

\bibitem{wei2020f3net}
J.~Wei, S.~Wang, and Q.~Huang, ``F$^3$net: Fusion, feedback and focus for
  salient object detection,'' in {\em AAAI}, 2020.

\bibitem{zhang2021_ucnet}
J.~Zhang, D.-P. Fan, Y.~Dai, S.~Anwar, F.~Saleh, S.~Aliakbarian, and N.~Barnes,
  ``Uncertainty inspired rgb-d saliency detection,'' {\em IEEE TPAMI}, vol.~44,
  no.~9, pp.~5761--5779, 2022.

\bibitem{fan2020bbs}
D.-P. Fan, Y.~Zhai, A.~Borji, J.~Yang, and L.~Shao, ``{BBS-Net}: {RGB-D}
  salient object detection with a bifurcated backbone strategy network,'' in
  {\em ECCV}, 2020.

\bibitem{jing2020weakly}
J.~Zhang, X.~Yu, A.~Li, P.~Song, B.~Liu, and Y.~Dai, ``Weakly-supervised
  salient object detection via scribble annotations,'' in {\em IEEE CVPR},
  2020.

\bibitem{imagesaliency}
L.~Wang, H.~Lu, Y.~Wang, M.~Feng, D.~Wang, B.~Yin, and X.~Ruan, ``Learning to
  detect salient objects with image-level supervision,'' in {\em IEEE CVPR},
  2017.

\bibitem{SOD_Survey_TPAMI_2021}
W.~Wang, Q.~Lai, H.~Fu, J.~Shen, H.~Ling, and R.~Yang, ``Salient object
  detection in the deep learning era: An in-depth survey,'' {\em IEEE TPAMI},
  vol.~44, no.~6, pp.~3239--3259, 2022.

\bibitem{Zhang_2018_CVPR}
J.~Zhang, T.~Zhang, Y.~Dai, M.~Harandi, and R.~Hartley, ``Deep unsupervised
  saliency detection: A multiple noisy labeling perspective,'' in {\em IEEE
  CVPR}, 2018.

\bibitem{deepusps}
D.~T. Nguyen, M.~Dax, C.~K. Mummadi, T.~P.~N. Ngo, T.~H.~P. Nguyen, Z.~Lou, and
  T.~Brox, ``{DeepUSPS}: Deep robust unsupervised saliency prediction with
  self-supervision,'' in {\em NeurIPS}, 2019.

\bibitem{Miao_2021_ACM_MM}
M.~{Zhang}, T.~{Liu}, Y.~{Piao}, S.~{Yao}, and H.~{Lu}, ``Auto-msfnet: Search
  multi-scale fusion network for salient object detection,'' in {\em ACM MM},
  2021.

\bibitem{xu2021locate}
B.~Xu, H.~Liang, R.~Liang, and P.~Chen, ``Locate globally, segment locally: A
  progressive architecture with knowledge review network for salient object
  detection,'' in {\em AAAI}, 2021.

\bibitem{wu2022edn}
Y.-H. Wu, Y.~Liu, L.~Zhang, M.-M. Cheng, and B.~Ren, ``Edn: Salient object
  detection via extremely-downsampled network,'' {\em IEEE TIP}, vol.~31,
  pp.~3125--3136, 2022.

\bibitem{yang2022biconnet}
Z.~Yang, S.~Soltanian-Zadeh, and S.~Farsiu, ``Biconnet: an edge-preserved
  connectivity-based approach for salient object detection,'' {\em Pattern
  Recognition}, vol.~121, p.~108231, 2022.

\bibitem{dmra_iccv19}
Y.~Piao, W.~Ji, J.~Li, M.~Zhang, and H.~Lu, ``Depth-induced multi-scale
  recurrent attention network for saliency detection,'' in {\em IEEE ICCV},
  2019.

\bibitem{Fu2020JLDCF}
K.~Fu, D.-P. Fan, G.-P. Ji, and Q.~Zhao, ``{JL-DCF}: Joint learning and
  densely-cooperative fusion framework for {RGB-D} salient object detection,''
  in {\em IEEE CVPR}, 2020.

\bibitem{jing2021_complementary}
J.~Zhang, D.-P. Fan, Y.~Dai, X.~Yu, Y.~Zhong, N.~Barnes, and L.~Shao, ``Rgb-d
  saliency detection via cascaded mutual information minimization,'' in {\em
  IEEE ICCV}, 2021.

\bibitem{resnet}
K.~He, X.~Zhang, S.~Ren, and J.~Sun, ``Deep residual learning for image
  recognition,'' in {\em IEEE CVPR}, 2016.

\bibitem{niu2012leveraging}
Y.~Niu, Y.~Geng, X.~Li, and F.~Liu, ``Leveraging stereopsis for saliency
  analysis,'' in {\em IEEE CVPR}, 2012.

\bibitem{liu2010sift}
C.~Liu, J.~Yuen, and A.~Torralba, ``Sift flow: Dense correspondence across
  scenes and its applications,'' {\em IEEE TPAMI}, vol.~33, no.~5,
  pp.~978--994, 2010.

\bibitem{peng2014rgbd}
H.~Peng, B.~Li, W.~Xiong, W.~Hu, and R.~Ji, ``{RGBD} salient object detection:
  A benchmark and algorithms,'' in {\em ECCV}, 2014.

\bibitem{zhang2012microsoft}
Z.~Zhang, ``Microsoft kinect sensor and its effect,'' {\em IEEE multimedia},
  vol.~19, no.~2, pp.~4--10, 2012.

\bibitem{cheng2014depth}
Y.~Cheng, H.~Fu, X.~Wei, J.~Xiao, and X.~Cao, ``Depth enhanced saliency
  detection method,'' in {\em ICIMCS}, 2014.

\bibitem{NJU2000}
R.~Ju, Y.~Liu, T.~Ren, L.~Ge, and G.~Wu, ``Depth-aware salient object detection
  using anisotropic center-surround difference,'' {\em Signal Processing: Image
  Communication}, vol.~38, pp.~115--126, 2015.

\bibitem{sun2010secrets}
D.~Sun, S.~Roth, and M.~J. Black, ``Secrets of optical flow estimation and
  their principles,'' in {\em IEEE CVPR}, 2010.

\bibitem{li2014saliency}
N.~Li, J.~Ye, Y.~Ji, H.~Ling, and J.~Yu, ``Saliency detection on light field,''
  in {\em IEEE CVPR}, 2014.

\bibitem{ng2005light}
R.~Ng, M.~Levoy, M.~Br{\'e}dif, G.~Duval, M.~Horowitz, and P.~Hanrahan, {\em
  Light field photography with a hand-held plenoptic camera}.
\newblock PhD thesis, Stanford University, 2005.

\bibitem{sip_dataset}
D.-P. Fan, Z.~Lin, Z.~Zhang, M.~Zhu, and M.-M. Cheng, ``{Rethinking RGB-D
  Salient Object Detection: Models, Datasets, and Large-Scale Benchmarks},''
  {\em IEEE TNNLS}, vol.~32, no.~5, pp.~2075--2089, 2020.

\bibitem{vgg_network}
K.~Simonyan and A.~Zisserman, ``Very deep convolutional networks for
  large-scale image recognition,'' in {\em ICLR}, 2015.

\bibitem{cpd_sal}
Z.~Wu, L.~Su, and Q.~Huang, ``Cascaded partial decoder for fast and accurate
  salient object detection,'' in {\em IEEE CVPR}, 2019.

\bibitem{kim2015salient}
J.~Kim, D.~Han, Y.-W. Tai, and J.~Kim, ``Salient region detection via
  high-dimensional color transform and local spatial support,'' {\em IEEE TIP},
  vol.~25, no.~1, pp.~9--23, 2015.

\bibitem{transformer_nips}
A.~Vaswani, N.~Shazeer, N.~Parmar, J.~Uszkoreit, L.~Jones, A.~N. Gomez,
  L.~Kaiser, and I.~Polosukhin, ``Attention is all you need,'' in {\em
  NeurIPS}, 2017.

\bibitem{dpt_transformer}
R.~Ranftl, A.~Bochkovskiy, and V.~Koltun, ``Vision transformers for dense
  prediction,'' in {\em IEEE ICCV}, 2021.

\bibitem{liu2021swin}
Z.~Liu, Y.~Lin, Y.~Cao, H.~Hu, Y.~Wei, Z.~Zhang, S.~Lin, and B.~Guo, ``Swin
  transformer: Hierarchical vision transformer using shifted windows,'' in {\em
  IEEE ICCV}, 2021.

\bibitem{on_calibration}
C.~Guo, G.~Pleiss, Y.~Sun, and K.~Q. Weinberger, ``On calibration of modern
  neural networks,'' in {\em ICML}, 2017.

\bibitem{gan_raw}
I.~Goodfellow, J.~Pouget-Abadie, M.~Mirza, B.~Xu, D.~Warde-Farley, S.~Ozair,
  A.~Courville, and Y.~Bengio, ``Generative adversarial nets,'' in {\em
  NeurIPS}, 2014.

\bibitem{mcmc_langevin}
R.~Neal, ``Mcmc using hamiltonian dynamics,'' {\em Handbook of Markov Chain
  Monte Carlo}, 2011.

\bibitem{ABP}
T.~Han, Y.~Lu, S.-C. Zhu, and Y.~N. Wu, ``Alternating back-propagation for
  generator network,'' in {\em AAAI}, 2017.

\bibitem{kendall2017uncertainties}
A.~Kendall and Y.~Gal, ``What uncertainties do we need in bayesian deep
  learning for computer vision?,'' in {\em NeurIPS}, 2017.

\bibitem{cao2022deep}
S.~Cao and Z.~Zhang, ``Deep hybrid models for out-of-distribution detection,''
  in {\em IEEE CVPR}, 2022.

\bibitem{dosovitskiy_ViT_ICLR_2021}
A.~Dosovitskiy, L.~Beyer, A.~Kolesnikov, D.~Weissenborn, X.~Zhai,
  T.~Unterthiner, M.~Dehghani, M.~Minderer, G.~Heigold, S.~Gelly, J.~Uszkoreit,
  and N.~Houlsby, ``An image is worth 16x16 words: Transformers for image
  recognition at scale,'' in {\em ICLR}, 2021.

\bibitem{itti1998model}
L.~Itti, C.~Koch, and E.~Niebur, ``A model of saliency-based visual attention
  for rapid scene analysis,'' {\em IEEE TPAMI}, vol.~20, no.~11,
  pp.~1254--1259, 1998.

\bibitem{global_contrast}
M.-M. Cheng, N.~J. Mitra, X.~Huang, P.~H. Torr, and S.-M. Hu, ``Global contrast
  based salient region detection,'' {\em IEEE TPAMI}, vol.~37, no.~3,
  pp.~569--582, 2014.

\bibitem{ronneberger_unet_2015}
O.~Ronneberger, P.~Fischer, and T.~Brox, ``U-net: Convolutional networks for
  biomedical image segmentation,'' in {\em MICCAI}, 2015.

\bibitem{nldf_sal}
Z.~Luo, A.~Mishra, A.~Achkar, J.~Eichel, S.~Li, and P.-M. Jodoin, ``Non-local
  deep features for salient object detection,'' in {\em IEEE CVPR}, 2017.

\bibitem{picanet}
N.~Liu, J.~Han, and M.-H. Yang, ``Picanet: Learning pixel-wise contextual
  attention for saliency detection,'' in {\em IEEE CVPR}, 2018.

\bibitem{wang2020progressive}
B.~Wang, Q.~Chen, M.~Zhou, Z.~Zhang, X.~Jin, and K.~Gai, ``Progressive feature
  polishing network for salient object detection.,'' in {\em AAAI}, 2020.

\bibitem{hou_DSS_cvpr_2017}
Q.~Hou, M.-M. Cheng, X.~Hu, A.~Borji, Z.~Tu, and P.~H. Torr, ``Deeply
  supervised salient object detection with short connections,'' in {\em IEEE
  CVPR}, 2017.

\bibitem{wang_iccv_2017_stagewise}
T.~Wang, A.~Borji, L.~Zhang, P.~Zhang, and H.~Lu, ``A stagewise refinement
  model for detecting salient objects in images,'' in {\em IEEE ICCV}, 2017.

\bibitem{zhao_eccv_2020_suppress}
X.~Zhao, Y.~Pang, L.~Zhang, H.~Lu, and L.~Zhang, ``Suppress and balance: A
  simple gated network for salient object detection,'' in {\em ECCV}, 2020.

\bibitem{chen2020global}
Z.~Chen, Q.~Xu, R.~Cong, and Q.~Huang, ``Global context-aware progressive
  aggregation network for salient object detection,'' in {\em AAAI}, 2020.

\bibitem{zhang_saliency_hierarchy_ECCV_2022}
W.~Zhang, L.~Zheng, H.~Wang, X.~Wu, and X.~Li, ``Saliency hierarchy modeling
  via generative kernels for salient object detection,'' in {\em ECCV}, 2022.

\bibitem{qin2019basnet}
X.~Qin, Z.~Zhang, C.~Huang, C.~Gao, M.~Dehghan, and M.~Jagersand, ``Basnet:
  Boundary-aware salient object detection,'' in {\em IEEE CVPR}, 2019.

\bibitem{pang_MINet_CVPR_2020}
Y.~Pang, X.~Zhao, L.~Zhang, and H.~Lu, ``Multi-scale interactive network for
  salient object detection,'' in {\em IEEE CVPR}, 2020.

\bibitem{tang2021disentangled}
L.~Tang, B.~Li, Y.~Zhong, S.~Ding, and M.~Song, ``Disentangled high quality
  salient object detection,'' in {\em IEEE ICCV}, 2021.

\bibitem{xie_hed_iccv_2015}
S.~Xie and Z.~Tu, ``Holistically-nested edge detection,'' in {\em IEEE ICCV},
  2015.

\bibitem{liu_rcf_cvpr_2017}
Y.~Liu, M.-M. Cheng, X.~Hu, K.~Wang, and X.~Bai, ``Richer convolutional
  features for edge detection,'' in {\em IEEE CVPR}, 2017.

\bibitem{basnet_sal}
X.~Qin, Z.~Zhang, C.~Huang, C.~Gao, M.~Dehghan, and M.~Jagersand, ``{BASNet}:
  Boundary-aware salient object detection,'' in {\em IEEE CVPR}, 2019.

\bibitem{zhao2019EGNet}
J.-X. Zhao, J.-J. Liu, D.-P. Fan, Y.~Cao, J.~Yang, and M.-M. Cheng,
  ``{EGNet}:edge guidance network for salient object detection,'' in {\em IEEE
  ICCV}, 2019.

\bibitem{zhang_progressive_attention_cvpr_2018}
X.~Zhang, T.~Wang, J.~Qi, H.~Lu, and G.~Wang, ``Progressive attention guided
  recurrent network for salient object detection,'' in {\em IEEE CVPR}, 2018.

\bibitem{zhang2021auto}
M.~Zhang, T.~Liu, Y.~Piao, S.~Yao, and H.~Lu, ``Auto-msfnet: Search multi-scale
  fusion network for salient object detection,'' in {\em ACM MM}, 2021.

\bibitem{han_cnnbased_fusion_2017}
J.~Han, H.~Chen, N.~Liu, C.~Yan, and X.~Li, ``Cnns-based rgb-d saliency
  detection via cross-view transfer and multiview fusion,'' {\em IEEE
  Transactions on Cybernetics}, vol.~48, no.~11, pp.~3171--3183, 2017.

\bibitem{lee_Superpixel_RGBD_ECCV_2022}
M.~Lee, C.~Park, S.~Cho, and S.~Lee, ``Spsn: Superpixel prototype sampling
  network for rgb-d salient object detection,'' in {\em ECCV}, 2022.

\bibitem{qu2017rgbd}
L.~Qu, S.~He, J.~Zhang, J.~Tian, Y.~Tang, and Q.~Yang, ``{RGBD} salient object
  detection via deep fusion,'' {\em IEEE TIP}, vol.~26, no.~5, pp.~2274--2285,
  2017.

\bibitem{wang2019adaptive}
N.~Wang and X.~Gong, ``Adaptive fusion for {RGB-D} salient object detection,''
  {\em IEEE Access}, vol.~7, pp.~55277--55284, 2019.

\bibitem{han2017cnns}
J.~Han, H.~Chen, N.~Liu, C.~Yan, and X.~Li, ``{CNNs}-based {RGB-D} saliency
  detection via cross-view transfer and multiview fusion,'' {\em IEEE TOC},
  vol.~48, no.~11, pp.~3171--3183, 2017.

\bibitem{A2dele_cvpr2020}
Y.~Piao, Z.~Rong, M.~Zhang, W.~Ren, and H.~Lu, ``A2dele: Adaptive and attentive
  depth distiller for efficient {RGB-D} salient object detection,'' in {\em
  IEEE CVPR}, 2020.

\bibitem{chen2018progressively}
H.~Chen and Y.~Li, ``Progressively complementarity-aware fusion network for
  {RGB-D} salient object detection,'' in {\em IEEE CVPR}, 2018.

\bibitem{chen2019multi}
H.~Chen, Y.~Li, and D.~Su, ``Multi-modal fusion network with multi-scale
  multi-path and cross-modal interactions for {RGB-D} salient object
  detection,'' {\em Pattern Recognition}, vol.~86, pp.~376--385, 2019.

\bibitem{chen2019three}
H.~Chen and Y.~Li, ``Three-stream attention-aware network for {RGB-D} salient
  object detection,'' {\em IEEE TIP}, vol.~28, no.~6, pp.~2825--2835, 2019.

\bibitem{zhao2019Contrast}
J.-X. Zhao, Y.~Cao, D.-P. Fan, M.-M. Cheng, X.-Y. Li, and L.~Zhang, ``Contrast
  prior and fluid pyramid integration for {RGBD} salient object detection,'' in
  {\em IEEE CVPR}, 2019.

\bibitem{ssf_rgbd}
M.~Zhang, W.~Ren, Y.~Piao, Z.~Rong, and H.~Lu, ``Select, supplement and focus
  for {RGB-D} saliency detection,'' in {\em IEEE CVPR}, 2020.

\bibitem{self_attention_rgbd}
N.~Liu, N.~Zhang, L.~Shao, and J.~Han, ``Learning selective mutual attention
  and contrast for rgb-d saliency detection,'' {\em IEEE TPAMI}, vol.~44,
  no.~12, pp.~9026--9042, 2022.

\bibitem{ji2020accurate}
W.~Ji, J.~Li, M.~Zhang, Y.~Piao, and H.~Lu, ``Accurate {RGB-D} salient object
  detection via collaborative learning,'' in {\em ECCV}, 2020.

\bibitem{HDFNet-ECCV2020}
Y.~Pang, L.~Zhang, X.~Zhao, and H.~Lu, ``Hierarchical dynamic filtering network
  for {RGB-D} salient object detection,'' in {\em ECCV}, 2020.

\bibitem{zhang2020bilateral}
Z.~Zhang, Z.~Lin, J.~Xu, W.-D. Jin, S.-P. Lu, and D.-P. Fan, ``Bilateral
  attention network for {RGB-D} salient object detection,'' {\em IEEE TIP},
  vol.~30, pp.~1949--1961, 2021.

\bibitem{cmms_rgbd}
C.~Li, R.~Cong, Y.~Piao, Q.~Xu, and C.~C. Loy, ``{RGB-D} salient object
  detection with cross-modality modulation and selection,'' in {\em ECCV},
  2020.

\bibitem{Li_2020_CMWNet}
G.~Li, Z.~Liu, L.~Ye, Y.~Wang, and H.~Ling, ``Cross-modal weighting network for
  {RGB-D} salient object detection,'' in {\em ECCV}, 2020.

\bibitem{Luo2020CascadeGN}
A.~Luo, X.~Li, F.~Yang, Z.~Jiao, H.~Cheng, and S.~Lyu, ``Cascade graph neural
  networks for {RGB-D} salient object detection,'' in {\em ECCV}, 2020.

\bibitem{chen2020eccv}
S.~Chen and Y.~Fu, ``Progressively guided alternate refinement network for
  {RGB-D} salient object detection,'' in {\em ECCV}, 2020.

\bibitem{ATSA}
M.~Zhang, S.~X. Fei, J.~Liu, S.~Xu, Y.~Piao, and H.~Lu, ``Asymmetric two-stream
  architecture for accurate {RGB-D} saliency detection,'' in {\em ECCV}, 2020.

\bibitem{li2021hierarchical}
G.~Li, Z.~Liu, M.~Chen, Z.~Bai, W.~Lin, and H.~Ling, ``Hierarchical alternate
  interaction network for rgb-d salient object detection,'' {\em IEEE TIP},
  vol.~30, pp.~3528--3542, 2021.

\bibitem{fu2021siamese}
K.~Fu, D.-P. Fan, G.-P. Ji, Q.~Zhao, J.~Shen, and C.~Zhu, ``Siamese network for
  rgb-d salient object detection and beyond,'' {\em IEEE TPAMI}, vol.~44,
  no.~9, pp.~5541--5559, 2022.

\bibitem{carion_DETR_ECCV_2020}
N.~Carion, F.~Massa, G.~Synnaeve, N.~Usunier, A.~Kirillov, and S.~Zagoruyko,
  ``End-to-end object detection with transformers,'' in {\em ECCV}, 2020.

\bibitem{zhu_deformableDETR_ICLR_2021}
X.~Zhu, W.~Su, L.~Lu, B.~Li, X.~Wang, and J.~Dai, ``Deformable {DETR}:
  Deformable transformers for end-to-end object detection,'' in {\em ICLR},
  2021.

\bibitem{dai_UPDETR_CVPR_2021}
Z.~Dai, B.~Cai, Y.~Lin, and J.~Chen, ``{UP-DETR}: Unsupervised pre-training for
  object detection with transformers,'' in {\em IEEE CVPR}, 2021.

\bibitem{wang_PVT_2021}
W.~Wang, E.~Xie, X.~Li, D.-P. Fan, K.~Song, D.~Liang, T.~Lu, P.~Luo, and
  L.~Shao, ``Pyramid vision transformer: A versatile backbone for dense
  prediction without convolutions,'' in {\em IEEE ICCV}, 2021.

\bibitem{zhang_metaDETR_arxiv_2021}
G.~Zhang, Z.~Luo, K.~Cui, S.~Lu, and E.~P. Xing, ``Meta-detr: Image-level
  few-shot detection with inter-class correlation exploitation,'' {\em IEEE
  TPAMI}, 2022.

\bibitem{xu_TransCenterTracking_2021}
Y.~Xu, Y.~Ban, G.~Delorme, C.~Gan, D.~Rus, and X.~Alameda-Pineda,
  ``Transcenter: Transformers with dense representations for multiple-object
  tracking,'' {\em IEEE TPAMI}, 2022.

\bibitem{yan_SpatialTemporalTransformerTrackingv2_2021}
B.~Yan, H.~Peng, J.~Fu, D.~Wang, and H.~Lu, ``Learning spatio-temporal
  transformer for visual tracking,'' in {\em IEEE ICCV}, 2021.

\bibitem{mao_TFPose_2021}
W.~Mao, Y.~Ge, C.~Shen, Z.~Tian, X.~Wang, Z.~Wang, and A.~v. den Hengel,
  ``Poseur: Direct human pose estimation with transformers,'' in {\em ECCV},
  2022.

\bibitem{Jiang_GMAFlow_2021}
S.~Jiang, D.~Campbell, Y.~Lu, H.~Li, and R.~Hartley, ``Learning to estimate
  hidden motions with global motion aggregation,'' in {\em IEEE ICCV}, 2021.

\bibitem{zheng_SETR_2020}
S.~Zheng, J.~Lu, H.~Zhao, X.~Zhu, Z.~Luo, Y.~Wang, Y.~Fu, J.~Feng, T.~Xiang,
  P.~H. Torr, and L.~Zhang, ``Rethinking semantic segmentation from a
  sequence-to-sequence perspective with transformers,'' in {\em IEEE CVPR},
  2021.

\bibitem{liu_ICCV_2021_VST}
N.~Liu, N.~Zhang, K.~Wan, L.~Shao, and J.~Han, ``Visual saliency transformer,''
  in {\em IEEE ICCV}, 2021.

\bibitem{zhang2021learning_nips}
J.~Zhang, J.~Xie, N.~Barnes, and P.~Li, ``Learning generative vision
  transformer with energy-based latent space for saliency prediction,'' in {\em
  NeurIPS}, 2021.

\bibitem{VAE_Kingma}
D.~Kingma and M.~Welling, ``Auto-encoding variational bayes,'' in {\em ICLR},
  2014.

\bibitem{LeCun06atutorial}
Y.~LeCun, S.~Chopra, R.~Hadsell, M.~Ranzato, and F.~Huang, ``A tutorial on
  energy-based learning,'' {\em Predicting structured data}, vol.~1, no.~0,
  2006.

\bibitem{cvae}
K.~Sohn, H.~Lee, and X.~Yan, ``Learning structured output representation using
  deep conditional generative models,'' in {\em NeurIPS}, 2015.

\bibitem{PHiSeg2019}
C.~F. Baumgartner, K.~C. Tezcan, K.~Chaitanya, A.~M. H{\"{o}}tker, U.~J.
  Muehlematter, K.~Schawkat, A.~S. Becker, O.~Donati, and E.~Konukoglu,
  ``Phiseg: Capturing uncertainty in medical image segmentation,'' in {\em
  MICCAI}, 2019.

\bibitem{probabilistic_unet}
S.~Kohl, B.~Romera-Paredes, C.~Meyer, J.~De~Fauw, J.~R. Ledsam, K.~Maier-Hein,
  S.~M.~A. Eslami, D.~Jimenez~Rezende, and O.~Ronneberger, ``A probabilistic
  u-net for segmentation of ambiguous images,'' in {\em NeurIPS}, 2018.

\bibitem{SuperVAE_AAAI19}
B.~Li, Z.~Sun, and Y.~Guo, ``Supervae: Superpixelwise variational autoencoder
  for salient object detection,'' in {\em AAAI}, 2019.

\bibitem{groenendijk2020benefit}
R.~Groenendijk, S.~Karaoglu, T.~Gevers, and T.~Mensink, ``On the benefit of
  adversarial training for monocular depth estimation,'' {\em CVIU}, vol.~190,
  p.~102848, 2020.

\bibitem{gan_maskerrcnn}
Q.~H. Le, K.~Youcef-Toumi, D.~Tsetserukou, and A.~Jahanian, ``Gan mask
  r-cnn:instance semantic segmentation benefits from generative adversarial
  networks,'' in {\em NeurIPS Workshop}, 2020.

\bibitem{gan_semi_seg}
N.~Souly, C.~Spampinato, and M.~Shah, ``Semi supervised semantic segmentation
  using generative adversarial network,'' in {\em IEEE ICCV}, 2017.

\bibitem{hung2018adversarial}
W.-C. Hung, Y.-H. Tsai, Y.-T. Liou, Y.-Y. Lin, and M.-H. Yang, ``Adversarial
  learning for semi-supervised semantic segmentation,'' in {\em BMVC}, 2018.

\bibitem{chen_IWGAN_arxiv_2021}
Y.~Chen, Q.~Gao, X.~Wang, {\em et~al.}, ``Inferential wasserstein generative
  adversarial networks,'' {\em Journal of the Royal Statistical Society Series
  B}, vol.~84, no.~1, pp.~83--113, 2022.

\bibitem{Lagging_Inference_Networks}
J.~He, D.~Spokoyny, G.~Neubig, and T.~Berg-Kirkpatrick, ``Lagging inference
  networks and posterior collapse in variational autoencoders,'' in {\em ICLR},
  2019.

\bibitem{hsu122017weakly}
K.-J. Hsu12, Y.-Y. Lin, and Y.-Y. Chuang, ``Weakly supervised saliency
  detection with a category-driven map generator,'' {\em BMVC}, 2017.

\bibitem{Guanbin_weaksalAAAI}
G.~Li, Y.~Xie, and L.~Lin, ``Weakly supervised salient object detection using
  image labels,'' in {\em AAAI}, 2018.

\bibitem{ahn2018learning}
J.~Ahn and S.~Kwak, ``Learning pixel-level semantic affinity with image-level
  supervision for weakly supervised semantic segmentation,'' in {\em IEEE
  CVPR}, 2018.

\bibitem{huang2018weakly}
Z.~Huang, X.~Wang, J.~Wang, W.~Liu, and J.~Wang, ``Weakly-supervised semantic
  segmentation network with deep seeded region growing,'' in {\em IEEE CVPR},
  2018.

\bibitem{zhang2021learning}
H.~Zhang, Y.~Zeng, H.~Lu, L.~Zhang, J.~Li, and J.~Qi, ``Learning to detect
  salient object with multi-source weak supervision,'' {\em IEEE TPAMI},
  vol.~44, no.~7, pp.~3577--3589, 2022.

\bibitem{song2019box}
C.~Song, Y.~Huang, W.~Ouyang, and L.~Wang, ``Box-driven class-wise region
  masking and filling rate guided loss for weakly supervised semantic
  segmentation,'' in {\em IEEE CVPR}, 2019.

\bibitem{dai2015boxsup}
J.~Dai, K.~He, and J.~Sun, ``Boxsup: Exploiting bounding boxes to supervise
  convolutional networks for semantic segmentation,'' in {\em IEEE ICCV}, 2015.

\bibitem{kulharia2020box2seg}
V.~Kulharia, S.~Chandra, A.~Agrawal, P.~Torr, and A.~Tyagi, ``Box2seg:
  Attention weighted loss and discriminative feature learning for weakly
  supervised segmentation,'' in {\em ECCV}, 2020.

\bibitem{lee2021bbam}
J.~Lee, J.~Yi, C.~Shin, and S.~Yoon, ``Bbam: Bounding box attribution map for
  weakly supervised semantic and instance segmentation,'' in {\em IEEE CVPR},
  2021.

\bibitem{tian2021boxinst}
Z.~Tian, C.~Shen, X.~Wang, and H.~Chen, ``Boxinst: High-performance instance
  segmentation with box annotations,'' in {\em IEEE CVPR}, 2021.

\bibitem{lin2016scribblesup}
D.~Lin, J.~Dai, J.~Jia, K.~He, and J.~Sun, ``Scribblesup: Scribble-supervised
  convolutional networks for semantic segmentation,'' in {\em IEEE CVPR}, 2016.

\bibitem{vernaza2017learning}
P.~Vernaza and M.~Chandraker, ``Learning random-walk label propagation for
  weakly-supervised semantic segmentation,'' in {\em IEEE CVPR}, 2017.

\bibitem{structure_consistency_scribble}
S.~Yu, B.~Zhang, J.~Xiao, and E.~G. Lim, ``Structure-consistent weakly
  supervised salient object detection with local saliency coherence,'' in {\em
  AAAI}, 2021.

\bibitem{bearman2016s}
A.~Bearman, O.~Russakovsky, V.~Ferrari, and L.~Fei-Fei, ``What’s the point:
  Semantic segmentation with point supervision,'' in {\em ECCV}, 2016.

\bibitem{chen2021seminar}
H.~Chen, J.~Wang, H.~C. Chen, X.~Zhen, F.~Zheng, R.~Ji, and L.~Shao, ``Seminar
  learning for click-level weakly supervised semantic segmentation,'' in {\em
  IEEE ICCV}, 2021.

\bibitem{obukhov2019gated}
A.~Obukhov, S.~Georgoulis, D.~Dai, and L.~{Van Gool}, ``Gated {CRF} loss for
  weakly supervised semantic image segmentation,'' in {\em NeurIPS}, 2019.

\bibitem{arbelaez2014multiscale}
P.~Arbel{\'a}ez, J.~Pont-Tuset, J.~T. Barron, F.~Marques, and J.~Malik,
  ``Multiscale combinatorial grouping,'' in {\em IEEE CVPR}, 2014.

\bibitem{rother2004grabcut}
C.~Rother, V.~Kolmogorov, and A.~Blake, ``"grabcut" interactive foreground
  extraction using iterated graph cuts,'' {\em ACM TOG}, vol.~23, no.~3,
  pp.~309--314, 2004.

\bibitem{zhang2020learningeccv}
J.~Zhang, J.~Xie, and N.~Barnes, ``Learning noise-aware encoder-decoder from
  noisy labels by alternating back-propagation for saliency detection,'' in
  {\em ECCV}, 2020.

\bibitem{rca_eccv}
Y.~Zhang, K.~Li, K.~Li, L.~Wang, B.~Zhong, and Y.~Fu, ``Image super-resolution
  using very deep residual channel attention networks,'' in {\em ECCV}, 2018.

\bibitem{denseaspp}
M.~Yang, K.~Yu, C.~Zhang, Z.~Li, and K.~Yang, ``Denseaspp for semantic
  segmentation in street scenes,'' in {\em IEEE CVPR}, 2018.

\bibitem{Godard_2017_CVPR}
C.~Godard, O.~Mac~Aodha, and G.~J. Brostow, ``Unsupervised monocular depth
  estimation with left-right consistency,'' in {\em IEEE CVPR}, 2017.

\bibitem{higgins2017betavae}
I.~Higgins, L.~Matthey, A.~Pal, C.~Burgess, X.~Glorot, M.~Botvinick,
  S.~Mohamed, and A.~Lerchner, ``beta-{VAE}: Learning basic visual concepts
  with a constrained variational framework,'' in {\em ICLR}, 2017.

\bibitem{yan2013hierarchical}
Q.~Yan, L.~Xu, J.~Shi, and J.~Jia, ``Hierarchical saliency detection,'' in {\em
  IEEE CVPR}, 2013.

\bibitem{Manifold-Ranking:CVPR-2013}
C.~Yang, L.~Zhang, H.~Lu, X.~Ruan, and M.-H. Yang, ``Saliency detection via
  graph-based manifold ranking,'' in {\em IEEE CVPR}, 2013.

\bibitem{li2015visual}
G.~Li and Y.~Yu, ``Visual saliency based on multiscale deep features,'' in {\em
  IEEE CVPR}, 2015.

\bibitem{pascal_s_dataset}
Y.~Li, X.~Hou, C.~Koch, J.~M. Rehg, and A.~L. Yuille, ``The secrets of salient
  object segmentation,'' in {\em IEEE CVPR}, 2014.

\bibitem{sod_dataset}
V.~Movahedi and J.~H. Elder, ``Design and perceptual validation of performance
  measures for salient object segmentation,'' in {\em IEEE CVPR Workshop},
  2010.

\bibitem{zhou2020interactive}
H.~Zhou, X.~Xie, J.-H. Lai, Z.~Chen, and L.~Yang, ``Interactive two-stream
  decoder for accurate and fast saliency detection,'' in {\em IEEE CVPR}, 2020.

\bibitem{zhao_CTDNet_ACMMM_2021}
Z.~Zhao, C.~Xia, C.~Xie, and J.~Li, ``Complementary trilateral decoder for fast
  and accurate salient object detection,'' in {\em ACM MM}, 2021.

\bibitem{xin2018c2s}
X.~Li, F.~Yang, H.~Cheng, W.~Liu, and D.~Shen, ``Contour knowledge transfer for
  salient object detection,'' in {\em ECCV}, 2018.

\bibitem{fan2018enhanced}
D.-P. Fan, C.~Gong, Y.~Cao, B.~Ren, M.-M. Cheng, and A.~Borji,
  ``Enhanced-alignment measure for binary foreground map evaluation,'' in {\em
  IJCAI}, 2018.

\bibitem{fan2017structure}
D.-P. Fan, M.-M. Cheng, Y.~Liu, T.~Li, and A.~Borji, ``Structure-measure: A new
  way to evaluate foreground maps,'' in {\em IEEE ICCV}, 2017.

\bibitem{franchi2022latent}
G.~Franchi, X.~Yu, A.~Bursuc, E.~Aldea, S.~Dubuisson, and D.~Filliat, ``Latent
  discriminant deterministic uncertainty,'' in {\em ECCV}, 2022.

\bibitem{degroot1983comparison}
M.~H. DeGroot and S.~E. Fienberg, ``The comparison and evaluation of
  forecasters,'' {\em Journal of the Royal Statistical Society: Series D (The
  Statistician)}, vol.~32, no.~1-2, pp.~12--22, 1983.

\bibitem{uncertainty_decomposition}
S.~Depeweg, J.-M. Hernandez-Lobato, F.~Doshi-Velez, and S.~Udluft,
  ``Decomposition of uncertainty in {B}ayesian deep learning for efficient and
  risk-sensitive learning,'' in {\em ICML}, 2018.

\bibitem{smoothness_loss}
Y.~Wang, Y.~Yang, Z.~Yang, L.~Zhao, P.~Wang, and W.~Xu, ``Occlusion aware
  unsupervised learning of optical flow,'' in {\em IEEE CVPR}, 2018.

\bibitem{naseer_IntriguingProperties_Arxiv_2021}
M.~Naseer, K.~Ranasinghe, S.~Khan, M.~Hayat, F.~S. Khan, and M.-H. Yang,
  ``Intriguing properties of vision transformers,'' in {\em NeurIPS}, 2021.

\bibitem{imagenet_1k}
J.~Deng, W.~Dong, R.~Socher, L.-J. Li, K.~Li, and L.~Fei-Fei, ``Imagenet: A
  large-scale hierarchical image database,'' in {\em IEEE CVPR}, 2009.

\bibitem{goodfellow2014explaining}
I.~J. Goodfellow, J.~Shlens, and C.~Szegedy, ``Explaining and harnessing
  adversarial examples,'' in {\em ICLR}, 2014.

\bibitem{madry2018towards}
A.~Madry, A.~Makelov, L.~Schmidt, D.~Tsipras, and A.~Vladu, ``Towards deep
  learning models resistant to adversarial attacks,'' in {\em ICLR}, 2018.

\bibitem{Bayesian_Uncertainty}
W.~Wright, ``Bayesian approach to neural-network modeling with input
  uncertainty,'' {\em IEEE TNN}, vol.~10, no.~6, pp.~1261--1270, 1999.

\bibitem{bayesian_conv}
Y.~Gal and Z.~Ghahramani, ``Bayesian convolutional neural networks with
  bernoulli approximate variational inference,'' {\em arXiv preprint
  arXiv:1506.02158}, 2015.

\bibitem{generalized_bnn}
M.~Vadera, B.~Jalaian, and B.~Marlin, ``Generalized bayesian posterior
  expectation distillation for deep neural networks,'' in {\em CUAI}, 2020.

\bibitem{Jordan99anintroduction}
M.~I. Jordan, Z.~Ghahramani, and et~al., ``An introduction to variational
  methods for graphical models,'' in {\em Machine Learning}, pp.~183--233, MIT
  Press, 1999.

\bibitem{Graphical_Models_VI}
M.~J. Wainwright and M.~I. Jordan, ``Graphical models, exponential families,
  and variational inference,'' {\em Foundations and Trends® in Machine
  Learning}, vol.~1, no.~1–2, pp.~1--305, 2008.

\bibitem{lee2022gibbs}
S.~Y. Lee, ``Gibbs sampler and coordinate ascent variational inference: A
  set-theoretical review,'' {\em Communications in Statistics-Theory and
  Methods}, vol.~51, no.~6, pp.~1549--1568, 2022.

\bibitem{mcmc_sampling}
W.~K. Hastings, ``{Monte Carlo sampling methods using Markov chains and their
  applications},'' {\em Biometrika}, vol.~57, pp.~97--109, 04 1970.

\bibitem{gal2016dropout}
Y.~Gal and Z.~Ghahramani, ``Dropout as a bayesian approximation: Representing
  model uncertainty in deep learning,'' in {\em ICML}, 2016.

\bibitem{deep_gaussian_process}
A.~Damianou and N.~D. Lawrence, ``Deep {G}aussian processes,'' in {\em
  AISTATS}, 2013.

\bibitem{szegedy2013intriguing}
C.~Szegedy, W.~Zaremba, I.~Sutskever, J.~Bruna, D.~Erhan, I.~Goodfellow, and
  R.~Fergus, ``Intriguing properties of neural networks,'' in {\em ICLR}, 2014.

\bibitem{wong2018provable}
E.~Wong and Z.~Kolter, ``Provable defenses against adversarial examples via the
  convex outer adversarial polytope,'' in {\em ICLR}, 2018.

\bibitem{amirul2018revisiting}
M.~A. Islam, M.~Kalash, and N.~D. Bruce, ``Revisiting salient object detection:
  Simultaneous detection, ranking, and subitizing of multiple salient
  objects,'' in {\em IEEE CVPR}, 2018.

\end{thebibliography}
}

\end{document}